\documentclass[letterpaper]{article} 
\usepackage[preprint]{aaai2027}  
\usepackage[hyphens]{url}  
\usepackage{graphicx} 
\usepackage{natbib}  
\usepackage{caption} 
\usepackage{algorithm}
\usepackage{algorithmic}

\usepackage{amsmath} 
\usepackage{xspace}
\usepackage{amssymb}
\usepackage{multirow}
\usepackage{cleveref}
\usepackage{subcaption}
\usepackage[table]{xcolor}
\usepackage[most]{tcolorbox}
\usepackage{mathtools}

\usepackage{newfloat}
\usepackage{listings}
\DeclareCaptionStyle{ruled}{labelfont=normalfont,labelsep=colon,strut=off} 
\floatstyle{ruled}
\newfloat{listing}{tb}{lst}{}
\floatname{listing}{Listing}

\usepackage{booktabs}

\title{DIVE: Unlocking Self-Improvement in Frozen Language Models Through Diversity-Driven Skill Evolution}
\author{
Siheng Xiong\textsuperscript{\rm 1},
Ali Payani\textsuperscript{\rm 2},
Oguzhan Gungordu\textsuperscript{\rm 1},
Faramarz Fekri\textsuperscript{\rm 1}
}

\affiliations{
\textsuperscript{\rm 1}Georgia Institute of Technology \qquad
\textsuperscript{\rm 2}Cisco Research\\
\texttt{sxiong45@gatech.edu
apayani@cisco.com
ogungordu3@gatech.edu
fekri@ece.gatech.edu}
}

\newcommand{\Ours}{\textsc{DIVE}\xspace}
\definecolor{promptbg}{RGB}{233,240,255}

\begin{document}

\maketitle

\begin{abstract}
Large language models (LLMs) cannot retain post-deployment experience without parameter updates. 
We introduce \Ours, a diversity-driven framework that enables frozen LLMs to improve by evolving persistent natural-language skills from task experience and verifier feedback. 
These skills encode reusable reasoning procedures, verification strategies, common failure modes, and output constraints and are both executed and revised by the same underlying model without access to a teacher model. 
Since natural-language skill evolution is a stochastic, non-convex search process, optimizing a single skill trajectory can overfit to sampled experience or converge to a suboptimal solution. 
\Ours mitigates this optimization variance by independently evolving multiple skill populations from bootstrapped experience, adaptively refining them through diverse transformations, and jointly selecting a complementary set of skills. 
Across six mathematical and logical reasoning tasks and multiple model families, \Ours consistently outperforms existing reasoning methods, prompt-optimization approaches, skill-development frameworks, and memory-based baselines. 
It achieves rapid self-improvement from accumulated experience, obtaining substantially larger performance gains with fewer rollouts than parameter-based methods such as SFT and GRPO, and prompt optimization with GEPA. 
Further, the resulting skills transfer across model scales and families, enabling smaller models such as GPT-5-nano to match or outperform larger counterparts, i.e., GPT-5, under conventional prompting. 
These results establish diversity-driven skill evolution as an effective, interpretable, and parameter-free approach to LLM self-improvement.
\end{abstract}

\section{Introduction}
\label{sec:introduction}

Large language models (LLMs) exhibit broad reasoning capabilities~\citep{yang2024can,yang2024harnessing,xiong2024large,yu2025causaleval,bao2025conflict,xiong2025enhancing,xiong2026enhancing}, yet their behavior after deployment is largely static. 
When a model repeatedly encounters examples from a task, receives feedback, or discovers a useful solution strategy, that experience does not modify its parameters or automatically persist across future queries. 
Fine-tuning can internalize such experience, but it requires access to model weights, substantial computation, and a carefully designed training pipeline. 
These requirements are increasingly restrictive as many capable language models are accessible only through APIs or deployed under limited adaptation budgets. 
This motivates a complementary question: \emph{can a frozen language model achieve self-improvement by converting experience into persistent natural-language skills?}

\begin{figure}[t]
\centering
\begin{subfigure}[t]{0.49\linewidth}
    \centering
    \includegraphics[width=\linewidth]{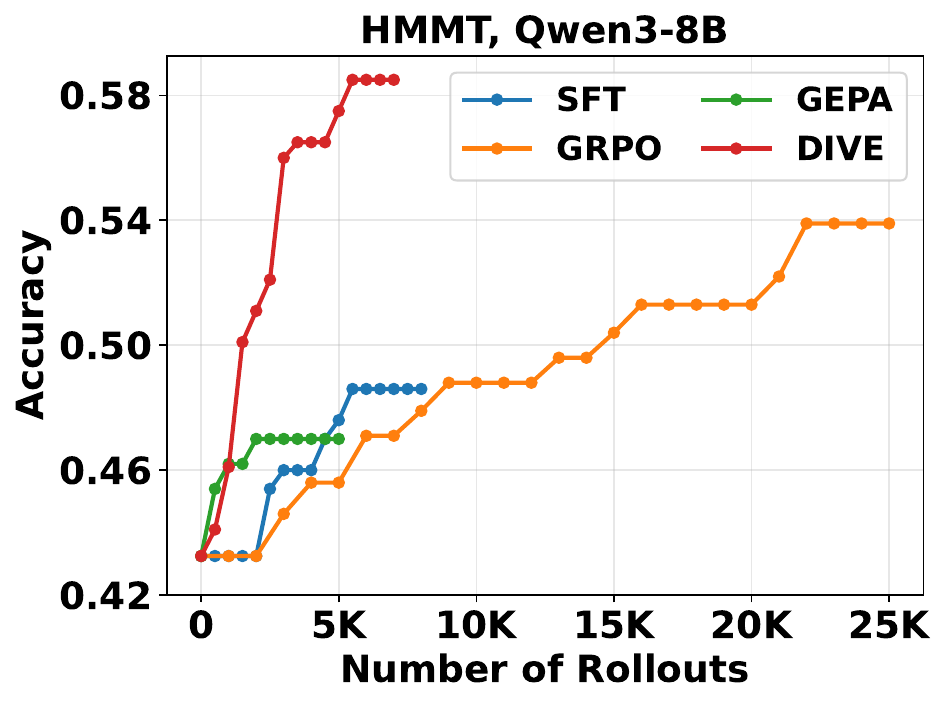}
    \caption{HMMT, Qwen3-8B}
\end{subfigure}
\hfill
\begin{subfigure}[t]{0.49\linewidth}
    \centering
    \includegraphics[width=\linewidth]{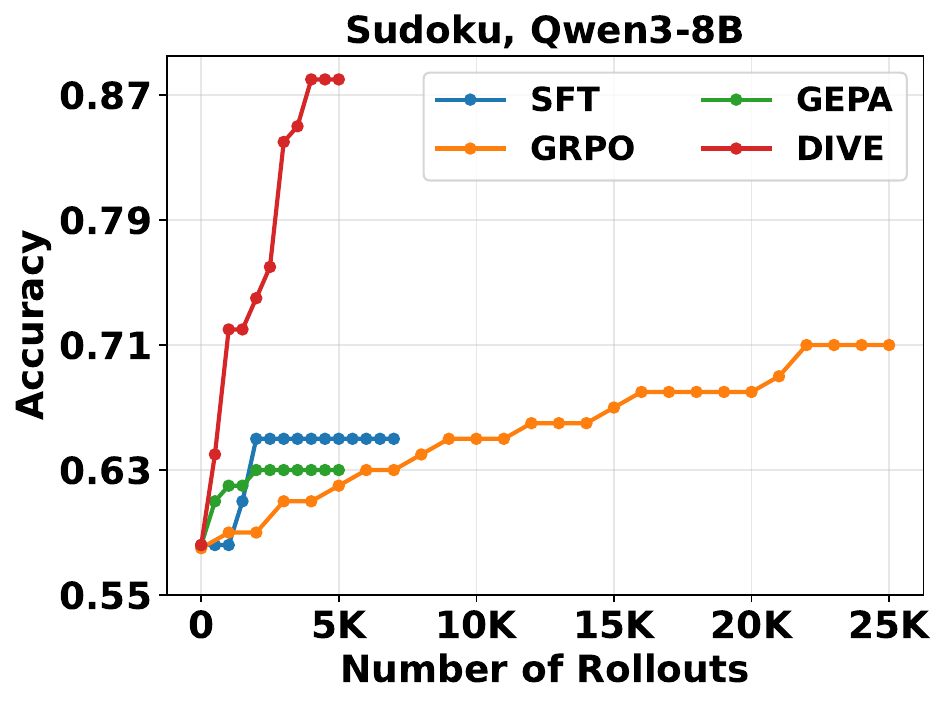}
    \caption{Sudoku, Qwen3-8B}
\end{subfigure}
\caption{Optimization efficiency on HMMT and Sudoku using Qwen3-8B. \Ours enables rapid self-improvement from accumulated experience, achieving substantially larger performance gains with fewer rollouts than parameter-based optimization (SFT, GRPO) and prompt optimization (GEPA).}
\label{fig:acc_rollout_curves}
\end{figure}

Prior work on textual reflection and prompt optimization suggests that natural-language context can serve as a writable substrate through adapting frozen language models~\citep{shinn2023reflexion,miprov2,agrawal2025gepa}.
Reliable self-improvement, however, requires more than generating a better prompt.
We consider the challenging setting in which the same frozen model must solve the task, interpret verifier feedback, and revise its own knowledge without access to a stronger teacher. 
This setting raises three key challenges. 
{First, self-generated revisions are inherently noisy:}
a locally beneficial edit may remove useful guidance, overfit to a small set of failures, or amplify an incorrect reflection, making greedy single-candidate refinement brittle. 
{Second, accumulated experience quickly exceeds the context budget, requiring demonstrations, traces, and feedback to be distilled into compact, reusable abstractions rather than appended indefinitely~\citep{xiong2025long,xiong2026adaptive}.}
{Third, skill evolution is path-dependent}: different initial skills, sampled experiences, and revision trajectories can lead to substantially different solutions.
Maintaining only a single evolving trajectory can therefore prematurely discard promising alternatives or converge to a suboptimal solution.
Effective self-improvement therefore requires preserving diverse solution trajectories throughout skill evolution.

We introduce \Ours, a diversity-driven framework for self-improvement of frozen LLMs through natural-language skill evolution. 
\Ours represents accumulated experience as persistent skills that encode reusable reasoning procedures, verification strategies, common failure modes, and output constraints. 
The same LLM both executes and revises these skills using task experience and verifier feedback, while its parameters remain fixed throughout optimization.

The central principle of \Ours is to preserve diversity throughout skill evolution and exploit the resulting complementarity at inference time. 
Rather than refining a single incumbent skill, \Ours independently evolves multiple skill populations from bootstrapped experience. 
Within each population, a portfolio of heterogeneous evolution operators, including Reflective Repair, Exploratory Revision, Compression, and Multi-Parent Recombination, proposes alternative revisions. 
An upper-confidence-bound policy adaptively allocates proposal budget across these operators based on their observed utility, while new operators can be generated from accumulated evolution history.

After evolution, candidate skills from all populations are evaluated on a shared validation set and jointly selected to form a compact, complementary skill set. 
At inference time, the selected skills independently generate candidate solutions, which are ranked by the frozen model to select the final response. 
In this way, \Ours uses diversity both to reduce the brittleness of textual skill optimization and to improve the reliability of downstream predictions through complementary skill hypotheses.

Unlike prior prompt optimization that primarily searches for a single improved prompt, \Ours maintains and evolves multiple skill hypotheses through independent populations, adaptive transformation operators, and joint skill selection.
We evaluate \Ours across six mathematical and logical reasoning tasks and multiple model families. 
\Ours consistently improves held-out performance without parameter updates, outperforming inference-time reasoning, skill-learning, memory-based, and prompt-optimization baselines. 
Ablations further demonstrate the complementary benefits of the key design components.

Our main contributions are:

\begin{itemize}

\item We formulate self-improvement of frozen LLMs as the evolution of persistent natural-language skills, enabling models to accumulate reusable task knowledge from experience and verifier feedback without weight updates or a stronger teacher model.

\item We introduce \Ours, a diversity-driven framework that independently evolves multiple skill populations from bootstrapped experience and adaptively allocates the evolution budget across heterogeneous transformation operators, preserving diverse skill hypotheses and evolution trajectories.

\item We jointly select a complementary set of skills, translating diversity in the evolution process into improved held-out performance. Experiments across diverse reasoning tasks and model families further demonstrate effective self-improvement and cross-model skill transfer.

\end{itemize}

\begin{figure*}[t]
    \centering
    \includegraphics[width=0.98\textwidth]{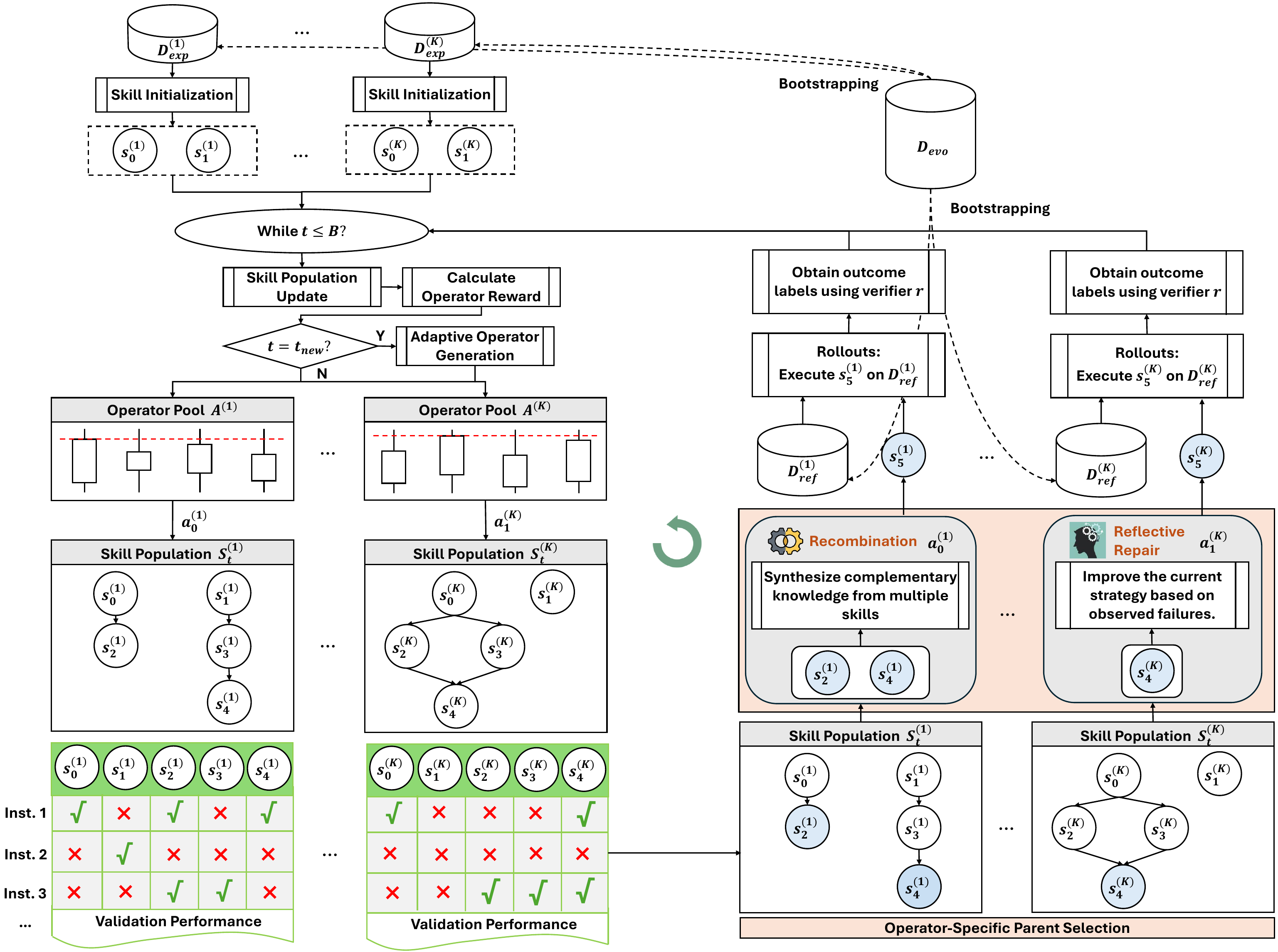}
    \caption{
    Overview of \Ours.
    We construct multiple independent skill populations from bootstrapped experience,
    evolve each population using adaptively selected operators,
    and jointly select complementary skills as the final skill set.
    }
    \label{fig:framework}
\end{figure*}

\section{Preliminaries}
\label{sec:preliminaries}

\subsection{Problem Definition}
\label{sec:problem_definition}

Given a question $x$, a language model $f_\theta$ produces a response $f_\theta(x)$.
An external verifier with access to the gold answer $y$ evaluates the response and returns a binary score
\begin{equation}
r(x,y,f_\theta(x))\in\{0,1\},
\end{equation}
where $r=1$ indicates a correct response and $r=0$ otherwise.
Depending on the task, the verifier may additionally return structured diagnostic signals, such as format violations, execution failures, or timeouts.
Importantly, the verifier does NOT provide natural-language critiques.

For a task distribution $\mathcal{T}$ over question--answer pairs $(x,y)$, we define the performance of the frozen model as
\begin{equation}
V_{\mathcal{T}}(f_\theta)
=
\mathbb{E}_{(x,y)\sim\mathcal{T}}
\left[
r\left(x,y,f_\theta(x)\right)
\right].
\end{equation}
Our goal is to improve task performance while keeping the model parameters $\theta$ fixed.

\subsection{Skill-Based Self-Improvement of Frozen LLMs}
\label{sec:skill_self_improvement}

To achieve this goal, the model must accumulate and reuse task-specific knowledge from experience and verifier feedback. 
We represent this acquired knowledge as a natural-language skill $s$. 
A skill is a compact textual artifact that guides subsequent inference by encoding reusable reasoning procedures, verification strategies, common failure modes, and output constraints.
The performance of a skill $s$ on task distribution $\mathcal{T}$ is
\begin{equation}
V_{\mathcal{T}}(s)
=\mathbb{E}_{(x,y)\sim\mathcal{T}}\left[r(x,y,f_{\theta}(x;s))\right],
\end{equation}
where $f_{\theta}(x;s)$ denotes the response generated by the frozen model conditioned on skill $s$.

Given a dataset $\mathcal{D}$ and a skill $s$, we denote by
\begin{equation}
\mathcal{E}(\mathcal{D};s)=\{(x,f_\theta(x;s),r(x,y,f_\theta(x;s)))\}_{(x,y)\in\mathcal{D}}
\end{equation}
the set of verifier-labeled skill-conditioned trajectories (we analogously write $\mathcal{E}(\mathcal{D})$ as zero-shot trajectories);
the same model proposes a revised skill as
\begin{equation}
s'=\operatorname{UpdateSkill}_{f_\theta}(s,\mathcal{E}(\mathcal{D};s)).
\end{equation}
Compared with weight-space self-improvement, this setting {remains applicable when model weights or training infrastructure are unavailable}, as is often the case for proprietary models.
Moreover, the resulting skills are {human-readable, editable, reversible, and transferable across models}.

\section{Methodology}
\label{sec:methodology}

\subsection{Overview}
\label{sec:method_overview}
We introduce \Ours, a framework that improves a frozen language model by evolving multiple skills and jointly selecting a complementary set of skills.
Skill evolution is a stochastic, non-convex search process in which different initializations, sampled examples, and revision trajectories can produce substantially different solutions.
\Ours therefore preserves multiple independent evolution trajectories and exploits their complementary strengths rather than relying on a single optimization run.

Specifically, we first partition the development data into an evolution set $\mathcal{D}_{\mathrm{evo}}$ and a validation set $\mathcal{D}_{\mathrm{val}}$, while reserving $\mathcal{D}_{\mathrm{test}}$ exclusively for final evaluation.
We then construct $K$ independent skill populations from bootstrapped subsets of $\mathcal{D}_{\mathrm{evo}}$.
Each population is evolved using a portfolio of heterogeneous evolution operators whose proposal budget is adaptively allocated via upper confidence bounds. 
After evolution, all candidate skills are evaluated on the shared validation set, from which we jointly construct a complementary final skill set.
At inference time, the selected skills independently generate candidate solutions, which are ranked by the frozen model to select the final response.
\textbf{Additional methodology details and theoretical analysis are provided in the supplementary material.}

\subsection{Independent Skill Population Construction}
\label{sec:independent_populations}

For each population $k\in[K]$, we independently sample a bootstrapped experience subset $\mathcal{D}_{\mathrm{exp}}^{(k)}$ and a bootstrapped reflection subset $\mathcal{D}_{\mathrm{ref}}^{(k)}$ from $\mathcal{D}_{\mathrm{evo}}$.
The two subsets serve distinct roles. 
The experience subset provides verifier-labeled trajectories from which reusable task knowledge is extracted, while the reflection subset is used to evaluate candidate skills, diagnose failures, and guide subsequent revisions. 
Together with an evolution budget $B$, these subsets define the $k$-th evolved population:
\begin{equation}
\mathcal{S}_B^{(k)} =\operatorname{DevelopSkill}_{f_\theta}\left(\mathcal{D}_{\mathrm{exp}}^{(k)},\mathcal{D}_{\mathrm{ref}}^{(k)}\right).
\label{eq:independent_population}
\end{equation}
Although all populations optimize the same task objective, their distinct bootstrapped subsets and evolution trajectories encourage the emergence of different reasoning procedures and verification strategies.

\paragraph{Seed Skill Initialization.}
For each population $k$, we construct an initial seed set $\mathcal{S}^{(k)}_0$ by distilling reusable task knowledge from the verifier-labeled trajectories collected on $\mathcal{D}_{\mathrm{exp}}^{(k)}$:
\begin{equation}
\mathcal{S}^{(k)}_0
=
\operatorname{InitializeSkill}_{f_\theta}\left(
\mathcal{E}(\mathcal{D}_{\mathrm{exp}}^{(k)})\right).
\label{eq:skill_initialization}
\end{equation}
The resulting seeds capture complementary aspects of the experience, including successful reasoning patterns, verification strategies, recurring failure modes, and output constraints, yielding an initial population with diverse inductive biases.

\subsection{Adaptive Skill Evolution}
\label{sec:adaptive_operator}

Each skill population is evolved independently.
For clarity, we consider a single population and omit the population superscript $k$ throughout this subsection.
At each evolution step $t$, the model uses verifier-labeled trajectories from the corresponding reflection subset to evaluate current skills, diagnose their failures, and propose candidate revisions.

\begin{table*}[t]
\centering
\resizebox{0.8\textwidth}{!}{
\begin{tabular}{l|l|cccccc|c}
\toprule
\textbf{Model} & \textbf{Method}
& \multicolumn{2}{c}{\textbf{Mathematical Reasoning}}
& \multicolumn{4}{c}{\textbf{Logical Reasoning}}
& \textbf{Avg.} \\
\cmidrule(lr){3-4}
\cmidrule(lr){5-8}
& & \textbf{HMMT} & \textbf{Equational Theories}
& \textbf{Sudoku} & \textbf{Cryptarithm}
& \textbf{Calcudoku} & \textbf{Futoshiki}
& \\
\midrule

\multirow{12}{*}{{GPT-5-nano}}
& Zero-shot
& 64.3 & 56.5 & 71.7 & 20.1 & 30.4 & 70.8 & 52.3 \\
& Few-shot
& 63.3 & 59.5 & 77.0 & 25.5 & 26.2 & 69.1 & 53.4 \\
& SC
& 80.1 & 59.1 & 86.5 & 48.9 & 49.1 & 82.7 & 67.7 \\
& ToT
& 74.3 & 62.3 & 92.8 & 55.1 & 68.4 & 93.1 & 74.3 \\
& Experience RAG
& 55.0 & 64.0 & 74.1 & 23.8 & 28.5 & 67.5 & 52.2 \\
& ExpeL
& 60.1 & 61.6 & 77.2 & 26.7 & 37.0 & 75.4 & 56.3 \\
& Direct Skill
& 57.5 & 55.3 & 70.3 & 24.7 & 38.7 & 71.7 & 53.0 \\
& SkillOpt
& 66.3 & 59.1 & 82.6 & 29.9 & 39.8 & 80.4 & 59.7 \\
& MIPROv2
& 59.0 & 55.2 & 75.1 & 24.4 & 34.7 & 74.5 & 53.8 \\
& GEPA
& 61.3 & 56.0 & 78.6 & 25.6 & 31.8 & 77.2 & 55.1 \\
& {\cellcolor[rgb]{0.925,0.957,1}}{\Ours ($M=1$)}
& {\cellcolor[rgb]{0.925,0.957,1}}74.3
& {\cellcolor[rgb]{0.925,0.957,1}}60.1
& {\cellcolor[rgb]{0.925,0.957,1}}85.8
& {\cellcolor[rgb]{0.925,0.957,1}}28.2
& {\cellcolor[rgb]{0.925,0.957,1}}36.8
& {\cellcolor[rgb]{0.925,0.957,1}}78.7
& {\cellcolor[rgb]{0.925,0.957,1}}60.7 \\
& {\cellcolor[rgb]{0.925,0.957,1}}{\Ours ($M=10$)}
& {\cellcolor[rgb]{0.925,0.957,1}}\textbf{82.9}
& {\cellcolor[rgb]{0.925,0.957,1}}\textbf{64.5}
& {\cellcolor[rgb]{0.925,0.957,1}}\textbf{96.0}
& {\cellcolor[rgb]{0.925,0.957,1}}\textbf{61.5}
& {\cellcolor[rgb]{0.925,0.957,1}}\textbf{84.8}
& {\cellcolor[rgb]{0.925,0.957,1}}\textbf{99.2}
& {\cellcolor[rgb]{0.925,0.957,1}}\textbf{81.5} \\

\midrule

\multirow{12}{*}{{DeepSeek-v4-flash}}
& Zero-shot
& 75.8 & 43.7 & 77.0 & 70.8 & 84.6 & 91.2 & 73.9 \\
& Few-shot
& 75.0 & 42.2 & 71.0 & 66.1 & 86.4 & 90.3 & 71.8 \\
& SC
& 90.1 & 65.0 & 93.1 & 90.0 & 90.3 & 92.5 & 86.8 \\
& ToT
& 86.7 & 53.5 & 95.0 & 91.2 & 93.1 & 93.1 & 85.4 \\
& Experience RAG
& 72.5 & 51.4 & 74.1 & 68.4 & 82.7 & 84.6 & 72.3 \\
& ExpeL
& 79.7 & 54.2 & 80.4 & 74.1 & 87.6 & 95.0 & 78.5 \\
& Direct Skill
& 80.0 & 53.0 & 77.9 & 71.3 & 85.5 & 96.7 & 77.4 \\
& SkillOpt
& 82.2 & 55.3 & 82.0 & 79.6 & 92.7 & 95.1 & 81.2 \\
& MIPROv2
& 75.4 & 48.3 & 78.1 & 74.0 & 88.5 & 93.2 & 76.3 \\
& GEPA
& 74.0 & 50.8 & 78.9 & 77.2 & 91.4 & 93.1 & 77.6 \\
& {\cellcolor[rgb]{0.925,0.957,1}}{\Ours ($M=1$)}
& {\cellcolor[rgb]{0.925,0.957,1}}90.2
& {\cellcolor[rgb]{0.925,0.957,1}}49.4
& {\cellcolor[rgb]{0.925,0.957,1}}84.3
& {\cellcolor[rgb]{0.925,0.957,1}}77.0
& {\cellcolor[rgb]{0.925,0.957,1}}93.0
& {\cellcolor[rgb]{0.925,0.957,1}}97.2
& {\cellcolor[rgb]{0.925,0.957,1}}81.9 \\
& {\cellcolor[rgb]{0.925,0.957,1}}{\Ours ($M=10$)}
& {\cellcolor[rgb]{0.925,0.957,1}}\textbf{93.3}
& {\cellcolor[rgb]{0.925,0.957,1}}\textbf{91.5}
& {\cellcolor[rgb]{0.925,0.957,1}}\textbf{99.1}
& {\cellcolor[rgb]{0.925,0.957,1}}\textbf{99.0}
& {\cellcolor[rgb]{0.925,0.957,1}}\textbf{97.9}
& {\cellcolor[rgb]{0.925,0.957,1}}\textbf{97.8}
& {\cellcolor[rgb]{0.925,0.957,1}}\textbf{96.4} \\

\midrule

\multirow{12}{*}{{Qwen3.5-27B}}
& Zero-shot
& 82.5 & 65.6 & 51.5 & 50.5 & 48.3 & 59.4 & 59.6 \\
& Few-shot & 81.6 & 66.4 & 51.2 & 50.8 & 47.0 & 58.1 & 59.2 \\
& SC
& 85.0 & 62.5 & 70.4 & 75.2 & 60.1 & 60.6 & 69.0 \\
& ToT
& 83.5 & 65.8 & \textbf{85.6} & 80.3 & 69.8 & \textbf{91.4} & 79.4 \\
& Experience RAG & 76.2 & 73.2 & 51.3 & 51.2 & 46.4 & 54.5 & 58.8 \\
& ExpeL
& 86.6 & 75.2 & 55.1 & 54.4 & 53.8 & 64.0 & 64.9 \\
& Direct Skill & 81.2 & 69.7 & 51.3 & 53.1 & 52.9 & 62.6 & 61.8 \\
& SkillOpt
& 93.1 & 84.2 & 59.0 & 57.4 & 55.6 & 66.9 & 69.4 \\
& MIPROv2
& 85.1 & 75.2 & 52.8 & 52.4 & 50.3 & 61.7 & 62.9 \\
& GEPA
& 92.5 & 82.5 & 54.4 & 52.1 & 50.0 & 62.4 & 65.7 \\
& {\cellcolor[rgb]{0.925,0.957,1}}{\Ours ($M=1$)}
& {\cellcolor[rgb]{0.925,0.957,1}}89.0
& {\cellcolor[rgb]{0.925,0.957,1}}86.1
& {\cellcolor[rgb]{0.925,0.957,1}}61.2
& {\cellcolor[rgb]{0.925,0.957,1}}62.0
& {\cellcolor[rgb]{0.925,0.957,1}}60.3
& {\cellcolor[rgb]{0.925,0.957,1}}71.5
& {\cellcolor[rgb]{0.925,0.957,1}}71.7 \\
& {\cellcolor[rgb]{0.925,0.957,1}}{\Ours ($M=10$)}
& {\cellcolor[rgb]{0.925,0.957,1}}\textbf{95.2}
& {\cellcolor[rgb]{0.925,0.957,1}}\textbf{90.1}
& {\cellcolor[rgb]{0.925,0.957,1}}{85.3}
& {\cellcolor[rgb]{0.925,0.957,1}}\textbf{82.5}
& {\cellcolor[rgb]{0.925,0.957,1}}\textbf{72.6}
& {\cellcolor[rgb]{0.925,0.957,1}}{90.7}
& {\cellcolor[rgb]{0.925,0.957,1}}\textbf{86.1} \\

\bottomrule
\end{tabular}
}
\caption{Performance comparison across mathematical and logical reasoning benchmarks. All logical reasoning results are evaluated on the hard subsets of the corresponding benchmarks. Best results for each model and benchmark are shown in bold.
}
\label{tab:main_results}
\end{table*}

\paragraph{Evolution Operators.}
An \emph{evolution operator} is a structured transformation that proposes skill revisions and induces a particular \emph{inductive bias} over the skill search space.
Relying on a single revision strategy may restrict the range of explored transformations and lead the search toward suboptimal solutions.
We therefore maintain a portfolio of \emph{heterogeneous} operators that support complementary forms of skill evolution.

At step $t$, an operator $a_t\in\mathcal{A}$ is selected, followed by a corresponding parent set
\begin{equation}
\mathcal{P}_t
=
\operatorname{SelectParents}
\left(
\mathcal{S}_t,a_t
\right).
\end{equation}
The selected operator then defines a conditional proposal skill:
\begin{equation}
s'_t
\sim
p_{f_\theta}
\left(
\cdot
\mid
\mathcal{P}_t,
\mathcal{E}(\mathcal{D}_{\mathrm{ref}};\mathcal{P}_t),
a_t
\right),
\label{eq:skill_proposal}
\end{equation}
where 
\begin{equation}
\mathcal{E}\left(\mathcal{D}_{\mathrm{ref}};\mathcal{P}_t\right)=\bigcup_{s\in\mathcal{P}_t}\mathcal{E}\left(\mathcal{D}_{\mathrm{ref}};s\right)
\end{equation}
collects the verifier-labeled trajectories generated under the selected parent skills.
The operator $a_t$ specifies the \emph{proposal kernel}, thereby shaping the distribution over candidate revisions.

\begin{algorithm}[t]
\caption{\Ours: Diversity-Driven Skill Evolution}
\label{alg:ours}
\begin{algorithmic}[1]
\small

\REQUIRE Frozen model $f_\theta$, evolution set $\mathcal{D}_{\mathrm{evo}}$,
validation set $\mathcal{D}_{\mathrm{val}}$, verifier $r$, evolution budget $B$,
number of populations $K$, initial operators $\mathcal{A}_0$,
UCB coefficient $\beta$, operator-generation step $t_{\mathrm{new}}$,
number of new operators $N_{\mathrm{new}}$, maximum final skill-set size $M$

\ENSURE Final skill set $\mathcal{S}_{\mathrm{final}}$

\STATE $\mathcal{C}\gets\emptyset$

\FOR{$k=1,\ldots,K$ \textbf{in parallel}}

    \STATE Bootstrap
    $\mathcal{D}_{\mathrm{exp}}^{(k)},
    \mathcal{D}_{\mathrm{ref}}^{(k)}
    \subseteq\mathcal{D}_{\mathrm{evo}}$

    \STATE
    $\mathcal{S}_0^{(k)}
    \gets
    \textsc{InitializeSkill}_{f_\theta}
    (\mathcal{E}(\mathcal{D}_{\mathrm{exp}}^{(k)}))$

    \STATE $\mathcal{A}^{(k)}\gets\mathcal{A}_0$
    \STATE Initialize $\{N_a,\widehat{\mu}_a\}_{a\in\mathcal{A}^{(k)}}$
    \STATE $\mathcal{H}^{(k)}\gets\emptyset$

    \FOR{$t=1,\ldots,B$}

        \IF{some $a\in\mathcal{A}^{(k)}$ has not been tried}
            \STATE $a_t\gets$ an untried operator
        \ELSE
            \STATE
            $a_t
            \gets
            \arg\max_{a\in\mathcal{A}^{(k)}}
            \left[
            \widehat{\mu}_a+
            \beta\sqrt{\frac{\log t}{N_a}}
            \right]$
        \ENDIF

        \STATE
        $\mathcal{P}_t
        \gets
        \textsc{SelectParents}
        (\mathcal{S}_{t-1}^{(k)},a_t)$

        \STATE
        $s'_t
        \sim
        p_{f_\theta}
        \left(
        \cdot
        \mid
        \mathcal{P}_t,
        \mathcal{E}(\mathcal{D}_{\mathrm{ref}}^{(k)};\mathcal{P}_t),
        a_t
        \right)$

        \STATE Evaluate $s'_t$ on $\mathcal{D}_{\mathrm{ref}}^{(k)}$

        \STATE
        $R_t
        \gets
        U(s'_t)
        -
        \max_{s\in\mathcal{P}_t}U(s)$

        \STATE Update $N_{a_t}$ and $\widehat{\mu}_{a_t}$ using $R_t$

        \STATE Update $\mathcal{H}^{(k)}$ with the current proposal and outcome

        \STATE
        $\mathcal{S}_{t}^{(k)}
        \gets
        \textsc{UpdatePopulation}
        (\mathcal{S}_{t-1}^{(k)},s'_t)$

        \IF{$t=t_{\mathrm{new}}$}
            \STATE
            $\mathcal{A}^{(k)}_{\mathrm{new}}
            \gets
            \textsc{GenerateOperator}_{f_\theta}
            (\mathcal{H}^{(k)},N_{\mathrm{new}})$
            \STATE
            $\mathcal{A}^{(k)}
            \gets
            \mathcal{A}^{(k)}
            \cup
            \mathcal{A}^{(k)}_{\mathrm{new}}$
            \STATE Initialize new operators as untried
        \ENDIF

    \ENDFOR

    \STATE
    $\mathcal{C}
    \gets
    \mathcal{C}
    \cup
    \mathcal{S}_{B}^{(k)}$

\ENDFOR

\STATE Evaluate all $s\in\mathcal{C}$ on $\mathcal{D}_{\mathrm{val}}$

\STATE
$\mathcal{S}_{\mathrm{final}}
\gets
\textsc{JointSelect}
\left(
\{\mathcal{S}^{(k)}_B\}_{k=1}^{K},
\mathcal{D}_{\mathrm{val}},
M
\right)$

\RETURN $\mathcal{S}_{\mathrm{final}}$

\end{algorithmic}
\end{algorithm}

\paragraph{Adaptive Allocation with Upper Confidence Bounds.}

To adaptively allocate the evolution budget across operators, we balance exploitation and exploration using an upper-confidence-bound (UCB) score:
\begin{equation}
\operatorname{UCB}_a(t)
=
\widehat{\mu}_a(t)
+
\beta
\sqrt{
\frac{\log t}{N_a(t)}
},
\label{eq:operator_ucb}
\end{equation}
where $N_a(t)$ denotes the number of times operator $a$ has been selected before step $t$, and
\begin{equation}
\widehat{\mu}_a(t)
=
\frac{1}{N_a(t)}
\sum_{\tau<t:\,a_\tau=a}
R_\tau
\end{equation}
is its empirical mean parent-relative reward.
Here, $R_\tau \triangleq U(s'_\tau)-\max_{s\in\mathcal{P}_\tau}U(s)$ denotes the improvement of the proposed skill over its best-performing parent at step $\tau$, where $U(s)$ denotes the average correctness of skill $s$.
Each operator is applied at least once for initialization, after which we select
\begin{equation}
a_t
=
\arg\max_{a\in\mathcal{A}}
\operatorname{UCB}_a(t).
\end{equation}
The first term favors operators that have yielded larger improvements, while the second encourages exploration of less frequently evaluated operators.
This mechanism adaptively concentrates the finite evolution budget on empirically productive transformations while continuing to explore under-evaluated alternatives.

\paragraph{Adaptive Operator Generation.}
Let ${\mathcal{H}}_t$ denote the evolution history accumulated up to step $t$, including the selected operators, parent skills, proposed revisions, and their parent-relative reward.
At a predefined step $t_{\mathrm{new}}$, we use this history to generate $N_{\mathrm{new}}$ operators that target recurring failure patterns or transformations insufficiently addressed by the existing operator portfolio:
\begin{equation}
\mathcal{A}_{\mathrm{new}}
=
\operatorname{GenerateOperator}_{f_\theta}
\left(
{\mathcal{H}}_{t_{\mathrm{new}}},N_{\mathrm{new}}
\right).
\end{equation}
The newly generated operators are added to the operator pool as untried operators,
$\mathcal{A}\leftarrow\mathcal{A}\cup\mathcal{A}_{\mathrm{new}}$.
Subsequent proposals are allocated across both existing and newly generated operators using the same UCB rule.
This allows the operator portfolio itself to adapt in response to observed evolution dynamics.

\subsection{Final Skill Set Construction}
\label{sec:final_skill_set}

After independently evolving all $K$ populations, we evaluate their candidate skills on $\mathcal{D}_{\mathrm{val}}$.
The shared validation set is used only for final skill set construction and does not induce further within-population revisions.

\paragraph{Joint Skill Set Selection.}
We construct the final skill set jointly according to the validation performance of the candidate generation and ranking pipeline~\citep{xiong2025deliberate2,xiong2025deliberate}.

For a candidate skill set $\mathcal{S}$, each skill $s\in\mathcal{S}$ independently generates a candidate response
$o_s(x)=f_\theta(x;s)$.
Given the resulting candidate set
$\mathcal{O}_{\mathcal{S}}(x)=\{o_s(x)\}_{s\in\mathcal{S}}$,
the frozen model jointly evaluates the candidate solutions based on their
reasoning, answer consistency, and adherence to task constraints, and selects
the highest-ranked response:
\begin{equation}
\widehat{o}_{\mathcal{S}}(x)
=
\operatorname{SelectTop}_{f_\theta}
\left(
x,\mathcal{O}_{\mathcal{S}}(x)
\right).
\label{eq:skill_candidate_selection}
\end{equation}

The empirical validation utility of a skill set is
\begin{equation}
\widehat{V}_{\mathcal{D}_{\mathrm{val}}}
\left(
\mathcal{S}
\right)
=
\frac{1}{|\mathcal{D}_{\mathrm{val}}|}
\sum_{(x,y)\in\mathcal{D}_{\mathrm{val}}}
r\left(
x,
y,
\widehat{o}_{\mathcal{S}}(x)
\right).
\label{eq:validation_ranking_utility}
\end{equation}

Given a maximum final-set size $M$, we optimize the following objective
\begin{equation}
\max_{\mathcal{S}}
\widehat{V}_{\mathcal{D}_{\mathrm{val}}}
\left(
\mathcal{S}
\right),
\label{eq:joint_skill_selection}
\end{equation}
subject to
$\mathcal{S}\subseteq\bigcup_{k=1}^{K}\mathcal{S}_B^{(k)}$,
$1\leq|\mathcal{S}|\leq M$, and
$|\mathcal{S}\cap\mathcal{S}_B^{(k)}|\leq1$ for all $k\in[K]$.
In practice, we first retain the top few candidates from each population according to their individual performance on $\mathcal{D}_{\mathrm{val}}$, and then greedily construct the final skill set $\mathcal{S}_{\mathrm{final}}$ according to each candidate's marginal improvement to $\widehat{V}_{\mathcal{D}_{\mathrm{val}}}$.

\paragraph{Inference-Time Candidate Ranking.}
At test time, every skill in $\mathcal{S}_{\mathrm{final}}$ independently generates a candidate response.
The candidates are then ranked by $f_\theta$, and the highest-ranked candidate is returned as the final response:
\begin{equation}
\widehat{o}_{\mathcal{S}_{\mathrm{final}}}(x)
=
\operatorname{SelectTop}_{f_\theta}
\left(
x,
\left\{
f_\theta(x;s)
\right\}_{s\in\mathcal{S}_{\mathrm{final}}}
\right).
\label{eq:test_time_ranking}
\end{equation}
The final skill set is fixed using $\mathcal{D}_{\mathrm{val}}$, and the same ranking procedure is used throughout test-time evaluation.

\begin{table*}[t]
\centering
\resizebox{0.9\linewidth}{!}{
\begin{tabular}{l|cccc|cccccc|c}
\toprule
\textbf{Method}
& \textbf{Calls}
& \textbf{Input Tokens}
& \textbf{Output Tokens}
& \textbf{Cost}
& \textbf{HMMT}
& \textbf{Eq. Theories}
& \textbf{Sudoku}
& \textbf{Crypt.}
& \textbf{Calc.}
& \textbf{Futo.}
& \textbf{Avg.} \\
\midrule

GPT-5-nano (Zero-shot)
& 1.0 & 170.1 & 19,609.3 & \$0.008
& 64.3 & 56.5 & 71.7 & 20.1 & 30.4 & 70.8 & 52.3 \\

GPT-5-nano (Few-shot)
& 1.0 & 4,966.3 & 17,590.2 & \$0.007
& 63.3 & 59.5 & 77.0 & 25.5 & 26.2 & 69.1 & 53.4 \\

{\cellcolor[rgb]{0.925,0.957,1}}GPT-5-nano (\Ours)
& {\cellcolor[rgb]{0.925,0.957,1}} 10.9
& {\cellcolor[rgb]{0.925,0.957,1}} 33,811.2
& {\cellcolor[rgb]{0.925,0.957,1}} 216,834.2
& {\cellcolor[rgb]{0.925,0.957,1}} \$0.088
& {\cellcolor[rgb]{0.925,0.957,1}}\textbf{82.9}
& {\cellcolor[rgb]{0.925,0.957,1}}64.5
& {\cellcolor[rgb]{0.925,0.957,1}}\textbf{96.0}
& {\cellcolor[rgb]{0.925,0.957,1}}61.5
& {\cellcolor[rgb]{0.925,0.957,1}}\textbf{84.8}
& {\cellcolor[rgb]{0.925,0.957,1}}\textbf{99.2}
& {\cellcolor[rgb]{0.925,0.957,1}}\textbf{81.5} \\

\midrule

GPT-5 (Zero-shot)
& 1.0 & 170.1 & 15,388.5 & \$0.154
& 71.3
& 76.5
& 81.3
& 64.5
& 70.5
& 95.0
& 76.5 \\

GPT-5 (Few-shot)
& 1.0 & 4,966.3 & 14,675.0 & \$0.153
& 75.0
& \textbf{81.5}
& 89.0
& \textbf{66.0}
& 72.0
& 95.0
& 79.8 \\

\bottomrule
\end{tabular}
}
\caption{Performance and per-example inference cost of small-model skill evolution compared with large-model prompting. All logical reasoning tasks are evaluated on the hard subsets. For \Ours, the final skill-set size is capped at $M=10$.}
\label{tab:gpt5_comparison}
\end{table*}
\begin{table*}[t]
\centering
\resizebox{0.9\linewidth}{!}{
\begin{tabular}{l|l|cccccc|c}
\toprule
\textbf{Model}
& \textbf{Method}
& \textbf{HMMT}
& \textbf{Eq. Theories}
& \textbf{Sudoku}
& \textbf{Crypt.}
& \textbf{Calc.}
& \textbf{Futo.}
& \textbf{Avg.} \\
\midrule

\multirow{2}{*}{Qwen3.5-9B}
& Zero-shot
& 42.5 & 46.4 & 17.4 & 22.5 & 16.9 & 20.6 & 27.2 
\\

& {\cellcolor[rgb]{0.925,0.957,1}}\Ours (Skills from Qwen3.5-9B)
& {\cellcolor[rgb]{0.925,0.957,1}}\textbf{65.0}
& {\cellcolor[rgb]{0.925,0.957,1}}\textbf{68.7}
& {\cellcolor[rgb]{0.925,0.957,1}}\textbf{30.0}
& {\cellcolor[rgb]{0.925,0.957,1}}\textbf{45.4}
& {\cellcolor[rgb]{0.925,0.957,1}}\textbf{55.3}
& {\cellcolor[rgb]{0.925,0.957,1}}\textbf{75.1}
& {\cellcolor[rgb]{0.925,0.957,1}}\textbf{56.6}
\\

\midrule

\multirow{3}{*}{Qwen3.5-27B}
& Zero-shot
& 82.5 & 65.6 & 51.5 & 50.5 & 48.3 & 59.4 & 59.6 \\

& {\cellcolor[rgb]{0.925,0.957,1}}\Ours (Skills from Qwen3.5-9B)
& {\cellcolor[rgb]{0.925,0.957,1}}90.2
& {\cellcolor[rgb]{0.925,0.957,1}}85.3
& {\cellcolor[rgb]{0.925,0.957,1}}81.4
& {\cellcolor[rgb]{0.925,0.957,1}}79.2
& {\cellcolor[rgb]{0.925,0.957,1}}68.5
& {\cellcolor[rgb]{0.925,0.957,1}}82.6
& {\cellcolor[rgb]{0.925,0.957,1}}81.2
\\

& \Ours (Skills from Qwen3.5-27B) & \textbf{95.2} & \textbf{90.1} & \textbf{85.3} & \textbf{82.5} & \textbf{72.6} & \textbf{90.7} & \textbf{86.1} \\

\midrule

\multirow{3}{*}{DeepSeek-v4-flash}
& Zero-shot
& 75.8 & 43.7 & 77.0 & 70.8 & 84.6 & 91.2 & 73.9 \\

& {\cellcolor[rgb]{0.925,0.957,1}}\Ours (Skills from Qwen3.5-9B)
& {\cellcolor[rgb]{0.925,0.957,1}}87.3
& {\cellcolor[rgb]{0.925,0.957,1}}86.5
& {\cellcolor[rgb]{0.925,0.957,1}}92.3
& {\cellcolor[rgb]{0.925,0.957,1}}93.2
& {\cellcolor[rgb]{0.925,0.957,1}}91.8
& {\cellcolor[rgb]{0.925,0.957,1}}92.0
& {\cellcolor[rgb]{0.925,0.957,1}}90.5
\\

& \Ours (Skills from DeepSeek-v4-flash) & \textbf{93.3} & \textbf{91.5} & \textbf{99.1} & \textbf{99.0} & \textbf{97.9} & \textbf{97.8} & \textbf{96.4} \\

\bottomrule
\end{tabular}
}
\caption{Cross-model transferability of evolved skills across model scales and families. All logical reasoning results are evaluated on the hard subsets. For \Ours, the final skill-set size is capped at $M=10$.}
\label{tab:skill_transfer}
\end{table*}

\section{Evaluation}

We partition each dataset into evolution, validation, and test splits.
The optimizer has access to the inputs and labels in the evolution split, which are used for skill initialization and optimization.
The validation split is used only to evaluate candidate skills and construct the final skill set; its instances are never included in skill-generation or revision prompts.
The test split remains strictly held out and is used only for final evaluation.
We evaluate \Ours on six mathematical and logical reasoning tasks: HMMT~\citep{dekoninck2026matharena}, Equational Theories~\citep{bolan2025equational}, Sudoku, Cryptarithm, Calcudoku, and Futoshiki~\citep{liu2026synlogic}.
Experiments use GPT-5-nano~\citep{singh2025openai}, DeepSeek-v4-flash~\citep{xu2026deepseek}, and Qwen3.5~\citep{team2026qwen3} under standardized inference settings.
We compare against standard prompting and inference-time reasoning methods, including in-context learning (ICL)~\citep{brown2020language}, Self-Consistency (SC)~\citep{wangself}, and Tree-of-Thought (ToT)~\citep{yao2023tree}; experience- and memory-based methods, including Experience RAG~\citep{lewis2020retrieval} and ExpeL~\citep{zhao2024expel}; skill-learning methods, including Direct Skill Generation and SkillOpt~\citep{yang2026skillopt}; and prompt-optimization methods, including MIPROv2~\citep{miprov2} and GEPA~\citep{agrawal2025gepa}.
We further compare against parameter-based adaptation using SFT and GRPO~\citep{shao2024deepseekmath} on Qwen3-8B~\citep{yang2025qwen3}.
{Additional details on benchmarks and experimental settings are provided in the supplementary material}.

\paragraph{Main Results.}
\Cref{tab:main_results} summarizes the main results.
Across model families and both mathematical and logical reasoning tasks, \Ours consistently achieves strong performance without parameter updates.
\Ours also substantially outperforms experience- and memory-based methods and prompt optimization, showing the benefit of iterative skill evolution over one-shot skill construction and single-trajectory prompt optimization.
These results support the value of maintaining diverse evolution trajectories, heterogeneous revision strategies, and complementary skill hypotheses throughout the self-improvement process.

\begin{figure*}[t]
\centering
\begin{minipage}[t]{0.49\textwidth}
\centering
\begin{subfigure}[t]{0.49\linewidth}
    \centering
    \includegraphics[width=\linewidth]{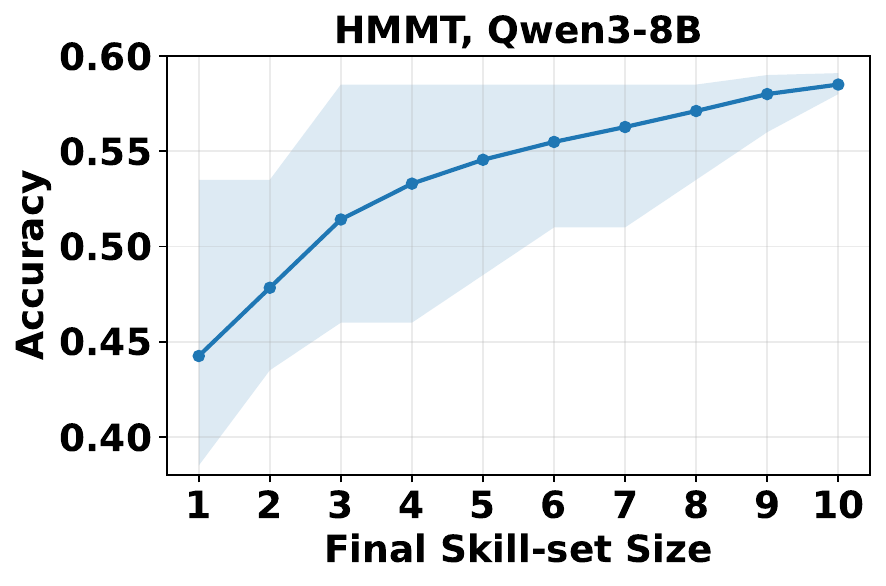}
    \caption{HMMT, Qwen3-8B}
\end{subfigure}
\hfill
\begin{subfigure}[t]{0.49\linewidth}
    \centering
    \includegraphics[width=\linewidth]{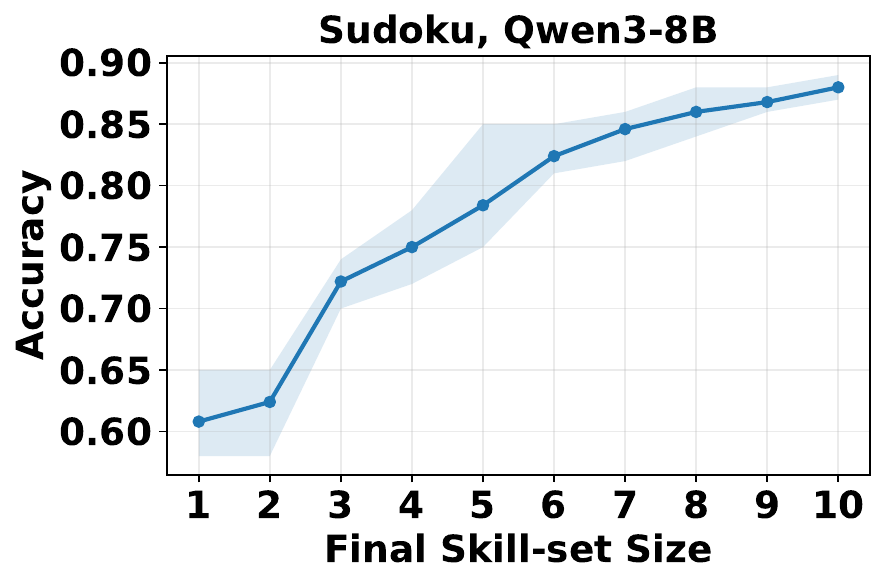}
    \caption{Sudoku, Qwen3-8B}
\end{subfigure}
\caption{Ablation of the final skill set size on HMMT and Sudoku using Qwen3-8B, with the number of skill populations fixed at $K=10$. Performance generally improves as more skills are included and gradually saturates.}
\label{fig:acc_vs_num_populations}
\end{minipage}
\hfill
\begin{minipage}[t]{0.49\textwidth}
\centering
\begin{subfigure}[t]{0.49\linewidth}
    \centering
    \includegraphics[width=\linewidth]{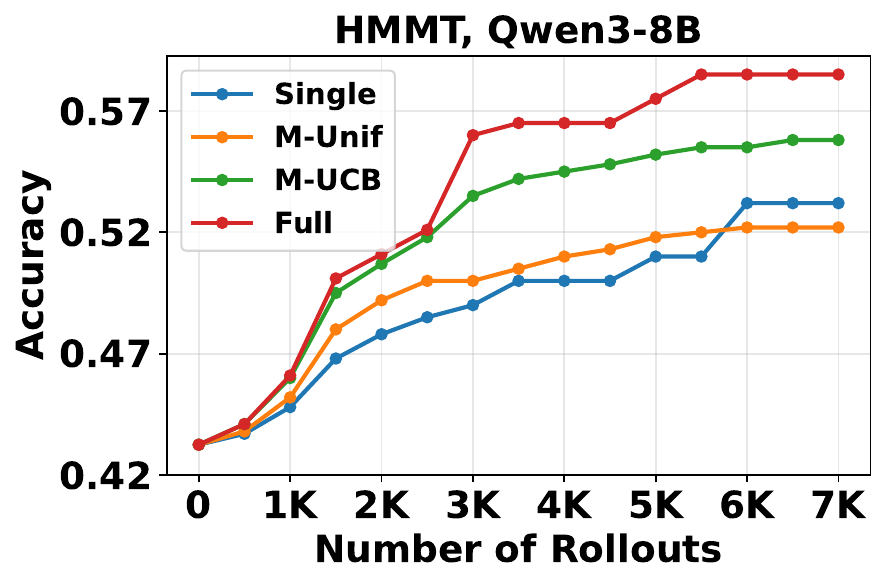}
    \caption{HMMT, Qwen3-8B}
\end{subfigure}
\hfill
\begin{subfigure}[t]{0.49\linewidth}
    \centering
    \includegraphics[width=\linewidth]{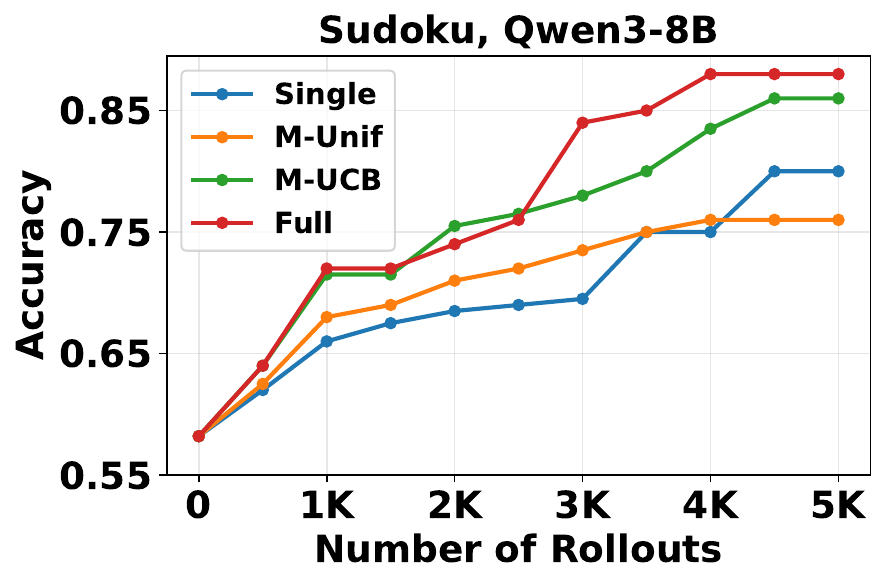}
    \caption{Sudoku, Qwen3-8B}
\end{subfigure}
\caption{Ablation of the skill-evolution strategy on HMMT and Sudoku using Qwen3-8B. We compare the best single evolution operator, multiple operators with uniform or UCB-based allocation, and the full method with operator generation.}
\label{fig:ablation_strategy}
\end{minipage}
\end{figure*}

\begin{table*}[t]
\centering
\resizebox{0.7\linewidth}{!}{
\begin{tabular}{l|cccccc|c}
\toprule
\textbf{Method}
& \textbf{HMMT}
& \textbf{Eq. Theories}
& \textbf{Sudoku}
& \textbf{Crypt.}
& \textbf{Calc.}
& \textbf{Futo.}
& \textbf{Avg.} \\
\midrule
Zero-shot
& 42.5 & 46.4 & 17.4 & 22.5 & 16.9 & 20.6 & 27.2\\

Best Single Skill
& 52.5 & 49.6 & \textbf{30.1} & 29.8 & 30.4 & 30.6 & 37.2 \\

Random $M$ Skills
& 62.4 & 56.1 & 28.7 & 40.6 & 42.0 & 65.9 & 49.3 \\

Top-$M$ Individual Skills
& \textbf{65.1} & 64.2 & 29.2 & 42.7 & 48.8 & 74.8 & 54.1 \\

{\cellcolor[rgb]{0.925,0.957,1}}Joint $M$-Skill Selection
& {\cellcolor[rgb]{0.925,0.957,1}}65.0
& {\cellcolor[rgb]{0.925,0.957,1}}\textbf{68.7}
& {\cellcolor[rgb]{0.925,0.957,1}}30.0
& {\cellcolor[rgb]{0.925,0.957,1}}\textbf{45.4}
& {\cellcolor[rgb]{0.925,0.957,1}}\textbf{55.3}
& {\cellcolor[rgb]{0.925,0.957,1}}\textbf{75.1}
& {\cellcolor[rgb]{0.925,0.957,1}}\textbf{56.6} \\

\bottomrule
\end{tabular}
}
\caption{Ablation of skill-set construction strategies using Qwen3.5-9B. Given $K=10$ populations and a maximum size of $M=10$, \textit{Random $M$ Skills} randomly selects one skill from each population, while \textit{Top-$M$ Individual Skills} selects the individually best-performing skills. \textit{Joint $M$-Skill Selection} denotes our proposed strategy. Logical reasoning uses hard subsets.
}
\label{tab:skill_set_construction}
\end{table*}

\paragraph{Small Models with Skills vs. Large Models.} 
We show that GPT-5-nano with \Ours achieves higher average performance than GPT-5 with ICL, while reducing inference cost by 42.5\% (\Cref{tab:gpt5_comparison}).
Skill evolution can substantially enhance the reasoning performance of a smaller model, enabling it to rival a stronger model under conventional prompting without parameter updates and at lower inference cost.

\paragraph{Optimization Efficiency.}
\Cref{fig:acc_rollout_curves} compares the optimization dynamics of \Ours, SFT, GRPO, and GEPA as a function of the number of rollouts on Qwen3-8B.
\Ours improves rapidly with additional experience, while SFT and GEPA plateau at substantially lower performance and GRPO improves more gradually despite requiring considerably more rollouts.
These results demonstrate a favorable performance--rollout trade-off, with \Ours attaining higher accuracy using substantially fewer rollouts than competing optimization approaches.

\paragraph{Cross-Model Skill Transfer.}
As shown in \Cref{tab:skill_transfer}, skills developed for Qwen3.5-9B transfer effectively to both Qwen3.5-27B, a larger model in the same family, and DeepSeek-v4-flash, a model from a different family.
These results show that evolved skills encode reusable task-solving knowledge that transfers across model scales and families, while target-specific evolution provides further gains by adapting skills to the behavior of the target model.

\paragraph{Ablation Analysis.}

\Cref{fig:acc_vs_num_populations} and \Cref{fig:ablation_strategy} validate the key design choices of \Ours.
Increasing the number of selected skills improves performance with diminishing returns, highlighting the benefit of complementary skill hypotheses, while heterogeneous operators outperform a single operator, UCB improves over uniform allocation, and adaptive operator generation yields further gains.
As shown in \Cref{tab:skill_set_construction}, jointly selecting complementary skills achieves the strongest overall performance, outperforming random or individually ranked skill selection.

\section{Related Work}

\textbf{Self-Improvement of Language Models.}
Language models can improve from experience through parameter updates such as supervised fine-tuning, reinforcement learning, and preference optimization~\citep{zelikman2022star,gulcehre2023reinforced,singh2023beyond,xiao2024cora,guo2025deepseek,zuo2026ttrl,zhang2026optimizing, bao2026rlearner}.
While effective, weight-space adaptation requires access to model parameters and training infrastructure and may incur substantial optimization cost.
Training-free alternatives instead improve model behavior through self-reflection and iterative refinement \citep{shinn2023reflexion,madaan2023self,10.1609/aaai.v39i28.35164}, in-context learning from interaction and feedback \citep{monea2025llmsincontextbanditreinforcement,xu2025provablylearninglanguagefeedback}, and natural-language reinforcement learning \citep{feng2025naturallanguagereinforcementlearning}.
These approaches primarily refine individual responses, retain interaction histories or reflections in context or memory, or study learning from richer language feedback.
In contrast, \Ours converts verifier-labeled experience into persistent task-level skills that explicitly encode reusable procedural knowledge, and optimizes multiple independent skill populations for reuse across future instances.

\noindent\textbf{Prompt Optimization and Evolution.}
Automatic prompt optimization \citep{zhou2022large,fernando2023promptbreeder,agarwal2024promptwizard,yang2024large} uses language models to search for effective instructions rather than relying on manual prompt design.
Evolutionary approaches extend this paradigm in different directions:
EvoPrompt~\citep{guo2024connecting} evolves prompt populations, Rainbow Teaming~\citep{samvelyan2024rainbow} generates adversarial prompts, and AlphaEvolve~\citep{novikov2025alphaevolve} applies evolutionary search to algorithm and code optimization, and Pathwise~\citep{gungordu2026pathwise} uses self-evolving LLMs for automated heuristic design.
More closely related to prompt optimization, GEPA~\citep{agrawal2025gepa} combines natural-language reflection with Pareto-based candidate selection for evolutionary prompt optimization, while MAPRO~\citep{zhang2026mapro} formulates multi-agent prompt optimization as structured probabilistic inference with feedback-driven prompt refinement.
\Ours differs in how diversity is used throughout optimization.
Rather than using population diversity primarily to search for strong individual artifacts, \Ours \emph{preserves diversity as an end-to-end design principle}, from independently bootstrapped evolution trajectories and heterogeneous transformations to the complementary skill set retained for inference.
This reduces dependence on any single evolution trajectory while preserving distinct skill hypotheses.

\noindent\textbf{Memory- and Skill-Based Adaptation.}
A growing body of work improves language-model agents by storing and reusing knowledge derived from prior interactions~\citep{zhong2024memorybank,zheng2024synapse,ouyang2025reasoningbank,zheng2025towards,zhang2025agentrouter,yang2026autoskill}.
Agent workflow memory \citep{wang2024agent} summarizes successful trajectories into reusable workflows, while other approaches induce reusable programmatic skills from experience \citep{wang2025inducing}.
These methods demonstrate that natural-language or structured external memory can support adaptation without weight updates.
Unlike approaches that primarily focus on how experience is stored and retrieved, \Ours focuses on \emph{how reusable knowledge should be searched, diversified, and selected under stochastic self-generated revisions}.
It therefore treats skill acquisition as an optimization problem over competing knowledge hypotheses rather than solely as memory construction or retrieval.

\section{Conclusion}

We introduced \Ours, a diversity-driven framework for self-improvement of frozen language models through natural-language skill evolution.
\Ours independently evolves multiple skill populations, adaptively allocates the evolution budget across heterogeneous operators, and jointly selects a complementary set of skills.
Across diverse reasoning tasks and model families, \Ours consistently improves performance, achieves a favorable performance--rollout trade-off, and produces skills that transfer across model scales and families.
These results highlight diversity-driven skill evolution as a promising approach for enabling frozen language models to improve from experience.
Future work may explore cross-task skill transfer and integration with long-term memory mechanisms to support more scalable and continual self-improvement.

\section*{Acknowledgments}
This work is supported in part by DARPA SciFy program, Award No.HR001125C0302, and CISCO Systems, Inc.

\bibliography{aaai2027}

\appendix
\newpage
\twocolumn[
\begin{center}
    {\LARGE \textbf{Supplementary Material}}
\end{center}
\vspace{2em}
]
\section{Additional Methodology Details}
\label{app:methodology_details}

This section provides further details on our methodology.
Because each skill population is evolved independently, we focus on a single population and \textbf{omit the population index $k$} throughout for notational simplicity.

\subsection{Seed Skill Initialization}
\label{app:seed_initialization}

For each population, we construct an initial seed set $\mathcal{S}_0=\left\{s_{\mathrm{minimal}},s_{\mathrm{const}},s_{\mathrm{veri}}\right\}$.
The three seeds provide complementary initialization biases.
The \emph{minimal seed} $s_{\mathrm{minimal}}$ contains only the task description and any mandatory output constraints.
The remaining two seeds are distilled from the verifier-labeled trajectories
collected on $\mathcal{D}_{\mathrm{exp}}$.
The \emph{solution-construction seed} $s_{\mathrm{const}}$ emphasizes
successful reasoning procedures, decompositions, representations, and
decision rules.
The \emph{diagnosis-and-verification seed} $s_{\mathrm{veri}}$ emphasizes
recurring failure modes, intermediate correctness checks, and final-answer
verification.
Together, these seeds provide a minimal starting hypothesis and two
complementary forms of experience-derived procedural knowledge for subsequent
evolution.

The prompt templates for summarization and skill initialization are shown below.

\begin{figure*}[!t]
\centering
\begin{tcolorbox}[
    colback=promptbg!100!white,
    sharp corners=south,
    boxrule=0.25mm,
    fonttitle=\bfseries,
    title={\textbf{\Ours's Summarization Prompt Template}},
    width=\textwidth,
    enhanced,
    drop shadow
]
\scriptsize
\setlength{\parskip}{0pt}

\#\#\# Task

\{\{\textcolor{blue}{task description}\}\}
\par\vspace{\baselineskip}
\#\#\# Verifier-Labeled Trajectories

\{\{\textcolor{blue}{questions, model responses, and verifier outcome labels}\}\}
\par\vspace{\baselineskip}
\#\#\# Instruction

Summarize the trajectories into compact evidence that can guide
subsequent skill development.
\par\vspace{\baselineskip}
Identify:

1. Recurring reasoning procedures, representations, or decision rules associated with successful responses;

2. Recurring failure patterns, including incorrect assumptions, invalid reasoning steps, calculation errors, incomplete solutions, and violated task constraints;

3. Verification strategies or intermediate checks that appear to distinguish successful responses from unsuccessful ones;

4. Output-format, parsing, execution, or timeout failures reported by the verifier;

5. Unresolved weaknesses for which the available trajectories do not yet suggest a reliable correction.
\par\vspace{\baselineskip}
Base the summary only on the provided trajectories and verifier outcomes.

Aggregate recurring patterns across instances rather than reproducing instance-specific solutions. 

Preserve concrete evidence when needed to make a pattern actionable, but omit unnecessary problem details, lengthy derivations, and complete answers. 

Do not propose a revised skill or solve the task.
\par\vspace{\baselineskip}
Organize the output using the following headings:

\#\# Successful Patterns

\#\# Failure Patterns

\#\# Verification and Constraint Checks

\#\# Unresolved Issues

Return only the summary.
\end{tcolorbox}
\end{figure*}

\begin{figure*}[!t]
\centering
\begin{tcolorbox}[
    colback=promptbg!100!white,
    sharp corners=south, 
    boxrule=0.25mm, 
    fonttitle=\bfseries, 
    title={\textbf{\Ours's Seed Skill Initialization Prompt Template}}, 
    width=\textwidth, 
    enhanced,
    drop shadow
]
\scriptsize
\setlength{\parskip}{0pt}
\setlength{\itemsep}{0pt}
\#\#\# Task

\{\{\textcolor{blue}{task description}\}\}
\par\vspace{\baselineskip}
\#\#\# Experience

\{\{\textcolor{blue}{summary of verifier-labeled successful and failed trajectories}\}\}
\par\vspace{\baselineskip}
\#\#\# Instruction

Distill reusable task knowledge from the observed experience into a
generalizable natural-language skill, with particular emphasis on \{\{\textcolor{blue}{focus}\}\}.

Extract principles and procedures that are useful beyond the observed
instances, while avoiding instance-specific solutions or unnecessary details.

Return only the skill.

\end{tcolorbox}
\end{figure*}

\subsection{Evolution Operator Portfolio}
\label{app:evolution_operators}

\Ours initializes each skill population with a portfolio of evolution operators designed to induce complementary search biases over the skill space:
\begin{equation}
\begin{aligned}
\mathcal{A}_0 = \{&
\text{Reflective Repair},
\text{Exploratory Revision},\\
&  \text{Compression}, \text{Recombination}
\}.
\end{aligned}
\end{equation}
The operators are defined as follows:
\begin{itemize}
    \item \textbf{Reflective Repair} (\textbf{1 parent}): Locally revises a skill to address systematic failures revealed by verifier-labeled trajectories while preserving useful existing guidance.


    \item \textbf{Exploratory Revision} (\textbf{1 parent}): Encourages exploration of a substantially different solution strategy, decomposition, or reasoning approach rather than incrementally modifying the parent skill.

    \item \textbf{Compression} (\textbf{1 parent}): Removes redundant, overly specific, or conflicting guidance and consolidates the remaining knowledge into a more concise and reusable skill.

    \item \textbf{Recombination} (\textbf{2 parents}): Synthesizes complementary strengths from two parent skills into a coherent child skill while resolving redundancy or conflicting guidance.
\end{itemize}
The prompt templates for each operator are shown below.

\begin{figure*}[!t]
\centering
\begin{tcolorbox}[
    colback=promptbg!100!white,
    sharp corners=south, 
    boxrule=0.25mm, 
    fonttitle=\bfseries, 
    title={\textbf{\Ours's Reflective Repair Prompt Template}}, 
    width=\textwidth, 
    enhanced,
    drop shadow
]
\scriptsize
\setlength{\parskip}{0pt}
\setlength{\itemsep}{0pt}
\#\#\# Task

\{\{\textcolor{blue}{task description}\}\}
\par\vspace{\baselineskip}
\#\#\# Parent Skill

\{\{\textcolor{blue}{parent skill}\}\}
\par\vspace{\baselineskip}
\#\#\# Experience

\{\{\textcolor{blue}{summary of verifier-labeled successful and failed trajectories}\}\}
\par\vspace{\baselineskip}
\#\#\# Instruction

Analyze the parent skill together with its successful and failed trajectories.
Identify systematic failure patterns and the aspects of the parent skill that contributed to successful behavior.
Minimally revise the skill to address the identified weaknesses while preserving useful existing guidance.

The revision should capture generalizable lessons rather than instance-specific solutions, avoid unnecessary changes or added complexity, and remain applicable to future task instances.

Return only the revised skill.

\end{tcolorbox}
\end{figure*}

\begin{figure*}[!t]
\centering
\begin{tcolorbox}[
    colback=promptbg!100!white,
    sharp corners=south, 
    boxrule=0.25mm, 
    fonttitle=\bfseries, 
    title={\textbf{\Ours's Exploratory Revision Prompt Template}}, 
    width=\textwidth, 
    enhanced,
    drop shadow
]
\scriptsize
\setlength{\parskip}{0pt}
\setlength{\itemsep}{0pt}
\#\#\# Task

\{\{\textcolor{blue}{task description}\}\}
\par\vspace{\baselineskip}
\#\#\# Parent Skill

\{\{\textcolor{blue}{parent skill}\}\}
\par\vspace{\baselineskip}
\#\#\# Experience

\{\{\textcolor{blue}{summary of verifier-labeled successful and failed trajectories}\}\}
\par\vspace{\baselineskip}
\#\#\# Instruction

Develop a substantially different solution strategy informed by the observed successes and failures.
Reconsider the assumptions, structure, and reasoning approach of the parent skill, and explore an alternative decomposition, representation, or problem-solving procedure.

The revised skill should remain generally applicable to future instances, preserve any indispensable task knowledge, and avoid merely making local edits to the parent skill.

Return only the revised skill.

\end{tcolorbox}
\end{figure*}

\begin{figure*}[!t]
\centering
\begin{tcolorbox}[
    colback=promptbg!100!white,
    sharp corners=south, 
    boxrule=0.25mm, 
    fonttitle=\bfseries, 
    title={\textbf{\Ours's Compression Prompt Template}}, 
    width=\textwidth, 
    enhanced,
    drop shadow
]
\scriptsize
\setlength{\parskip}{0pt}
\setlength{\itemsep}{0pt}
\#\#\# Task

\{\{\textcolor{blue}{task description}\}\}
\par\vspace{\baselineskip}
\#\#\# Parent Skill

\{\{\textcolor{blue}{parent skill}\}\}
\par\vspace{\baselineskip}
\#\#\# Experience

\{\{\textcolor{blue}{summary of verifier-labeled successful and failed trajectories}\}\}
\par\vspace{\baselineskip}
\#\#\# Instruction

Compress the parent skill into a more concise and coherent form.
Identify and remove redundant, overly specific, conflicting, or unnecessary guidance, while preserving the reasoning procedures, verification strategies, and constraints that contribute to successful behavior.

Abstract repeated or instance-specific guidance into generalizable principles whenever possible.
Do not sacrifice essential knowledge or introduce new strategies unless needed to resolve inconsistencies.

Return only the revised skill.

\end{tcolorbox}
\end{figure*}

\begin{figure*}[!t]
\centering
\begin{tcolorbox}[
    colback=promptbg!100!white,
    sharp corners=south, 
    boxrule=0.25mm, 
    fonttitle=\bfseries, 
    title={\textbf{\Ours's Recombination Prompt Template}}, 
    width=\textwidth, 
    enhanced,
    drop shadow
]
\scriptsize
\setlength{\parskip}{0pt}
\setlength{\itemsep}{0pt}
\#\#\# Task

\{\{\textcolor{blue}{task description}\}\}
\par\vspace{\baselineskip}
\#\#\# Parent Skills

\{\{\textcolor{blue}{parent skill 1}\}\}

\{\{\textcolor{blue}{parent skill 2}\}\}
\par\vspace{\baselineskip}
\#\#\# Experience

\{\{\textcolor{blue}{summary of verifier-labeled successful and failed trajectories}\}\}
\par\vspace{\baselineskip}
\#\#\# Instruction

Compare the parent skills together with their observed successes and failures.
Identify their complementary reasoning strategies, verification procedures, and specialized knowledge, and synthesize these strengths into a single coherent skill.

Resolve redundant, inconsistent, or conflicting guidance rather than simply concatenating the parent skills.
Preserve generalizable knowledge that contributes to successful behavior and avoid instance-specific solutions.

Return only the revised skill.

\end{tcolorbox}
\end{figure*}

\subsection{Parent Selection}
\label{app:parent_selection}

Parent selection is conditioned on the selected evolution operator and the
instance-level behavior of the current skill population.
For each skill $s\in\mathcal{S}_t$, we evaluate its correctness on the
population-specific reflection set $\mathcal{D}_{\mathrm{ref}}$, i.e.,
\begin{equation}
U(s)
=
\frac{1}{\left|\mathcal{D}_{\mathrm{ref}}\right|}
\sum_{(x,y)\in\mathcal{D}_{\mathrm{ref}}}
r\left(x,y,f_\theta(x;s)\right).
\label{eq:parent_skill_utility}
\end{equation}

The selected operator $a_t$ determines both the required number of parents
and the corresponding sampling criterion.
For each single-parent operator, we define an eligible parent set
$\mathcal{G}_{a_t}\subseteq\mathcal{S}_t$ and an operator-specific parent
score $G_{a_t}(s)$.
The parent is sampled according to
\begin{equation}
p_{\mathrm{single}}(s\mid a_t)
=
\frac{
    \mathbb{I}[s\in\mathcal{G}_{a_t}]
    \exp\left(G_{a_t}(s)/\tau_{\mathrm{p}}\right)
}{
    \displaystyle
    \sum_{\tilde{s}\in\mathcal{G}_{a_t}}
    \exp\left(G_{a_t}(\tilde{s})/\tau_{\mathrm{p}}\right)
},
\label{eq:single_parent_sampling}
\end{equation}
where $\tau_{\mathrm{p}}>0$ controls the sampling temperature.
Lower values place greater probability on skills with the highest
operator-specific scores, whereas larger values preserve greater parent
diversity.
For $\tau_{\mathrm{p}}=0$, we use greedy selection.

\texttt{Reflective Repair} is applied preferentially to high-performing skills so that
systematic errors can be corrected while preserving useful existing
guidance.
All current skills are eligible, and the parent score is its reflection-set
performance:
\begin{equation}
\mathcal{G}_{\mathrm{repair}}
=
\mathcal{S}_t,
\quad
G_{\mathrm{repair}}(s)
=
U(s).
\label{eq:repair_parent_score}
\end{equation}

\texttt{Exploratory Revision} targets lower-performing skills for which incremental
repair may be insufficient and a substantially different strategy may be
beneficial.
All skills are eligible, with
\begin{equation}
\mathcal{G}_{\mathrm{explore}}
=
\mathcal{S}_t,
\quad
G_{\mathrm{explore}}(s)
=
1-U(s).
\label{eq:exploratory_parent_score}
\end{equation}

\texttt{Compression} is applied only to skills whose token length exceeds a predefined
threshold.
Let $\ell(s)$ denote the token length of skill $s$, and let
$\ell_{\mathrm{comp}}$ denote the compression threshold.
The eligible set is
\begin{equation}
\mathcal{G}_{\mathrm{compress}}
=
\left\{
s\in\mathcal{S}_t:
\ell(s)>\ell_{\mathrm{comp}}
\right\}.
\label{eq:compression_eligible_set}
\end{equation}
Among the eligible skills, we favor those with higher performance:
\begin{equation}
G_{\mathrm{compress}}(s)
=
U(s),
\quad
s\in\mathcal{G}_{\mathrm{compress}}.
\label{eq:compression_parent_score}
\end{equation}
This encourages the operator to
shorten effective but overly long skills.
If no current skill exceeds $\ell_{\mathrm{comp}}$, Compression is considered
inapplicable at that evolution step.

For the \texttt{Recombination} operator, we first sample a strong parent
according to its reflection-set performance:
\begin{equation}
p_{\mathrm{dual}}(s_1=s)
=
\frac{
    \exp\left(U(s)/\tau_{\mathrm{p}}\right)
}{
    \displaystyle
    \sum_{\tilde{s}\in\mathcal{S}_t}
    \exp\left(U(\tilde{s})/\tau_{\mathrm{p}}\right)
},
\quad
s\in\mathcal{S}_t.
\label{eq:recombination_first_parent_sampling}
\end{equation}

Given the first parent $s_1$, we quantify the marginal coverage provided by
another skill $s$ as
\begin{equation}
\Delta(s\mid s_1)
=
\frac{1}{\left|\mathcal{D}_{\mathrm{ref}}\right|}
\sum_{(x,y)\in\mathcal{D}_{\mathrm{ref}}}
\left(1-c(s_1;x,y)\right)c(s;x,y).
\label{eq:marginal_parent_coverage}
\end{equation}
This score measures the fraction of reflection instances solved by $s$ but
not by $s_1$.

To favor second parents that are both individually effective and
complementary to the first parent, we define
\begin{equation}
Q(s\mid s_1)
=
\lambda U(s)
+
(1-\lambda)\Delta(s\mid s_1),
\label{eq:complementary_parent_score}
\end{equation}
where $\lambda\in[0,1]$ balances individual performance and marginal
coverage.
The second parent is then sampled according to
\begin{equation}
p_{\mathrm{dual}}(s_2=s\mid s_1)
=
\frac{
    \exp\left(Q(s\mid s_1)/\tau_{\mathrm{p}}\right)
}{
    \displaystyle
    \sum_{\tilde{s}\in\mathcal{S}_t\setminus\{s_1\}}
    \exp\left(Q(\tilde{s}\mid s_1)/\tau_{\mathrm{p}}\right)
},
\label{eq:recombination_second_parent_sampling}
\end{equation}
where $s\in\mathcal{S}_t\setminus\{s_1\}$.
The resulting parent set is
$\mathcal{P}_t=\{s_1\}$ for a single-parent operator and
$\mathcal{P}_t=\{s_1,s_2\}$ for \texttt{Recombination}.
Overall, \texttt{Reflective Repair} favors strong skills, \texttt{Exploratory Revision} favors
weaker skills, \texttt{Compression} favors strong skills that exceed the length
constraint, and \texttt{Recombination} pairs a strong first parent with a second
parent exhibiting complementary instance-level coverage.

\begin{figure*}[!t]
\centering
\begin{tcolorbox}[
    colback=promptbg!100!white,
    sharp corners=south,
    boxrule=0.25mm,
    fonttitle=\bfseries,
    title={\textbf{\Ours's Adaptive Operator Generation Prompt Template}},
    width=\textwidth,
    enhanced,
    drop shadow
]
\scriptsize
\setlength{\parskip}{0pt}
\setlength{\itemsep}{0pt}

\#\#\# Task

\{\{\textcolor{blue}{task description}\}\}

\par\vspace{\baselineskip}
\#\#\# Existing Evolution Operators

\{\{\textcolor{blue}{operator names, parent arities, and transformation instructions}\}\}

\par\vspace{\baselineskip}
\#\#\# Evolution History

\{\{\textcolor{blue}{summary of selected operators, parent skills, proposed revisions, and parent-relative rewards}\}\}

\par\vspace{\baselineskip}
\#\#\# Instruction

Analyze the evolution history and the existing operator portfolio.
Identify recurring failure patterns, unproductive revision behaviors, or
useful transformations that are not adequately addressed by the current
operators.

Propose exactly
\{\{\textcolor{blue}{$N_{\mathrm{new}}$}\}\}
new evolution operators that introduce distinct and practically useful
transformation strategies.
\par\vspace{\baselineskip}
For each proposed operator, provide:

1. The portfolio gap or recurring failure pattern it is intended to address;

2. Supporting evidence from the evolution history;

3. An explanation of how it differs from the most similar existing operator;

4. A concise operator name;

5. The required number of parent skills, which must be either 1 or 2;

6. A clear instruction specifying how the parent skill or skills should be
transformed.
\par\vspace{\baselineskip}
Each operator should be generalizable beyond the observed instances.
Avoid superficial renaming, unnecessary duplication, and minor variants of
existing operators. Do not include task-specific solutions or
instance-specific guidance.

Return exactly
\{\{\textcolor{blue}{$N_{\mathrm{new}}$}\}\}
operators as a JSON list using the following schema:

\begin{verbatim}
[
  {
    "targeted_gap": "<failure pattern or missing transformation addressed>",
    "supporting_evidence": "<evidence from the evolution history>",
    "distinction": "<how this differs from the closest existing operator>",
    "name": "<concise operator name>",
    "parent_arity": <1 or 2>,
    "instruction": "<transformation instruction>"
  }
]
\end{verbatim}

Return only the JSON list.

\end{tcolorbox}
\end{figure*}

\subsection{Adaptive Operator Generation}
\label{app:adaptive_operator_generation}

Although the initial operator portfolio provides several complementary
transformation strategies, a fixed portfolio may not adequately address all
failure patterns that emerge during evolution.
We therefore allow each skill population to generate additional operators
from its own accumulated evolution history.

For each population, we record the evolution history up to step $t$ as
\begin{equation}
\mathcal{H}_t
=
\left\{
\left(
a_\tau,
\mathcal{P}_\tau,
s'_\tau,
R_\tau
\right)
\right\}_{\tau=1}^{t},
\end{equation}
where $a_\tau$ is the selected operator,
$\mathcal{P}_\tau$ is the corresponding parent set,
$s'_\tau$ is the proposed revision, and
$R_\tau$ is its parent-relative reward.
We compute this reward on the population-specific reflection set:
\begin{equation}
R_\tau
=
U(s'_\tau)
-
\max_{s\in\mathcal{P}_\tau}
U(s).
\label{eq:operator_parent_relative_reward}
\end{equation}
For multi-parent operators, comparison against the strongest parent prevents
a proposal from receiving a positive reward merely by improving over a weaker
parent.

Because the complete evolution history may exceed the context
window, we convert it into a compact structured summary $\overline{\mathcal{H}}_t$.
For each existing operator $a$, the summary includes its number of
applications,
\begin{equation}
N_a(t)
=
\sum_{\tau=1}^{t}
\mathbb{I}[a_\tau=a],
\end{equation}
its empirical mean reward,
\begin{equation}
\widehat{\mu}_a(t)
=
\frac{1}{N_a(t)}
\sum_{\tau\leq t:\,a_\tau=a}
R_\tau,
\end{equation}
and its positive-improvement rate,
\begin{equation}
q_a^{+}(t)
=
\frac{1}{N_a(t)}
\sum_{\tau\leq t:\,a_\tau=a}
\mathbb{I}[R_\tau>0].
\end{equation}
The summary additionally contains representative successful and unsuccessful
revisions, recurring failure patterns that remain unresolved across the
population, and effective transformations observed in the history but not
explicitly represented by the current operator portfolio.

At a predefined step $t_{\mathrm{new}}$, we use the summarized evolution
history to generate $N_\mathrm{new}$ new operators in a single model call:
\begin{equation}
\mathcal{A}_{\mathrm{new}}
=
\operatorname{GenerateOperator}_{f_\theta}
\left(
\overline{\mathcal{H}}_{t_{\mathrm{new}}},
N_\mathrm{new}
\right).
\label{eq:adaptive_operator_generation}
\end{equation}
The model is instructed to propose $N_\mathrm{new}$ different operators that
address recurring failure patterns or useful transformations insufficiently
covered by the existing portfolio.
The generated operators are directly added to the
operator pool:
\begin{equation}
\mathcal{A}
\leftarrow
\mathcal{A}
\cup
\mathcal{A}_{\mathrm{new}}.
\end{equation}

Each generated proposal contains both diagnostic metadata and an executable
operator specification:
\begin{equation}
\widetilde{a}
=
\left(
g_a,
e_a,
d_a,
n_a,
h_a,
\pi_a
\right),
\end{equation}
where $g_a$ identifies the targeted portfolio gap,
$e_a$ summarizes supporting evidence from the evolution history,
$d_a$ explains how the proposal differs from its closest existing operator,
$n_a$ is the operator name,
$h_a$ is its required number of parents, and
$\pi_a$ is its transformation instruction.
The diagnostic fields $(g_a,e_a,d_a)$ encourage grounded and non-redundant
operator generation, whereas only the executable specification
$(n_a,h_a,\pi_a)$ is retained for subsequent evolution.

The adaptive operator generation prompt is shown below.

\begin{figure*}
\centering
\begin{tcolorbox}[
    colback=promptbg!100!white,
    sharp corners=south,
    boxrule=0.25mm,
    fonttitle=\bfseries,
    title={\textbf{\Ours's Candidate Ranking Prompt Template}},
    width=\textwidth,
    enhanced,
    drop shadow
]
\scriptsize
\setlength{\parskip}{0pt}
\setlength{\itemsep}{0pt}

\#\#\# Task

\{\{\textcolor{blue}{task description}\}\}
\par\vspace{\baselineskip}
\#\#\# Candidate Solutions

\#\#\#\# Candidate 1

\{\{\textcolor{blue}{Candidate 1's response}\}\}

\#\#\#\# Candidate 2

\{\{\textcolor{blue}{Candidate 2's response}\}\}

\hspace{0.5em}\(\vdots\)

\#\#\#\# Candidate $M$

\{\{\textcolor{blue}{Candidate $M$'s response}\}\}
\par\vspace{\baselineskip}
\#\#\# Instruction

Evaluate the candidate solutions and select the single candidate
most likely to correctly solve the question.

Consider:

1. Whether the reasoning is logically valid and sufficiently complete;

2. Whether intermediate conclusions and calculations are internally
   consistent;
   
3. Whether the final answer follows from the reasoning;

4. Whether the solution satisfies all task-specific constraints and output requirements.

Select exactly one of the provided candidates. 

Return only the index of the selected candidate.
\end{tcolorbox}
\end{figure*}

\subsection{Joint Skill Set Selection}
\label{app:joint_skill_selection}

After independently evolving all populations, we construct the final
skill set using the shared validation set $\mathcal{D}_{\mathrm{val}}$.
Rather than selecting skills solely according to their individual
performance, we seek a set whose predictions are
complementary under the complete inference-time
procedure.

We can define the empirical validation utility of a single skill $s$ as
\begin{equation}
\widehat{V}_{\mathcal{D}_{\mathrm{val}}}(\{s\})
=
\frac{1}{|\mathcal{D}_{\mathrm{val}}|}
\sum_{(x,y)\in\mathcal{D}_{\mathrm{val}}}
r\left(
x,
y,
\widehat{o}_{\{s\}}(x)
\right),
\label{eq:skill_set_validation_utility}
\end{equation}
where $\widehat{o}_{\{s\}}(x)=
f_\theta(x;s)$ for a singleton set.

To reduce the cost of joint selection, we first retain the top $L$ skills
from each evolved population according to their validation
performance:
\begin{equation}
\mathcal{C}^{(k)}
=
\operatorname{TopL}
\left(
\mathcal{S}_B^{(k)};
\widehat{V}_{\mathcal{D}_{\mathrm{val}}}(\{s\})
\right),
\end{equation}
and form the shortlisted candidate pool
\begin{equation}
\mathcal{C}
=
\bigcup_{k=1}^{K}\mathcal{C}^{(k)}.
\end{equation}
Let $\kappa(s)\in[K]$ denote the index of the population from which skill
$s$ originates.

We greedily construct the final skill set.
The first skill is selected according to its validation utility:
\begin{equation}
s_1^\star
=
\arg\max_{s\in\mathcal{C}}
\widehat{V}_{\mathcal{D}_{\mathrm{val}}}(\{s\}),
\quad
\mathcal{S}_1=\{s_1^\star\}.
\label{eq:first_skill_selection}
\end{equation}

At selection step $j\geq2$, the eligible candidate set is
\begin{equation}
\mathcal{C}_j
=
\left\{
s\in\mathcal{C}\setminus\mathcal{S}_{j-1}
:
\kappa(s)\notin
\left\{
\kappa(\widetilde{s})
:
\widetilde{s}\in\mathcal{S}_{j-1}
\right\}
\right\},
\label{eq:eligible_skill_candidates}
\end{equation}
which ensures that at most one skill is selected from each population.

For every eligible skill $s$, we compute its marginal contribution to the
validation utility of the complete skill set:
\begin{equation}
\Delta
\left(
s\mid\mathcal{S}_{j-1}
\right)
=
\widehat{V}_{\mathcal{D}_{\mathrm{val}}}
\left(
\mathcal{S}_{j-1}\cup\{s\}
\right)
-
\widehat{V}_{\mathcal{D}_{\mathrm{val}}}
\left(
\mathcal{S}_{j-1}
\right).
\label{eq:skill_marginal_utility}
\end{equation}
We then select
\begin{equation}
s_j^\star
=
\arg\max_{s\in\mathcal{C}_j}
\Delta
\left(
s\mid\mathcal{S}_{j-1}
\right),
\quad
\mathcal{S}_j
=
\mathcal{S}_{j-1}\cup\{s_j^\star\}.
\label{eq:greedy_joint_skill_selection}
\end{equation}

The procedure continues until the maximum skill-set size $M$ is reached or
no eligible skill provides a positive marginal improvement.
To reduce evaluation variance and computation, we generate and cache each
skill's response to every validation instance once.

\subsection{Inference-Time Candidate Ranking}
\label{app:candidate_ranking}

Given a question $x$ and a selected skill set
$\mathcal{S}$, each skill independently generates one
candidate response:
\begin{equation}
\mathcal{O}_{\mathcal{S}}(x)=\{o_s(x)\}_{s\in\mathcal{S}}
\end{equation}
For HMMT, we extract the final answer from each candidate response and select the answer using majority voting.
For the remaining datasets, we present all candidate responses (no lengthy reasoning content) to the same frozen model in a
single listwise ranking call:
\begin{equation}
\widehat{o}_{\mathcal{S}}(x)
=
\operatorname{SelectTop}_{f_\theta}
\left(
x, \mathcal{O}_{\mathcal{S}}(x)
\right).
\label{eq:listwise_candidate_ranking}
\end{equation}

The model jointly compares the candidate solutions based
on the validity and completeness of their reasoning, the consistency between
their reasoning and final answers, and their adherence to task-specific
constraints and output requirements.
On $\mathcal{D}_{\mathrm{val}}$, the external verifier evaluates the selected
candidate only after ranking to compute
$\widehat{V}_{\mathcal{D}_{\mathrm{val}}}(\mathcal{S})$.
The same ranking procedures are used throughout test-time
evaluation without access to test labels.
The ranking prompt template is shown below.

\section{Theoretical Analysis}
\label{app:theoretical_analysis}

This section provides theoretical motivation for two central components of
\Ours: adaptive allocation among heterogeneous evolution operators and
independent evolution followed by joint skill set selection.

\subsection{Adaptive Allocation of Evolution Operators}
\label{app:adaptive_operator_analysis}

An evolution operator determines not only how a parent skill is revised but
also which parents are eligible and how they are sampled.
We therefore treat an operator $a$ as a complete proposal mechanism consisting
of its parent arity, parent-selection distribution, and natural-language
transformation instruction.

Let $q_a\left(
\mathcal{P}\mid\mathcal{S}_t
\right)$
denote the distribution over parent sets induced by operator $a$ at step $t$.
After sampling
$\mathcal{P}_t\sim q_a(\cdot\mid\mathcal{S}_t)$, the frozen model proposes
\begin{equation}
s'_t
\sim
p_{f_\theta}
\left(
\cdot
\mid
\mathcal{P}_t,
\mathcal{E}(\mathcal{D}_{\mathrm{ref}};\mathcal{P}_t),
a
\right).
\label{eq:theory_operator_proposal}
\end{equation}
Thus, the observed performance of an operator reflects the complete
combination of parent selection and skill transformation.

\paragraph{Useful Proposal Mass under a Finite Budget.}

For a parent set $\mathcal{P}$ and improvement threshold $\epsilon>0$, define
the set of $\epsilon$-improving revisions as
\begin{equation}
\mathcal{I}_{\epsilon}(\mathcal{P})
=
\left\{
s':
U(s')
-
\max_{s\in\mathcal{P}}U(s)
\geq
\epsilon
\right\}.
\label{eq:theory_improvement_region}
\end{equation}
Conditioned on the evolution history $\mathcal{H}_{t-1}$, the useful proposal
mass of operator $a$ is
\begin{equation}
p_{a,t}(\epsilon)
=
\Pr\left(
s'_t\in
\mathcal{I}_{\epsilon}(\mathcal{P}_t)
\mid
a_t=a,\mathcal{H}_{t-1}
\right).
\label{eq:theory_useful_proposal_mass}
\end{equation}
where the probability is taken over both
$\mathcal{P}_t\sim q_a(\cdot\mid\mathcal{S}_t)$ and the model-generated
revision in \Cref{eq:theory_operator_proposal}.

This quantity measures how much probability the complete proposal mechanism
places on revisions that improve over their own starting points.
An operator can therefore be useful even when it does not produce the
highest-utility child on average, provided that it assigns substantial
probability to transformations not easily reached by the other operators.

Suppose the optimizer makes $T$ proposals, and define the conditional
improvement hazard
\begin{equation}
\overline{p}_t(\epsilon)
=
\Pr\left(
R_t\geq\epsilon
\mid
R_1<\epsilon,\ldots,R_{t-1}<\epsilon
\right).
\label{eq:conditional_improvement_hazard}
\end{equation}
By the chain rule,
\begin{equation}
\Pr\left(
\max_{1\leq t\leq T} R_t \geq \epsilon
\right)
=
1-
\prod_{t=1}^{T}
\left(
1-\overline{p}_t(\epsilon)
\right).
\label{eq:theory_hit_probability}
\end{equation}
Accordingly, finite-budget search improves when the selected proposal
mechanisms maintain high useful proposal mass over the sequence of search
states encountered during evolution.

A heterogeneous portfolio is valuable because the most productive proposal
mechanism may change as the population changes.
\texttt{Reflective Repair} may be effective once a strong but imperfect skill has been
identified, \texttt{Exploratory Revision} may be more useful when current strategies
share a common failure mode, and \texttt{Recombination} may become useful only after
complementary skills have emerged.
A single fixed proposal mechanism need not assign high useful mass in all of
these search states.

\paragraph{Parent-Relative Credit Assignment.}

The reward compares the proposed
child with its strongest parent.
For a multi-parent operator, this prevents a child from receiving positive
credit merely because it improves over a weak parent.
It also separates proposal quality from absolute parent quality.

For operator $a$, define its conditional expected parent-relative reward as
\begin{equation}
\mu_{a,t}
=
\mathbb{E}
\left[
R_t
\mid
a_t=a,\mathcal{H}_{t-1}
\right].
\label{eq:theory_operator_expected_reward}
\end{equation}
Both $\mu_{a,t}$ and $p_{a,t}(\epsilon)$ may vary with $t$ because the
population, eligible parents, unresolved failure modes, and available
evolution history change during optimization.
The operator-allocation problem is therefore generally non-stationary.

The UCB rule used by \Ours estimates the historical average reward of each
operator and balances this empirical productivity against uncertainty from
limited observations.
After every available operator has
received an initial trial, its score is
\begin{equation}
\operatorname{UCB}_a(t)
=
\widehat{\mu}_a(t)
+
\beta
\sqrt{
\frac{\log t}{N_a(t)}
}.
\label{eq:theory_ucb}
\end{equation}
The first term favors proposal mechanisms that have achieved larger
parent-relative improvements, while the second continues to test mechanisms
with fewer observations.

Since the reward distribution changes as the skill population evolves, we do
not invoke the standard regret guarantees for stationary stochastic bandits.
Instead, UCB serves as a practical finite-budget allocation rule that
concentrates proposals on empirically productive mechanisms without
prematurely eliminating underexplored alternatives.

Adaptive operator generation further expands the portfolio when the existing
mechanisms place insufficient proposal mass on transformations suggested by
the accumulated evolution history.

\subsection{Independent Populations and Joint Skill Set Selection}
\label{app:population_selection_analysis}

Natural-language skill evolution is stochastic and path-dependent.
Different bootstrapped examples, initial skills, sampled parents, and
model-generated revisions can lead to different regions of the skill space.
Independent populations improve the probability of discovering useful
candidate skills and can produce skills with complementary instance-level
success patterns.

\paragraph{Candidate-Pool Search Coverage.}

For a skill $s$, define its population risk on the task distribution
$\mathcal{T}$ as
\begin{equation}
\mathcal{R}(s)
=
\mathbb{E}_{(x,y)\sim\mathcal{T}}
\left[
1-r\left(x,y,f_\theta(x;s)\right)
\right].
\label{eq:theory_skill_risk}
\end{equation}

Let
\begin{equation}
G_k(\epsilon)
=
\left\{
\min_{s\in\mathcal{S}_B^{(k)}}
\mathcal{R}(s)
\leq
\epsilon
\right\}.
\end{equation}
denote the event that evolved population $k$ contains at least one skill with
risk at most $\epsilon$.

Under the idealized assumptions that the events
${G_k(\epsilon)}_{k=1}^{K}$ are independent and that
$\Pr(G_k(\epsilon))=q_\epsilon$ for every population,
\begin{equation}
\Pr\left(
\bigcup_{k=1}^{K}G_k(\epsilon)
\right)
=
1-
\left(
1-q_\epsilon
\right)^K.
\label{eq:theory_search_coverage}
\end{equation}
Thus, independent evolution increases the probability that the overall
candidate pool contains at least one low-risk skill, with diminishing returns
as the number of populations grows.

The populations are not perfectly independent in practice.
Nevertheless, independently bootstrapped experience, population-specific
reflection subsets, stochastic revisions, and separate evolution histories
encourage lower dependence across runs.
A discovered skill must still survive validation shortlisting, contribute to
the selected skill set, and be successfully aggregated at inference time.

\paragraph{Instance-Level Complementarity.}

For an instance $(x,y)$, define the set of skills producing correct candidates
as
\begin{equation}
C_{\mathcal{S}}(x,y)
=
\left\{
s\in\mathcal{S}:
r\left(x,y,o_s(x)\right)=1
\right\}.
\label{eq:theory_correct_candidate_set}
\end{equation}
The oracle coverage of $\mathcal{S}$ is
\begin{equation}
A_{\mathrm{oracle}}(\mathcal{S})
=
\Pr_{(x,y)\sim\mathcal{T}}
\left(
C_{\mathcal{S}}(x,y)\neq\varnothing
\right).
\label{eq:theory_oracle_coverage}
\end{equation}
This is the accuracy that would be obtained by an oracle capable of selecting
a correct candidate whenever at least one exists.

For a new skill $s\notin\mathcal{S}$, its marginal oracle coverage is
\begin{equation}
\begin{split}
\Delta_{\mathrm{cov}}
\left(
s\mid\mathcal{S}
\right)
=
\Pr_{(x,y)\sim\mathcal{T}}
\Big(
&
r\left(x,y,o_s(x)\right)=1,\\
&
C_{\mathcal{S}}(x,y)=\varnothing
\Big).
\end{split}
\label{eq:theory_marginal_coverage}
\end{equation}
Therefore,
\begin{equation}
A_{\mathrm{oracle}}
\left(
\mathcal{S}\cup\{s\}
\right)
-
A_{\mathrm{oracle}}(\mathcal{S})
=
\Delta_{\mathrm{cov}}
\left(
s\mid\mathcal{S}
\right).
\label{eq:theory_oracle_marginal_gain}
\end{equation}
A skill contributes to oracle coverage only on instances not already solved by
the current set.
Consequently, a skill with lower standalone accuracy may be more useful than a
higher-accuracy skill when its correct predictions occur on different
instances.
This motivates selecting skills according to their marginal contribution
rather than solely according to individual validation performance.

\paragraph{Candidate Coverage and Selection Regret.}

Let
$
\widehat{o}_{\mathcal{S}}(x)
$
denote the output returned by the aggregation
procedure applied to the candidates generated by $\mathcal{S}$.
Its realized accuracy is
\begin{equation}
A_{\mathrm{agg}}(\mathcal{S})
=
\Pr_{(x,y)\sim\mathcal{T}}
\left[
r\left(
x,y,\widehat{o}_{\mathcal{S}}(x)
\right)=1
\right].
\label{eq:theory_aggregation_accuracy}
\end{equation}

Because the aggregation procedure selects from the candidate outputs,
a correct final answer is possible only when at least one candidate is
correct.
We can therefore decompose aggregation accuracy as
\begin{equation}
A_{\mathrm{agg}}(\mathcal{S})
=
A_{\mathrm{oracle}}(\mathcal{S})
-
L_{\mathrm{sel}}(\mathcal{S}).
\label{eq:theory_selection_decomposition}
\end{equation}
where
\begin{equation}
\begin{split}
L_{\mathrm{sel}}(\mathcal{S})
=
\Pr_{(x,y)\sim\mathcal{T}}
\Big(
&
C_{\mathcal{S}}(x,y)\neq\varnothing,\\
&
r\left(
x,y,\widehat{o}_{\mathcal{S}}(x)
\right)=0
\Big).
\end{split}
\label{eq:theory_selection_regret}
\end{equation}
is the selection regret: the probability that at least one correct candidate
is available but the aggregation procedure returns an incorrect one.

Combining
\Cref{eq:theory_oracle_marginal_gain,eq:theory_selection_decomposition}, the
realized marginal gain from adding skill $s$ is
\begin{equation}
\begin{split}
&
A_{\mathrm{agg}}
\left(
\mathcal{S}\cup\{s\}
\right)
-
A_{\mathrm{agg}}(\mathcal{S})
\\
&\qquad =
\Delta_{\mathrm{cov}}
\left(
s\mid\mathcal{S}
\right)
-
\Delta_{\mathrm{sel}}
\left(
s\mid\mathcal{S}
\right).
\end{split}
\label{eq:theory_realized_marginal_gain}
\end{equation}
where
\begin{equation}
\Delta_{\mathrm{sel}}
\left(
s\mid\mathcal{S}
\right)
\coloneqq
L_{\mathrm{sel}}
\left(
\mathcal{S}\cup\{s\}
\right)
-
L_{\mathrm{sel}}(\mathcal{S}).
\label{eq:theory_marginal_selection_loss}
\end{equation}
Adding a skill may increase coverage while also making selection more
difficult.
Accordingly, test accuracy need not improve monotonically with skill-set size.
A useful skill is one whose additional coverage exceeds any additional
selection regret it introduces.

The validation objective used by \Ours directly estimates
$A_{\mathrm{agg}}(\mathcal{S})$ under the complete
aggregation pipeline.
Its greedy marginal-selection criterion therefore estimates the realized
quantity in \Cref{eq:theory_realized_marginal_gain}.
Stopping when no remaining candidate yields positive validation improvement
avoids adding skills whose incremental coverage does not compensate for their
effect on candidate selection.

Overall, independent populations improve the coverage of the evolved
candidate pool, while joint validation-based selection converts this search
diversity into a compact set of instance-level complementary skills.
The final benefit depends jointly on candidate competence, diversity in their
error patterns, and the ability of the aggregation procedure to maintain low
selection regret.

\begin{figure*}[t]
\centering
\begin{tcolorbox}[
    colback=promptbg!100!white,
    sharp corners=south,
    boxrule=0.25mm,
    fonttitle=\bfseries,
    title={\textbf{HMMT}},
    width=\textwidth,
    enhanced,
    drop shadow
]
\scriptsize
\setlength{\parskip}{0pt}
\setlength{\itemsep}{0pt}
\textbf{Question:} Mark writes the expression $\sqrt{\underline{a b c d}}$ on the board, where $\underline{a b c d}$ is a four-digit number and $a \neq 0$. Derek, a toddler, decides to move the $a$, changing Mark's expression to $a \sqrt{\underline{b c d}}$. Surprisingly, these two expressions are equal. Compute the only possible four-digit number $\underline{a b c d}$.

\textbf{Answer:} 3375
\par\vspace{\baselineskip}
\textbf{Question:} Trapezoid $A B C D$, with $A B \| C D$, has side lengths $A B=11, B C=8, C D=19$, and $D A=4$. Compute the area of the convex quadrilateral whose vertices are the circumcenters of $\triangle A B C, \triangle B C D$, $\triangle C D A$, and $\triangle D A B$.

\textbf{Answer:} $9\sqrt{15}$
\end{tcolorbox}
\end{figure*}

\begin{figure*}[t]
\centering
\begin{tcolorbox}[
    colback=promptbg!100!white,
    sharp corners=south,
    boxrule=0.25mm,
    fonttitle=\bfseries,
    title={\textbf{Equational Theories}},
    width=\textwidth,
    enhanced,
    drop shadow
]
\scriptsize
\setlength{\parskip}{0pt}
\setlength{\itemsep}{0pt}
\textbf{[Normal Set]}

\textbf{Question:}
You are a mathematician specializing in equational theories of magmas.

Your task is to determine whether Equation 1 (x = x * (x * ((y * z) * z))) implies Equation 2 (x = (x * x) * (y * (z * y))) over all magmas.

Output format (use exact headers without any additional text or formatting):

VERDICT: must be exactly TRUE or FALSE (in the same line).

REASONING: must be non-empty.

PROOF: required if VERDICT is TRUE, empty otherwise.

COUNTEREXAMPLE: required if VERDICT is FALSE, empty otherwise.

\textbf{Answer:} False
\par\vspace{\baselineskip}
\textbf{[Hard Set]}

\textbf{Question:}
You are a mathematician specializing in equational theories of magmas.

Your task is to determine whether Equation 1 (x = ((y * y) * z) * x) implies Equation 2 (x = x * (((y * x) * x) * x)) over all magmas.

Output format (use exact headers without any additional text or formatting):

VERDICT: must be exactly TRUE or FALSE (in the same line).

REASONING: must be non-empty.

PROOF: required if VERDICT is TRUE, empty otherwise.

COUNTEREXAMPLE: required if VERDICT is FALSE, empty otherwise.

\textbf{Answer:} True
\end{tcolorbox}
\end{figure*}

\begin{figure*}[t]
\centering
\begin{tcolorbox}[
    colback=promptbg!100!white,
    sharp corners=south,
    boxrule=0.25mm,
    fonttitle=\bfseries,
    title={\textbf{Sudoku}},
    width=\textwidth,
    enhanced,
    drop shadow
]
\scriptsize
\setlength{\parskip}{0pt}
\setlength{\itemsep}{0pt}
\textbf{[Normal Set]}

\textbf{Question:} This is a standard 9x9 Sudoku, where X needs to be filled with digits 1-9:

X65XX821X

7XX6X1X54

8X1X45673

X42X7X19X

9X8XX2365

1X6389XX2

539X6XX21

6X4923XX7

XX7154936

Please complete this Sudoku.

Sudoku rules: Fill in digits 1-9 so that each digit appears exactly once in each row, column, and 3x3 sub-grid.

Please provide your answer at the end using a Python markdown code block, represented as a tuple, for example: 

\texttt{\char96}\texttt{\char96}\texttt{\char96}python

((1,2,3,4,5,6,7,8,9),(4,5,6,7,8,9,1,2,3),...)

\texttt{\char96}\texttt{\char96}\texttt{\char96}

\textbf{Answer:} ((4, 6, 5, 7, 3, 8, 2, 1, 9), (7, 2, 3, 6, 9, 1, 8, 5, 4), (8, 9, 1, 2, 4, 5, 6, 7, 3), (3, 4, 2, 5, 7, 6, 1, 9, 8), (9, 7, 8, 4, 1, 2, 3, 6, 5), (1, 5, 6, 3, 8, 9, 7, 4, 2), (5, 3, 9, 8, 6, 7, 4, 2, 1), (6, 1, 4, 9, 2, 3, 5, 8, 7), (2, 8, 7, 1, 5, 4, 9, 3, 6))
\par\vspace{\baselineskip}
\textbf{[Hard Set]}

\textbf{Question:} Try to solve this Sudoku, where X represents unknown digits:

XX15X4X7X

X48X9XXX5

5XXXXXX8X

XX56XXXXX

XXXXX37X8

XX2XX14XX

XX643XX2X

XXXXXXXX9

X7XX1583X

Please fill in all the Xs.

The rules of Sudoku are simple: each row, column, and 3x3 box must contain the numbers 1-9 without repetition.

Please provide your answer at the end using a Python markdown code block, represented as a tuple, for example: 

\texttt{\char96}\texttt{\char96}\texttt{\char96}python

((1,2,3,4,5,6,7,8,9),(4,5,6,7,8,9,1,2,3),...)

\texttt{\char96}\texttt{\char96}\texttt{\char96}

\textbf{Answer:} ((3, 6, 1, 5, 8, 4, 9, 7, 2), (2, 4, 8, 3, 9, 7, 6, 1, 5), (5, 9, 7, 1, 6, 2, 3, 8, 4), (7, 3, 5, 6, 4, 8, 2, 9, 1), (6, 1, 4, 9, 2, 3, 7, 5, 8), (9, 8, 2, 7, 5, 1, 4, 6, 3), (8, 5, 6, 4, 3, 9, 1, 2, 7), (1, 2, 3, 8, 7, 6, 5, 4, 9), (4, 7, 9, 2, 1, 5, 8, 3, 6))
\end{tcolorbox}
\end{figure*}

\begin{figure*}[t]
\centering
\begin{tcolorbox}[
    colback=promptbg!100!white,
    sharp corners=south,
    boxrule=0.25mm,
    fonttitle=\bfseries,
    title={\textbf{Cryptarithm}},
    width=\textwidth,
    enhanced,
    drop shadow
]
\scriptsize
\setlength{\parskip}{0pt}
\setlength{\itemsep}{0pt}
\textbf{[Normal Set]}

\textbf{Question:} In this alphametic puzzle: CTYYR + CRTC + RTT = CRYCT (where CTYYR is 5-digit number, CRTC is 4-digit number, RTT is 3-digit number, CRYCT is 5-digit number), each letter represents a distinct digit from 0-9. Determine the digits that make the equation valid. Please end your response in the last line with the following format: The answer is \$YOUR\_ANSWER. \$YOUR\_ANSWER should be the equation with letters replaced by digits.

\textbf{Answer:} 16889 + 1961 + 966 = 19816
\par\vspace{\baselineskip}
\textbf{[Hard Set]}

\textbf{Question:} In this verbal arithmetic problem: DZSQL * DFQD - DSZ - LFDQ = FQDLQDFF (where DZSQL is a 5-digit number, DFQD is a 4-digit number, DSZ is a 3-digit number, LFDQ is a 4-digit number, FQDLQDFF is a 8-digit number), substitute each letter with a unique digit to make the equation true. Find the correct digit assignment. Please end your response in the last line with the following format: The answer is \$YOUR\_ANSWER. \$YOUR\_ANSWER should be the equation with letters replaced by digits.

\textbf{Answer:} 20418 * 2512 - 240 - 8521 = 51281255
\end{tcolorbox}
\end{figure*}

\section{Dataset Statistics}

In this section, we provide statistics for all benchmark datasets used in our study.
We evaluate \Ours on six mathematical and logical reasoning tasks: HMMT~\citep{dekoninck2026matharena}, Equational Theories~\citep{bolan2025equational}, Sudoku, Cryptarithm, Calcudoku, and Futoshiki~\citep{liu2026synlogic}.

HMMT is a prestigious high-school mathematics competition featuring challenging problems in algebra, geometry, combinatorics, and number theory. 
Because the official HMMT datasets do not include a training split, we construct one from NuminaMath-1.5~\citep{numina_math_datasets} by retaining valid, non-synthetic, short-answer competition problems from the \texttt{olympiads} and \texttt{amc\_aime} sources. 
We remove malformed and duplicate examples and exclude any problem with normalized overlap with the HMMT evaluation set, yielding 1,200 training problems. 
For evaluation, we combine the MathArena HMMT February 2025 and November 2025 datasets~\citep{dekoninck2026matharena}. 

Equational Theories is a mathematical reasoning benchmark based on the Equational Theories Project~\citep{bolan2025equational} and the SAIR Mathematics Distillation Challenge\footnote{https://competition.sair.foundation/competitions/mathematics-distillation-challenge-equational-theories-stage1/overview}. 
Each problem presents two identities over a magma and asks whether the first equation logically implies the second over all possible magmas.
We construct the training set by merging the publicly available subsets from the \texttt{SAIRfoundation/equational-theories-selec} \texttt{ted-problems} dataset and removing all examples that appear in the evaluation benchmark, resulting in 1,400 training problems. 
For evaluation, we use 400 problems from the \texttt{SAIRfoundation/equational-theories-bench} \texttt{mark} dataset, comprising 200 normal and 200 hard problems.

Sudoku, Cryptarithm, Calcudoku, and Futoshiki are synthetic logical reasoning tasks drawn from SynLogic \citep{liu2026synlogic}. 
For each task and difficulty setting, we use 1200 training instances and 200 test instances. 
We construct normal and hard subsets using task-specific generation parameters. 
In the normal setting, Sudoku uses difficulty level 3; Cryptarithm uses four letters, two operators, and operator level 2; Calcudoku uses a $5\times5$ grid; and Futoshiki uses a $5\times5$ grid with 10 inequality signs and 5 prefilled cells. 
In the hard setting, Sudoku uses difficulty level 4; Cryptarithm uses six letters, three operators, and operator level 3; Calcudoku uses a $6\times6$ grid; and Futoshiki uses a $6\times6$ grid with 14 inequality signs and 7 prefilled cells.

We evaluate Qwen3-8B on the normal subsets and all other models on the hard subsets. 
HMMT contains a single evaluation set, which is used for all models. 
Representative examples from each dataset are provided below.

\paragraph{Verifiers.}
We implement dataset-specific verifiers that reflect the structure and validity conditions of each task. 
For HMMT, the verifier extracts and normalizes the final answer, then checks it against the reference using symbolic equivalence.
For Equational Theories, the response is parsed into the required verdict, reasoning, proof, and counterexample fields. 
The normalized verdict is compared with the benchmark label. 
For the SynLogic tasks, Sudoku, Cryptarithm, Calcudoku, and Futoshiki, we use task-specific parsers and constraint checks. 
Each verifier extracts the proposed solution, validates the required output format, and checks whether the solution satisfies the underlying task constraints.

\section{Implementation Details}
\label{app:implementation_details}

For each dataset, we allocate $1{,}000$ training examples to the evolution set $\mathcal{D}_{\mathrm{evo}}$ and use the remaining as the validation set $\mathcal{D}_{\mathrm{val}}$. 
We maintain $K=10$ independent populations. 
For each population $k$, we independently bootstrap (with replacement) an experience subset $\mathcal{D}_{\mathrm{exp}}^{(k)}$ and a reflection subset $\mathcal{D}_{\mathrm{ref}}^{(k)}$ from $\mathcal{D}_{\mathrm{evo}}$, with$\frac{|\mathcal{D}_{\mathrm{exp}}^{(k)}|}{|\mathcal{D}_{\mathrm{evo}}|}=0.05$, $\frac{|\mathcal{D}_{\mathrm{ref}}^{(k)}|}{|\mathcal{D}_{\mathrm{evo}}|}=0.10$.

Each population is initialized with $|\mathcal{S}_0^{(k)}|=3$ seed skills and evolved for a budget of $B=10$ steps.
At each evolution step, the proposed child skill is added to the current population without removing its parent skills.
Operators for which no eligible parents are available are excluded from UCB selection at that step.
We set the UCB exploration coefficient to $\beta=0.3$ and the parent-sampling temperature to $\tau_{\mathrm{p}}=0.6$. 
Adaptive operator generation is performed at step $t_{\mathrm{new}}=8$, at which point $N_\mathrm{new}=1$ new operator is generated.
The Compression operator is applicable only to skills longer than $\ell_{\mathrm{comp}}=4{,}096$ tokens. 
For Recombination, we set the trade-off between individual performance and complementary coverage to $\lambda=0.5$.
We use one rollout per skill--instance pair when evaluating skills on $\mathcal{D}_{\mathrm{ref}}$ and $\mathcal{D}_{\mathrm{val}}$. 
Verifier-labeled trajectories are partitioned into chunks that fit within the model context window; the chunks are summarized independently and then merged into a final summary.
To construct the final skill set, we retain the top $L=3$ skills from each population according to their individual validation performance and jointly select a set of at most $M=10$ skills. 
Greedy selection terminates either when $M=10$ skills have been selected or when the largest remaining marginal improvement on the validation set is non-positive.

At inference time, models with native thinking or reasoning modes use their default reasoning configurations.
For models without such default configurations, we use a sampling temperature of $0.6$, $\texttt{top\_p}=0.95$, and a maximum output length of $32{,}768$ tokens. 
Unless otherwise stated, the same decoding configuration is used for skill initialization, evolution, summarization, operator generation, and inference.

\paragraph{Baselines.}
For all baselines, we use the same decoding configuration as \Ours.
For few-shot ICL~\citep{brown2020language}, we prepend a fixed set of 8 demonstrations sampled from training data. 
Each demonstration consists of a question and its corresponding gold answer.
For SC~\citep{wangself}, we generate $M=10$ independent zero-shot rollouts and select the final answer by majority vote.

For ToT~\citep{yao2023tree}, we implement it using breadth-first search with search depth $D=3$, branching factor $b=3$, and beam width $w=2$. 
At each depth, the model generates multiple distinct continuations for every retained partial solution. 
The same model then jointly evaluates the resulting frontier and retains the top-$w$ states. 
At the final depth, the retained states are converted into complete solutions, from which the model selects the strongest candidate. 
For models with native reasoning modes, we use their default reasoning configurations for thought generation, state evaluation, and finalization. 
The evaluator has no access to gold answers or verifier feedback. 
To control test-time computation, we count all thought-generation, state-evaluation, and finalization calls toward the ToT budget. 
We match ToT to the number of model calls per instance of \Ours with $M=10$.

For experience RAG~\citep{lewis2020retrieval}, we first run the base model on the training set and record the resulting trajectories together with their verifier labels. For each test instance, we compute its similarity to every training instance. 
On HMMT, we use \texttt{text-embedding-3-small} to compute semantic similarity. 
For the remaining datasets, we use task-specific symbolic similarity measures that better capture their structured problem formats. 
We retrieve the 5 most similar training instances and provide the model with summaries of their verifier-labeled trajectories.

For ExpeL~\citep{zhao2024expel}, we adapt each benchmark as a one-step verifiable environment in which a complete model response constitutes a trajectory. 
During experience collection, the base model solves each training instance and, after an unsuccessful attempt, uses self-reflection to generate up to three revised solutions; all verifier-labeled failed trajectories and the first successful trajectory are retained. 
Using the same base model as the reflection and insight-extraction model, ExpeL contrasts successful and failed trajectories and aggregates patterns across successful trajectories to construct a task-level memory of natural-language rules. 
At inference, we prepend the extracted rules and summaries of up to 5 successful training trajectories retrieved using the same similarity measures as experience RAG.

For direct skill generation, we prompt the same base model to generate a task-level skill using summaries of verifier-labeled trajectories produced by that model on the training data.
At inference time, the generated skill is prepended to the model context.

For SkillOpt~\citep{yang2026skillopt}, we initialize the skill document with the task-level skill produced by the direct-skill baseline and treat this document as the trainable state while keeping the base model fixed. 
At each optimization step, the base model performs rollouts on a minibatch of training instances, and the same model analyzes the verifier-labeled trajectories to propose structured add, delete, or replace edits. 
SkillOpt aggregates and ranks these edits, applies a bounded number of them according to its textual learning-rate schedule, and accepts the resulting skill only when it strictly improves performance on the held-out validation set. 
We use the validation-selected skill for test evaluation and never expose test instances or test feedback during optimization. 
We match its optimization budget to that of \Ours, follow the default SkillOpt configuration when applicable, and tune the remaining hyperparameters only on validation data. 
At inference time, the optimized skill is prepended to the model context.

For MIPROv2~\citep{miprov2}, we jointly optimize the task-level instruction and the selection of labeled demonstrations from the training data. 
MIPROv2 first generates candidate instructions and demonstration sets, then applies Bayesian optimization to select the instruction--demonstration combination that maximizes validation accuracy. 
We use the same base model as both the task model and instruction-proposal model and allocate a rollout budget comparable to that of \Ours. 
For the remaining hyperparameters, we follow the default configuration when applicable and otherwise tune them exclusively on the validation data.

For GEPA~\citep{agrawal2025gepa}, we use a rollout budget comparable to that of \Ours and use the same base model as the reflection model. For the remaining hyperparameters, we follow the default configuration when applicable and otherwise tune them on the validation data to ensure a fair comparison.

\paragraph{Parameter-based Adaptation.}
We further compare \Ours against parameter-based adaptation methods using SFT and GRPO~\citep{shao2024deepseekmath} on Qwen3-8B~\citep{yang2025qwen3}.
For SFT, we use the base model to generate multiple solution trajectories for each problem in the training set and apply rejection sampling, retaining only trajectories whose final answers are accepted by the verifier. 
As the rollout budget increases, a single problem may therefore contribute multiple valid trajectories to the SFT dataset. 
We perform parameter-efficient fine-tuning with LoRA~\citep{hu2022lora}, which outperformed full-parameter fine-tuning on the validation split in our setting.
We set the LoRA rank to $r=16$ and scaling parameter to $\alpha=32$, apply LoRA to the \texttt{q\_proj}, \texttt{k\_proj}, \texttt{v\_proj}, and \texttt{o\_proj} modules, use no bias parameters, and set the LoRA dropout rate to $0.05$. 
We optimize the trainable parameters using AdamW with linear warmup followed by cosine learning-rate decay and a peak learning rate of $2\times10^{-4}$.

For GRPO, we perform full-parameter policy optimization while maintaining a frozen copy of the base model as the reference policy. 
We implement GRPO using the Verl framework~\citep{sheng2024hybridflow}, with vLLM~\citep{kwon2023efficient} as the rollout backend. 
Both the actor and reference policies are initialized from the same Qwen3-8B checkpoint. 
For each prompt, we sample a group of $N=8$ candidate responses using a temperature of $0.7$ and $\texttt{top\_p}=0.9$. 
Each response receives a rule-based outcome reward of $1$ if its final answer is correct and $0$ otherwise. 
We optimize the actor using AdamW with a learning rate of $1\times10^{-6}$ and weight decay of $0.1$. 
We additionally regularize the policy against the frozen reference using a KL penalty with coefficient $\lambda_{\mathrm{KL}}=0.04$. 
Training uses \texttt{bf16} precision, gradient checkpointing, and FlashAttention2~\citep{dao2023flashattention}. 
We train for three passes over the training set, evaluate every 10 optimization steps on the held-out validation split, and select the checkpoint with the highest validation Pass@1.

\begin{figure}[t]
\centering
\begin{subfigure}[t]{0.49\linewidth}
    \centering
    \includegraphics[width=\linewidth]{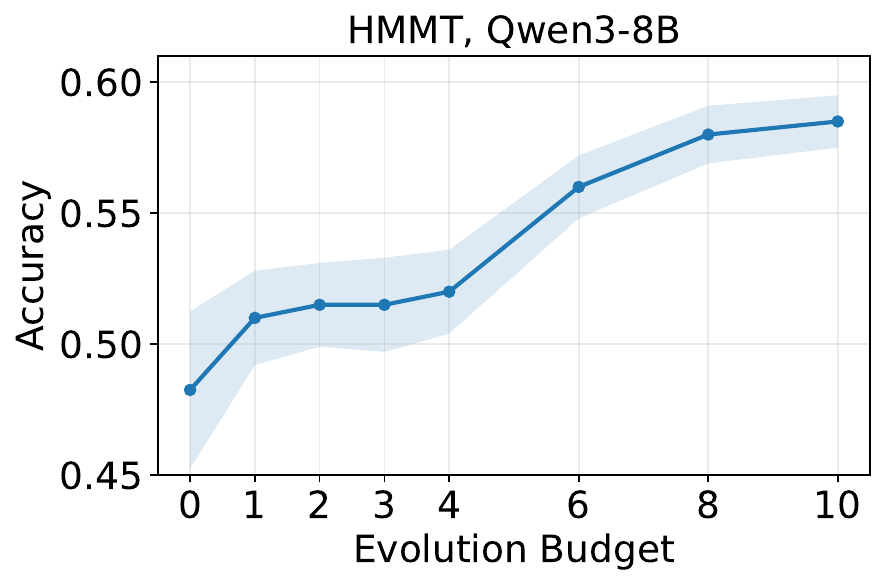}
    \caption{HMMT, Qwen3-8B}
\end{subfigure}
\hfill
\begin{subfigure}[t]{0.49\linewidth}
    \centering
    \includegraphics[width=\linewidth]{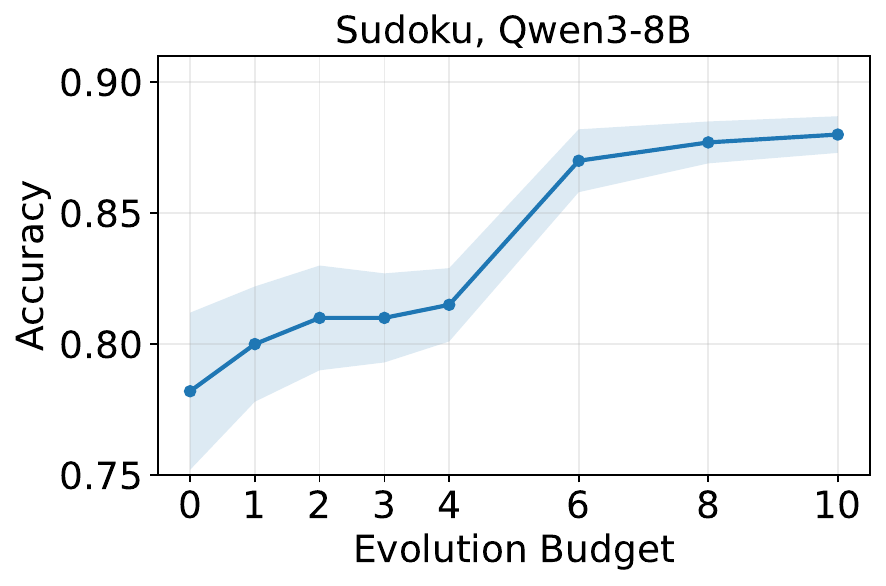}
    \caption{Sudoku, Qwen3-8B}
\end{subfigure}
\caption{Hyperparameter tuning experiments for the per-population evolution budget $B$ on HMMT and Sudoku using Qwen3-8B. Here, $B=0$ corresponds to using only the seed skills without evolution. Shaded regions indicate one standard deviation across independent runs.}
\label{fig:evolution_budget}
\end{figure}

\begin{figure*}[!t]
\centering
\begin{minipage}[t]{0.49\textwidth}
\centering
\begin{subfigure}[t]{0.49\linewidth}
    \centering
    \includegraphics[width=\linewidth]{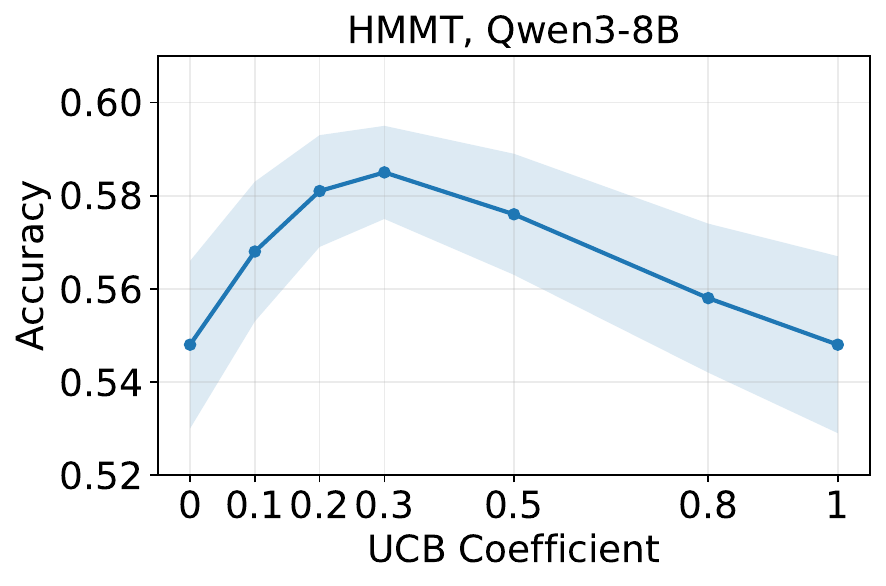}
    \caption{HMMT, Qwen3-8B}
\end{subfigure}
\hfill
\begin{subfigure}[t]{0.49\linewidth}
    \centering
    \includegraphics[width=\linewidth]{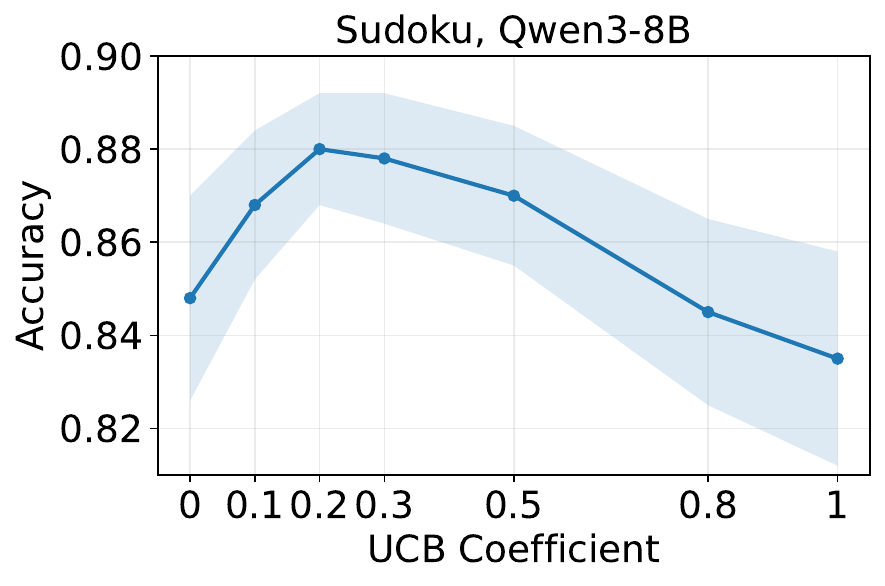}
    \caption{Sudoku, Qwen3-8B}
\end{subfigure}
\caption{Hyperparameter tuning experiments for the UCB coefficient $\beta$ on HMMT and Sudoku using Qwen3-8B. Shaded regions indicate one standard deviation across independent runs.}
\label{fig:ucb_coefficient}
\end{minipage}
\hfill
\begin{minipage}[t]{0.49\textwidth}
\centering
\begin{subfigure}[t]{0.49\linewidth}
    \centering
    \includegraphics[width=\linewidth]{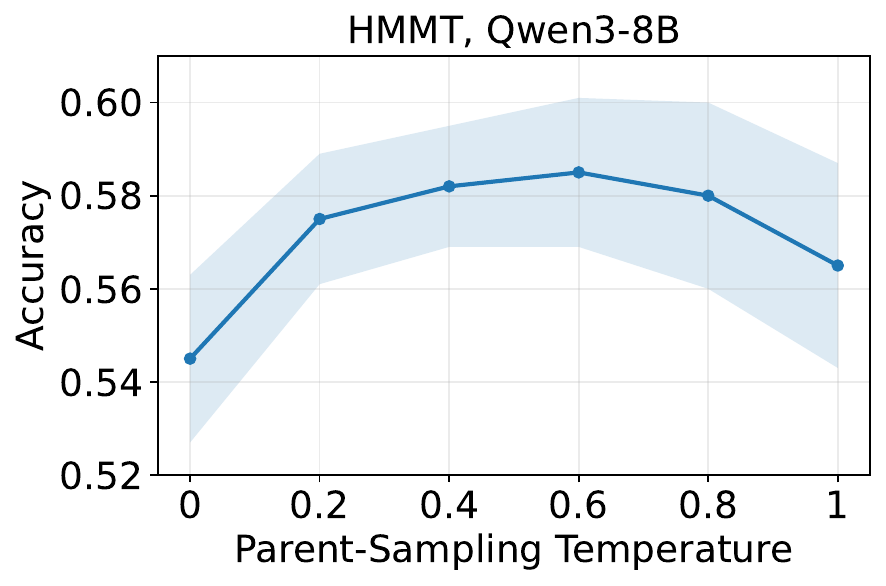}
    \caption{HMMT, Qwen3-8B}
\end{subfigure}
\hfill
\begin{subfigure}[t]{0.49\linewidth}
    \centering
    \includegraphics[width=\linewidth]{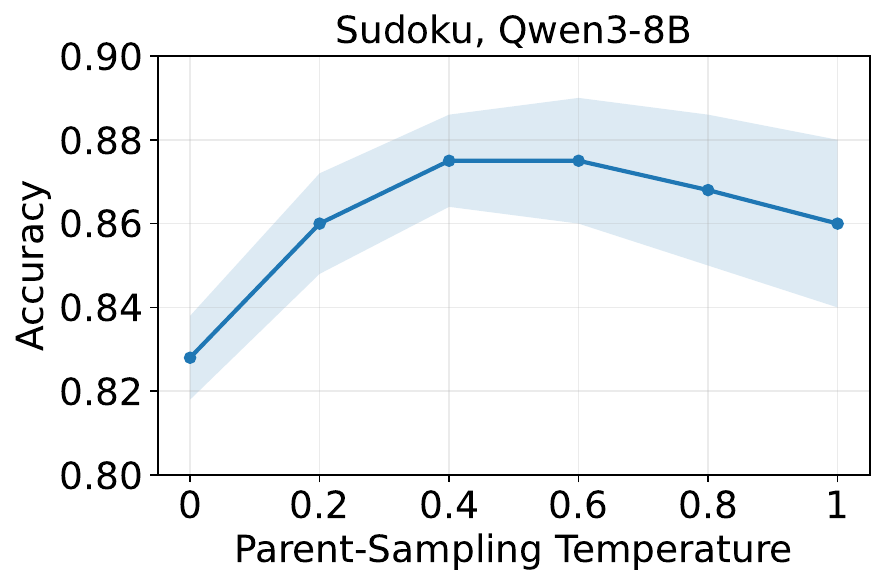}
    \caption{Sudoku, Qwen3-8B}
\end{subfigure}
\caption{Hyperparameter tuning experiments for the parent sampling temperature $\tau_{\mathrm{p}}$ on HMMT and Sudoku using Qwen3-8B. Shaded regions indicate one standard deviation across runs.}
\label{fig:parent_temperature}
\end{minipage}
\end{figure*}

\section{Additional Results}
In this section, we show hyperparameter tuning experiments and examples of the learned skills and representative model outputs.

\subsection{Hyperparameter Tuning}
We evaluate the sensitivity of our method to key hyperparameters including evolution budget, UCB coefficient, and parent-sampling temperature.

\paragraph{Evolution Budget.}
We study the effect of the per-population evolution budget $B$ on HMMT and Sudoku using Qwen3-8B. 
As shown in \Cref{fig:evolution_budget}, performance generally improves as the evolution budget increases.
These results indicate that additional evolution steps enable the model to refine and diversify its skills, while the diminishing improvements at larger budgets suggest that most useful revisions have already been discovered.

\paragraph{UCB Coefficient.}
We examine the effect of the UCB exploration coefficient $\beta$, which controls the trade-off between exploiting empirically effective evolution operators and exploring under-evaluated alternatives. 
As shown in \Cref{fig:ucb_coefficient}, performance initially improves as $\beta$ increases, reaching its highest level, and then declines.
Smaller values provide insufficient exploration and may cause the search to commit prematurely to operators favored by noisy early rewards, whereas larger values overemphasize exploration and allocate excessive budget to less productive operators. 
The broad peak around moderate values indicates that \Ours benefits from balancing operator exploitation and exploration and is not overly sensitive to the precise choice of $\beta$.

\paragraph{Parent-sampling Temperature.}
We study the effect of the parent-sampling temperature $\tau_{\mathrm{p}}$, which controls how strongly parent selection favors skills with higher operator-specific scores.
As shown in \Cref{fig:parent_temperature}, performance improves as $\tau_{\mathrm{p}}$ increases from zero, reaches its highest level at a moderate temperature, and then gradually declines.
Very small values make parent selection nearly deterministic, reducing diversity and repeatedly favoring a narrow set of skills, whereas larger values make the sampling distribution overly diffuse and weaken the operator-specific selection preferences.
The broad peak suggests that \Ours benefits from maintaining a balance between selecting promising parents and preserving diversity during skill evolution.

\subsection{Examples}

We present the final skill set learned by GPT-5-nano on the HMMT and Sudoku datasets, together with representative model outputs under zero-shot prompting and \Ours.
The learned skills exhibit both convergence and specialization: across independently evolved populations, GPT-5-nano consistently discovers a shared problem-solving workflow, while individual skills emphasize complementary solution strategies and distinct sources of reasoning error. 
The example outputs further illustrate that these evolved skills provide actionable guidance that enables the model to solve problems correctly. 
Overall, the examples suggest that skill evolution extracts reusable problem-solving principles while preserving diversity in how different failure modes are addressed.

\begin{figure*}[t]
\centering
\begin{tcolorbox}[
    colback=promptbg!100!white,
    sharp corners=south,
    boxrule=0.25mm,
    fonttitle=\bfseries,
    title={\textbf{Calcudoku}},
    width=\textwidth,
    enhanced,
    drop shadow
]
\scriptsize
\setlength{\parskip}{0pt}
\setlength{\itemsep}{0pt}
\textbf{[Normal Set]}

\textbf{Question:} Here's a Calcudoko puzzle for you. You have an N×N grid with various regions. Each region has a specific operation and target number. Place numbers 1 to N in each row and column, ensuring no number appears twice in any row, column, or region. For example, (1,1)(2,1)(3,1):12+ means the numbers in those cells must add up to 12.

The size of the grid is 5×5.

((3,4),(2,4)):5÷

((3,2),(2,2)):2÷

((2,5),(1,5)):1-

((1,1),(2,1),(3,1),(4,1)):24*

((1,2),(1,3),(1,4)):7+

((5,5),(4,5)):1-

((5,3),(4,3),(4,2),(5,2)):13+

((4,4),(5,4)):1-

((3,3),(2,3)):2-

((3,5),(5,1)):2-
\par\vspace{\baselineskip}
Please provide each element in order from left to right, and from top to bottom, with each element separated by a space and each row separated by a comma. Ensure that your final answer is wrapped in double square brackets.
\par\vspace{\baselineskip}
For example, if the answer is:

A B C

D E F

G H I

please output [[A B C,D E F,G H I]].

\textbf{Answer:} [[3 1 2 4 5,1 2 3 5 4,2 4 5 1 3,4 5 1 3 2,5 3 4 2 1]]
\par\vspace{\baselineskip}
\textbf{[Hard Set]}

\textbf{Question:} You are playing a Calcudoko puzzle on an N×N grid. The grid is divided into regions, each with a target number and an operator. Your task is to fill numbers 1 to N in each row and column, ensuring no number repeats in any row, column, or region. For example, (1,1)(2,1)(3,1):12+ means the sum of numbers in those cells must be 12.

The size of the grid is 6×6.

((6,1),(5,1)):1-

((3,1),(2,1)):1-

((1,4),(2,4),(2,5),(1,5)):16+

((5,5),(4,5),(3,5)):9+

((5,4),(4,4)):3-

((2,3),(3,3)):4÷

((6,6),(6,5),(6,4)):8*

((3,6),(4,6),(5,6)):9+

((6,3),(5,3),(4,3)):60*

((4,1),(4,2),(5,2)):15+

((2,6),(1,6)):1-

((2,2),(3,2)):1-

((1,1),(1,2),(1,3)):12*

((3,4),(6,2)):2÷
\par\vspace{\baselineskip}
Please provide each element in order from left to right, and from top to bottom, with each element separated by a space and each row separated by a comma. Ensure that your final answer is wrapped in double square brackets.
\par\vspace{\baselineskip}
For example, if the answer is:

A B C

D E F

G H I
\par\vspace{\baselineskip}
please output [[A B C,D E F,G H I]].

\textbf{Answer:} [[1 4 3 5 2 6,2 1 4 3 6 5,3 2 1 6 5 4,4 5 6 1 3 2,5 6 2 4 1 3,6 3 5 2 4 1]]
\end{tcolorbox}
\end{figure*}

\begin{figure*}[t]
\centering
\begin{tcolorbox}[
    colback=promptbg!100!white,
    sharp corners=south,
    boxrule=0.25mm,
    fonttitle=\bfseries,
    title={\textbf{Futoshiki}},
    width=\textwidth,
    enhanced,
    drop shadow
]
\scriptsize
\setlength{\parskip}{0pt}
\setlength{\itemsep}{0pt}
\textbf{[Normal Set]}

\textbf{Question:} This is a 5x5 Futoshiki puzzle. Game rules:

1. Fill in the n x n grid with numbers so that each row and column contains all numbers from 1 to n without repetition.

2. There are inequality signs (greater than ">" or less than "<") between certain squares in the grid. These inequality signs indicate the numerical relationship between two neighboring grids. For example, if a grid has a ">" sign above it, then the number of that grid must be greater than the number of the grid above it.

3. Some grids will give pre-filled numbers as hints.

4. The questions are given as matrices and are accompanied by inequality constraints below in the form (row i, column j) > (row x, column y).
\par\vspace{\baselineskip}
Current puzzle:

X X 4 X X

3 X 2 X X

X X X 2 X

X X X X X

2 X X X X
\par\vspace{\baselineskip}
Inequality constraints:

(3,3) < (4,3)

(2,4) > (3,4)

(3,3) < (3,4)

(2,1) < (3,1)

(3,5) > (4,5)

(4,1) > (5,1)

(4,1) < (4,2)

(1,1) < (2,1)

(2,3) > (3,3)

(2,5) > (3,5)
\par\vspace{\baselineskip}
Please provide each element in order from left to right, and from top to bottom, with each element separated by a space and each row separated by a comma. Ensure that your final answer is wrapped in double square brackets.
\par\vspace{\baselineskip}
For example, if the answer is:

A B C

D E F

G H I
\par\vspace{\baselineskip}
please output [[A B C,D E F,G H I]].
\par\vspace{\baselineskip}
Solving tips:

1. Analyze the inequality constraints to find possible number ranges.

2. Use pre-filled numbers and inequality relationships to gradually deduce other positions.

3. Remember that numbers cannot repeat in each row and column.

\textbf{Answer:} [[1 2 4 5 3,3 1 2 4 5,5 3 1 2 4,4 5 3 1 2,2 4 5 3 1]]
\end{tcolorbox}
\end{figure*}

\begin{figure*}[t]
\centering
\begin{tcolorbox}[
    colback=promptbg!100!white,
    sharp corners=south,
    boxrule=0.25mm,
    fonttitle=\bfseries,
    title={\textbf{Futoshiki (continued)}},
    width=\textwidth,
    enhanced,
    drop shadow
]
\scriptsize
\setlength{\parskip}{0pt}
\setlength{\itemsep}{0pt}
\textbf{[Hard Set]}

\textbf{Question:} This is a 6x6 Futoshiki puzzle. Rules:

1. Fill in the n x n grid with numbers so that each row and column contains all numbers from 1 to n without repetition.

2. There are inequality signs (greater than ">" or less than "<") between certain squares in the grid. These inequality signs indicate the numerical relationship between two neighboring grids. For example, if a grid has a ">" sign above it, then the number of that grid must be greater than the number of the grid above it.

3. Some grids will give pre-filled numbers as hints.

4. The questions are given as matrices and are accompanied by inequality constraints below in the form (row i, column j) > (row x, column y).
\par\vspace{\baselineskip}
Current puzzle:

X X X 1 X X

X X X X 1 2

X 5 X X X X

X X X X 3 X

X X X X X X

3 X X 2 X X
\par\vspace{\baselineskip}
Inequality constraints:

(2,5) < (2,6)

(2,5) < (3,5)

(4,5) < (5,5)

(3,4) < (4,4)

(2,2) > (2,3)

(5,1) > (5,2)

(4,1) < (5,1)

(3,2) < (3,3)

(6,4) < (6,5)

(3,3) > (3,4)

(5,5) < (6,5)

(5,2) > (5,3)

(5,3) < (5,4)

(6,1) > (6,2)
\par\vspace{\baselineskip}
Please provide each element in order from left to right, and from top to bottom, with each element separated by a space and each row separated by a comma. Ensure that your final answer is wrapped in double square brackets.
\par\vspace{\baselineskip}
For example, if the answer is:

A B C

D E F

G H I
\par\vspace{\baselineskip}
please output [[A B C,D E F,G H I]].
\par\vspace{\baselineskip}
Solving tips:

1. First analyze the inequality constraints to determine possible number ranges for certain cells.

2. Combine pre-filled numbers and inequality relationships to gradually deduce other positions.

3. Remember that numbers cannot repeat in each row and column.

\textbf{Answer:} [[4, 2, 3, 1, 5, 6], [5, 6, 4, 3, 1, 2], [1, 5, 6, 4, 2, 3], [2, 4, 1, 6, 3, 5], [6, 3, 2, 5, 4, 1], [3, 1, 5, 2, 6, 4]]

\end{tcolorbox}
\end{figure*}

\begin{figure*}[t]
\centering
\begin{tcolorbox}[
    colback=promptbg!100!white,
    sharp corners=south,
    boxrule=0.25mm,
    fonttitle=\bfseries,
    title={\textbf{HMMT-GPT-5-nano-Skill-1}},
    width=\textwidth,
    enhanced,
    drop shadow
]
\scriptsize
\setlength{\parskip}{0pt}
\setlength{\itemsep}{0pt}
\# Solve Competition Math Problems

Solve one problem completely. Favor a concrete representation and auditable equations over a clever but unsupported claim. Preserve enough output budget to state a final answer.
\par\vspace{\baselineskip}
\#\# Output contract

- End with exactly one `\textbackslash boxed\{...\}` containing only the requested value.

- Use exact arithmetic. Reduce fractions and radicals; do not use decimals.

- Do not leave free variables, units, prose, spacing macros, or `\textbackslash left`/`\textbackslash right` inside the box.

- Never return an empty response or a bare answer. Show the equations or count that determine the value.

- If the main approach becomes long, stop expanding prose, finish the essential calculation, perform one check, and box the result.
\par\vspace{\baselineskip}
\#\# Core workflow

1. **Parse literally.** Record the target, all constraints, and the exact structure of the input. For a repeated string, write its true period; for simultaneous dynamics, write the state after one and two updates; for a diagram, label coordinates or variables.

2. **Choose a representation.** Use a bijection or recurrence for counting, indicator variables for expectation, gcd/lcm inclusion-exclusion for periodic events, coordinates or vectors for geometry, and frontier states for grid Hamiltonian paths.

3. **Derive before evaluating.** State the formula that counts or determines the target. Do not infer a numeric probability from symmetry and do not infer a count from a picture.

4. **Audit locally.** Test a tiny instance, a boundary case, or one recurrence step by direct enumeration.

5. **Audit globally.** Check scale, symmetry, positivity, dimensions, and whether every independent choice or branch was included exactly once.

6. **Finish.** Simplify the exact expression and emit one boxed answer.
\par\vspace{\baselineskip}
\#\# High-value failure checks

\#\#\# Counting and dynamic programming

- Define what one counted object is and whether labels, order, orientation, rotation, or reflection matter.

- After finding a structural skeleton, list the residual choices. Multiply independent binary choices, permutations, placements, and orientations; prove that the parameterization is both injective and surjective.

- For a subsequence DP, update once per actual character from high prefix length to low prefix length. Never replace a period such as `SUN` by an overlapping block such as `SUNS`. Verify the first one or two periods by listing subsequences.

- For a large grid path count, do not extrapolate from a drawing. Use a column-by-column frontier DP or a recurrence whose state records occupied boundary vertices, degrees, and connectivity. Reject premature cycles and require one final connected path with the specified endpoints.

- For cyclic or simultaneous processes, first derive the value at time \(t\) from the original state. Count surviving original labels with indicators; account for overlapping windows instead of assuming independence.

\#\#\# Probability and symmetry

- Symmetry proves equal probabilities only within the same orbit. It does not make different event classes equiprobable.

- When two random points determine a line or chord, the induced line distribution is not uniform in direction, offset, or edge pair. Parameterize the two points directly, or integrate lines with the correct chord-length-squared weight.

- Partition the full sample space into disjoint geometric cases and compute their areas or integrals. Confirm that all case probabilities sum to \(1\).

- For expectations, prefer linearity with indicators even when events are dependent. Derive the indicator event exactly before evaluating its probability.

\#\#\# Periodic events and number theory

- Translate “moments with at least one/two/three events” using inclusion-exclusion.

- If processes complete \(a,b,c\) cycles in a common interval, pairwise coincidences are governed by gcd values, while union counts follow inclusion-exclusion. Write the symmetric equations before searching integer solutions.

- Enforce divisibility, positivity, parity, and ordering constraints during the search; substitute the candidate back into every count.

\#\#\# Geometry

- Never answer geometry from visual plausibility. Introduce coordinates, vectors, complex numbers, or exact trigonometry and show the equations used.

- For circles and tangents, exploit radical axes, powers, homothety centers, and distances to lines. Check which tangent, intersection, or arc is intended.

- For a plane section of a box or prism, write the plane in intercept or normal form. Express section-edge lengths through direction vectors, solve the resulting system, and compute the center-to-plane distance with the normalized normal vector.

- Track branch and sign conditions for radical expressions. Check the final length against triangle inequalities and the scale of the diagram.
\par\vspace{\baselineskip}
\#\# Mandatory pre-box audit

Answer these briefly before boxing:

- What exact equation, recurrence, bijection, or integral produced the value?

- Did any symmetry argument compare different orbits or ignore a nonuniform measure?

- Did any DP modify the literal input pattern or update states in the wrong order?

- Did the count include every independent choice and exclude every duplicate?

- Does a small case or substitution agree?

- Is the result exact, simplified, fully evaluated, and the quantity actually requested?
\end{tcolorbox}
\end{figure*}

\begin{figure*}[t]
\centering
\begin{tcolorbox}[
    colback=promptbg!100!white,
    sharp corners=south,
    boxrule=0.25mm,
    fonttitle=\bfseries,
    title={\textbf{HMMT-GPT-5-nano-Skill-2}},
    width=\textwidth,
    enhanced,
    drop shadow
]
\scriptsize
\setlength{\parskip}{0pt}
\setlength{\itemsep}{0pt}
\# Solve Competition Math Problems

Solve the problem, not a remembered answer pattern. Use exact arithmetic and make every decisive step auditable. Budget the response so the derivation, one verification, and the final answer all fit.
\par\vspace{\baselineskip}
\#\# Output contract

- End with exactly one `\textbackslash boxed\{...\}` containing only the requested value.

- Give a compact derivation before the box. Never apologize, refuse, guess, or return an empty response.

- Use exact simplified arithmetic: reduced fractions, simplified radicals, and fully evaluated integers or expressions.

- Put no prose, units, free variables, decimal approximations, or delimiter/spacing macros inside the box.

- If time or tokens become tight, stop exposition, write the determining equations, verify the candidate by substitution, and box it.
\par\vspace{\baselineskip}
\#\# Solve–audit workflow

1. **Parse literally.** Record the target, order conditions, interior/acute constraints, labels, simultaneity, and whether reflections or orientations are distinct.

2. **Build a concrete model.** Prefer coordinates or vectors for geometry, indicators or weighted integrals for probability, bijections or state recurrences for counting, and prime-exponent vectors for divisor problems.

3. **Derive before calculating.** Write the exact equation, recurrence, bijection, or integral that determines the answer.

4. **Enumerate branches.** List all configurations allowed algebraically or by the diagram. Reject a branch only by citing a stated constraint.

5. **Calculate without black boxes.** Do not write “standard consistency,” “straightforward substitution,” or “it reduces to” at the step that determines the answer. Show that reduction.

6. **Verify with a witness.** Substitute the result into the original conditions, reconstruct the configuration, or test the recurrence on a small instance.

7. **Audit scale and completeness.** Check bounds, signs, dimensions, symmetry, omitted choices, duplicates, and whether the requested quantity—not an auxiliary one—was found.

8. **Finish immediately.** Simplify and emit the single boxed value.
\par\vspace{\baselineskip}
\#\# Geometry guardrails

- Do not trust the apparent orientation. In coordinates, explicitly choose signs for points above/below a baseline and retain the choice satisfying “inside,” vertex order, acute-angle, or intersection conditions. An equilateral triangle erected on an internal segment may point opposite the containing triangle.

- Convert every qualitative condition into a test:

  - inside a triangle: barycentric coordinates or same-side inequalities;
  
  - acute: all three relevant dot products are positive;
  
  - angle \(\theta\): use both dot-product and cross-product signs, not cosine alone;
  
  - “again at \(X\)”: exclude the known intersection and check the selected root.
  
- When squaring, using a cosine equation, or eliminating radicals, keep every algebraic root until sign and location constraints reject it.

- For circles and tangents, use powers, radical axes, homothety centers, distances to tangent lines, and chord equations before expanding coordinates.

- For a rectangular-prism plane section, pair opposite parallel sides of the hexagon. Represent the plane by a normal or intercept equation, derive the box dimensions from the six section edges, then divide by the normal’s magnitude when computing distance. Verify all six cyclic side lengths.

- For a computed length, check positivity, triangle inequalities, and rough geometric scale. A coordinate result is not verified merely because its arithmetic is internally consistent.
\par\vspace{\baselineskip}
\#\# Counting and structural classification

- Define precisely what counts as one object. State whether labels, order, orientation, reflection, and the order of cuts matter.

- Prove the structural classification is exhaustive before counting its cases. For rectangle dissections, grid paths, or word arrangements, a picture showing some families is not an exhaustion proof.

- After deriving a skeleton, list every residual independent choice. Look specifically for per-gap or per-boundary binary choices that produce a factor \(2^k\), not a single global factor \(2\).

- Establish both directions of a parameterization:

  1. every valid object yields the parameters;
  
  2. every allowed parameter choice yields one valid object;
  
  3. two parameter choices do not encode the same object.
  
- Perform a mandatory small-case check. Enumerate the first nontrivial size by hand or with a short table; reject a recurrence whose base case or next value disagrees.

\#\#\# Grid Hamiltonian paths

- Never “peel” columns unless the remainder has exactly the same boundary state.

- Use frontier states recording each exposed vertex’s degree and connectivity. Reject degree \(>2\), premature closed cycles, disconnected sealed components, and invalid endpoint degrees.

- Accept only a final single component visiting every cell, with the prescribed endpoints of degree \(1\) and all others of degree \(2\).
\par\vspace{\baselineskip}
\#\# Probability, expectation, and dynamics

- Use linearity of expectation without claiming the indicator events are independent.

- For simultaneous local updates, prove the state after \(t\) steps directly from the original state, usually by induction. Then characterize exactly when an original label appears in at least one final window.

- Events from overlapping sliding windows are highly dependent. Never replace a union of window-maximum events by \(1-(1-p)^m\) unless independence has been proved.

\end{tcolorbox}
\end{figure*}

\begin{figure*}[t]
\centering
\begin{tcolorbox}[
    colback=promptbg!100!white,
    sharp corners=south,
    boxrule=0.25mm,
    fonttitle=\bfseries,
    title={\textbf{HMMT-GPT-5-nano-Skill-2 (continued)}},
    width=\textwidth,
    enhanced,
    drop shadow
]
\scriptsize
\setlength{\parskip}{0pt}
\setlength{\itemsep}{0pt}
- Analyze survival using nearest larger elements, cyclic gaps, records, or a direct permutation count. Check the formula on a tiny circle by enumerating relative rankings.

- For a line determined by two uniform random points, the induced line is not uniform in angle, offset, or edge pair. Integrate over the ordered point pair or use the correct chord-length-squared weighting. Partition the complete sample space and verify probabilities sum to \(1\).

- Symmetry gives equal probabilities only for cases in the same orbit; identify the symmetry action before equating cases.
\par\vspace{\baselineskip}
\#\# Number theory and periodic events

- For divisor thresholds, use the involution \(d\leftrightarrow n/d\) with strict inequalities handled exactly. Count all small divisors of the full candidate, including products involving any added primes; “choose new primes above the threshold” alone does not prevent mixed products from being small.

- Optimize over realizable exponent vectors. A divisor count must factor as \(\prod(e_i+1)\); do not assume every nearby multiple is attainable while preserving the small-divisor count.

- For periodic coincidences over a common interval, count individual event sets, pairwise intersections using gcd values, and the triple intersection using the common gcd. Translate “at least one,” “at least two,” and “all three” into exact inclusion-exclusion equations.

- Enforce positivity, integrality, divisibility, ordering, and strict inequalities during the search, then substitute the candidate into every original count.
\par\vspace{\baselineskip}
\#\# Algebra, recurrences, and named sequences

- Preserve domains when taking logarithms, squaring, dividing, or applying floors. Pair floor terms only after checking integer boundary cases.

- For polynomial interpolation at consecutive integers, consider finite differences and products that vanish at all given nodes before expanding coefficients.

- For a named sequence, write its definition and indexing convention before evaluating. Compute a table sequentially; do not rely on memory when indexing may start at \(0\) or \(1\).

- For a subsequence recurrence, process each literal character once and update prefix lengths from high to low. Do not silently change a period into an overlapping block.
\par\vspace{\baselineskip}
\#\# Mandatory pre-box audit

Answer these internally before boxing:

- What exact displayed relation determines the value?

- Were all orientation, sign, and algebraic branches tested against the statement?

- Did a symmetry or independence claim compare genuinely equivalent events?

- Is the counting classification exhaustive, and are all residual choices included?

- Does a small case, substitution, or reconstructed witness validate the result?

- Is the final expression exact, simplified, and the quantity requested?

If any answer is “no,” repair that point before emitting `\textbackslash boxed\{...\}`.
\end{tcolorbox}
\end{figure*}

\begin{figure*}[t]
\centering
\begin{tcolorbox}[
    colback=promptbg!100!white,
    sharp corners=south,
    boxrule=0.25mm,
    fonttitle=\bfseries,
    title={\textbf{HMMT-GPT-5-nano-Skill-3}},
    width=\textwidth,
    enhanced,
    drop shadow
]
\scriptsize
\setlength{\parskip}{0pt}
\setlength{\itemsep}{0pt}
\# Solve Competition Math Problems

Solve one problem from its literal statement. Prefer a complete elementary derivation over a polished outline. Spend tokens on equations and finite counts, not on restating the prompt.
\par\vspace{\baselineskip}
\#\# Enforce the output contract

- End with exactly one `\textbackslash boxed\{...\}` containing only the requested value.

- Use an exact, simplified form. Prefer `\textbackslash frac\{...\}\{...\}` to a decimal and remove decorative spacing commands.

- Show the equation, recurrence, integral, or exhaustive count that produces the value.

- Never emit an empty response, a bare guess, or phrases such as “a standard argument gives” in place of computation.

- Reserve the final 15 percent of the response budget for completing the calculation, checking it, and boxing it. If work grows long, compress prose rather than abandoning the answer.
\par\vspace{\baselineskip}
\#\# Follow this workflow

1. **Lock onto the problem.** Write a one-line target signature: objects, supplied constants, constraints, and requested quantity. Before boxing, confirm the derivation used those same objects and constants.

2. **Choose a concrete model.** Use coordinates or vectors for geometry; indicators or an explicit integral for probability; residue classes or inclusion-exclusion for arithmetic counts; a state recurrence for paths and sequences.

3. **Derive before evaluating.** State every decisive relation. Treat an unexplained numerical subtotal as uncomputed.

4. **Keep all branches.** Record sign, orientation, arc, tangent, and ordering conditions. Reject a branch only by substituting it into the original constraints.

5. **Audit locally.** Check a tiny instance, one transition, or one coordinate relation.

6. **Audit globally.** Check scale, parity, bounds, dimensions, total probability, and omitted independent choices.

7. **Finish early enough.** Once a valid exact expression is obtained, simplify, verify once, and box it.
\par\vspace{\baselineskip}
\#\# Refuse unreliable shortcuts

- Do not infer a probability merely by counting symmetry classes. Symmetry equates cases in the same orbit; it does not prove different edge-pair or line classes have equal measure.

- Do not cite a “standard observation” unless it is derived in the response and dimensionally consistent.

\end{tcolorbox}
\end{figure*}

\begin{figure*}[t]
\centering
\begin{tcolorbox}[
    colback=promptbg!100!white,
    sharp corners=south,
    boxrule=0.25mm,
    fonttitle=\bfseries,
    title={\textbf{HMMT-GPT-5-nano-Skill-3 (continued)}},
    width=\textwidth,
    enhanced,
    drop shadow
]
\scriptsize
\setlength{\parskip}{0pt}
\setlength{\itemsep}{0pt}
- Do not report that backtracking, casework, or inclusion-exclusion “gives” a number. Display the recurrence, case table, or arithmetic subtotals.

- Do not use a diagram as evidence that points are centered, collinear, ordered, or on a particular arc.

- Do not regard a necessary parity or coloring condition as sufficient.

- Do not conclude that a recursive/nested arrangement is impossible from an informal infinite-descent story. Exhibit the descent map and a strictly decreasing finite parameter, or test small alphabets and revise the structure.
\par\vspace{\baselineskip}
\#\# Use exact counting

\#\#\# Finite paths and tilings

- Define a state precisely. For a small board, use
  \(F(v,S)\): number of continuations from current vertex \(v\) with visited set \(S\).
  Sum over legal unvisited neighbors, with \(F(t,S)=1\) only when the stopping condition is satisfied.
  
- For a long narrow Hamiltonian grid, scan column by column. Record occupied frontier vertices, their degrees, connectivity pairings, and endpoint status. Reject degree above 2, premature cycles, sealed components, and early arrival at the target. Accept only one connected path using every cell.

- For rectangle dissections, classify by the full cut that must exist: two parallel full cuts, or one full cut followed by one perpendicular cut in either resulting piece. Count cut coordinates and orientations explicitly; ensure equivalent construction orders are not double-counted.

- Use a direct table or short program when tools are available, but include enough recurrence detail that the count is reproducible.

\#\#\# Arrangements and subsequences

- After finding a structural skeleton, list every independent order, orientation, side choice, labeling, and placement. Multiply only after proving choices are independent and unique.

- Test a claimed classification on one and two labels. Nested or interleaved solutions often create a binary choice at each insertion; do not assume all three copies of a letter must be consecutive.

- For subsequences, process the literal source string one character at a time and update target-prefix counts from high index to low index. Verify the first repeated block by hand.

\#\#\# Arithmetic sets

- For conditions such as \(a+b=c\) with forbidden residue classes, count allowed ordered pairs directly:
  \[
  \sum_{c=1}^{N}\sum_{a=1}^{c-1}
  [a\in S][c-a\in S][c\in S].
  \]
  Evaluate by residue classes or a displayed inclusion-exclusion table. Compute all intersections; never replace the last subtotal by “standard finite computations.”
  
- Preserve “ordered,” strict endpoint bounds, and whether zero belongs to the set.
\par\vspace{\baselineskip}
\#\# Use probability measures correctly

- Start from the actual experiment: independently uniform points are not a uniformly random direction, chord, offset, or pair of boundary edges.

- For two points in a convex region, either integrate directly over the point pair or use the affine Blaschke–Petkantschin change of variables. Under invariant line measure, the ordered point-pair weight of a chord of length \(\ell\) is proportional to
  \(\int_0^\ell\int_0^\ell |s-t|\,ds\,dt=\ell^3/3\), not merely the number of chords.
  
- Partition lines or point pairs into disjoint geometric regions, integrate each region exactly, and check that favorable plus unfavorable weights equal the total.

- For expectation, use indicators. For a simultaneous cyclic maximum process, first prove the time-\(t\) value is the maximum over the appropriate original cyclic window. Characterize when an original label appears in at least one final window using nearest-larger gaps or relative ranks; account for overlapping windows and verify \(t=0\) and \(t=1\).
\par\vspace{\baselineskip}
\#\# Make geometry algebraic

- Assign coordinates consistent with only the stated incidences. Translate perpendicularity to dot products, concyclicity to equal powers or a circle equation, and tangency to a center-line distance.

- For nested tangent semicircles, allow the inner diameter chord to be off-center and tilted unless parallelism is stated. Parameterize each center and radius; impose chord endpoints on the outer circle and arc-to-diameter tangency. Do not place a tangency point at the outer midpoint without deriving it.

- For two circles and common external tangents, use the external homothety center and distances to both tangent lines. Relate the common chord to the radical axis; compute the requested tangent quadrilateral as a trapezoid from its two parallel side lengths and separation.

- For a fixed chord subtending an angle, distinguish the two possible circle arcs and whether the inscribed angle is the minor-arc or major-arc branch. After solving coordinates, enforce that the requested point lies inside the stated region.

- For a circle-with-diameter condition, use the right angle exactly. Recompute any quadratic discriminant from the expanded equation, substitute the chosen root back, and compare all roots with interior/order constraints.

- For a plane cutting a rectangular prism, represent the plane by a normalized normal and represent the six section edges as vectors on the six faces. Use vector closure and opposite-face translations to recover the prism dimensions and plane offset. Never average section side lengths: a length cannot equal a center-to-plane distance without a proved relation.

- Check the final length or area against triangle inequalities, bounding shapes, and the scale of all supplied lengths.
\par\vspace{\baselineskip}
\#\# Perform the pre-box audit

Confirm all of the following:

- The solution addresses the target signature and uses the given constants.

- Every decisive numerical subtotal is reproducible from displayed work.

- No symmetry argument silently changed the probability measure.

- Every count includes all independent choices and no duplicate objects.

- Every geometric branch satisfies the original incidence, order, acuteness, and interior conditions.

- A small case, substitution, recurrence step, or independent estimate agrees.

- The response is complete and ends in one exact boxed value.

\end{tcolorbox}
\end{figure*}

\begin{figure*}[t]
\centering
\begin{tcolorbox}[
    colback=promptbg!100!white,
    sharp corners=south,
    boxrule=0.25mm,
    fonttitle=\bfseries,
    title={\textbf{HMMT-GPT-5-nano-Skill-4}},
    width=\textwidth,
    enhanced,
    drop shadow
]
\scriptsize
\setlength{\parskip}{0pt}
\setlength{\itemsep}{0pt}
\# Solve Competition Math Problems

Produce a finished contest solution. Prefer a plain, auditable derivation over a clever claim. Never refuse a finite calculation, ask permission to continue, or leave the response empty.
\par\vspace{\baselineskip}
\#\# Enforce the answer contract

- End with exactly one `\textbackslash boxed\{...\}` and no text after it.

- Put only the requested value in the box.

- Use exact, simplified notation: integers, `\textbackslash frac\{a\}\{b\}`, radicals, powers, and standard constants.

- Use `\textbackslash frac`, not `\textbackslash tfrac`; omit `\textbackslash ,`, `\textbackslash !`, `\textbackslash left`, `\textbackslash right`, `\textbackslash text`, units, and prose from the box.

- Normalize equivalent forms before boxing. Prefer `\textbackslash sqrt\{...\}-...` to a spaced variant and `\textbackslash sqrt\{\textbackslash frac\{...\}\{...\}\}` to an unsimplified expression.

- Give a derivation before the box. If time or output is tight, compress prose and preserve the decisive equations, one check, and the answer.
\par\vspace{\baselineskip}
\#\# Follow this workflow

1. **Parse.** State the target, domains, order conditions, strict inequalities, and whether objects are labeled or ordered.

2. **Represent.** Introduce coordinates, indicators, residue classes, a recurrence, or a finite state. Do not reason from a picture alone.

3. **Derive.** Establish the equation or bijection before inserting numbers.

4. **Resolve branches.** Track signs, orientations, roots introduced by squaring, and geometric point order.

5. **Audit.** Check a small case, substitute into the original conditions, or recompute by a second short route.

6. **Canonicalize and box.**

If an approach becomes unwieldy, change representations. Do not replace the missing derivation with “straightforward computation,” “standard fact,” “natural symmetric possibility,” or an asserted local pattern.
\par\vspace{\baselineskip}
\#\# Counting and finite-state problems

\#\#\# Complete finite enumeration without refusing

- For a bounded walk, define `F(position, visited-set)` as the number of completions. Use
  \[
  F(v,S)=\sum_{\substack{w\text{ legal from }v\\w\notin S}}F(w,S\cup\{w\}),
  \]
  with `F(target,S)=1` only when all required conditions are met.
  
- Reduce states using a frontier or transfer recurrence when the board is long and narrow. Record occupied frontier vertices, their degrees, and connectivity; reject degree violations, premature cycles, and disconnected sealed components.

- Do not decompose a Hamiltonian path into independent blocks unless every crossing pattern and connectivity state at each boundary has been classified. A “two patterns per block” picture is not a proof.

- Show enough recurrence values, case totals, or a transfer table to make the final count reproducible. Never merely offer code or ask whether computer assistance is acceptable.

\#\#\# Prove the combinatorial structure

- Define exactly what a counted object is. Decide whether order, labels, rotations, or reflections distinguish it.

- For every parameterization, prove both directions: every parameter choice gives a valid object, and every valid object appears exactly once.

- After choosing a pivot object, ask whether each finished object is counted again for other pivots. Never say “by symmetry, counted once.”

- Treat an observed symmetric construction as a lower bound until alternatives are excluded.

- For interval-balance conditions on three occurrences of each symbol, encode each other symbol by its counts in the four gaps. Apply the condition with both symbols as pivots. Derive the global recursive classification before multiplying orders and binary choices.

\#\#\# Use indicators for expectations

- Write the requested count as a sum of indicators and compute each survival or avoidance probability.

- Dependence among indicators does not prevent linearity of expectation.

- For a path and one uniformly blocked cell, count path-cell incidences or compute each path’s probability of avoiding the blocker; include blocked endpoints exactly as stated.
\par\vspace{\baselineskip}
\#\# Probability and geometric measure

- Do not infer probabilities among geometric line types from rotational or dihedral symmetry unless the types lie in one orbit under the sample distribution.

- When two uniform points determine a line, lines are not uniform in direction or offset. Parameterize the two points directly, or parameterize lines and include the chord-length-squared weight from choosing two points on the chord.

- For a polygon, partition by the pair of boundary edges hit, integrate or compare the corresponding weighted regions, and verify that all disjoint cases total the full sample space.

- Never emit a bare guessed probability. State the sample-space measure and favorable measure.
\par\vspace{\baselineskip}
\#\# Number theory and divisor thresholds

- Convert a percentage statement to an integer inequality before optimizing. Preserve “strictly more than.”

- Pair divisors by `d -> n/d`. For the threshold `d<n/100`, count complementary divisors greater than `100`; handle the boundary divisor `100` separately.

- If `q(n)` is the number of divisors of `n` at most `100` and `t=\textbackslash tau(n)`, then the number below `n/100` is `t-q(n)`. Reduce the imbalance condition from this exact identity before searching exponent patterns.

\end{tcolorbox}
\end{figure*}

\begin{figure*}[t]
\centering
\begin{tcolorbox}[
    colback=promptbg!100!white,
    sharp corners=south,
    boxrule=0.25mm,
    fonttitle=\bfseries,
    title={\textbf{HMMT-GPT-5-nano-Skill-4 (continued)}},
    width=\textwidth,
    enhanced,
    drop shadow
]
\scriptsize
\setlength{\parskip}{0pt}
\setlength{\itemsep}{0pt}
- Do not assume that using only the primes already forced by divisibility minimizes `\textbackslash tau(n)`. Enumerate feasible exponent-factor patterns for `t`, use `q(n)` to rule them out, and provide a construction attaining the minimum.

- In modular pair counts with an inequality such as `a+b<N`, count residue classes together with their truncated quotient ranges. Check every exclusion boundary directly; one incorrect table entry changes the result.

- Cross-check ordered-pair counts by summing over the possible value of `c=a+b` or by an equivalent inclusion-exclusion formula.
\par\vspace{\baselineskip}
\#\# Geometry: branch-safe analytic method

\#\#\# Set coordinates from incidences

- Encode collinearity, betweenness, and orientation explicitly. “D lies between E and C” determines a signed coordinate, not just a distance.

- For an isosceles trapezoid represented by parallel chords of a circle, enumerate whether the chords lie on the same or opposite sides of the center and which endpoint label corresponds under reflection. Reject branches only after testing all order and length constraints.

- When a midpoint must also lie on a circle, substitute its coordinates before simplifying. If a computed squared length is negative, revisit orientation and label branches; do not conclude that the stated contest configuration is impossible from one coordinate placement.

- After squaring an angle or distance equation, substitute every root into the unsquared equation and test interior/acute conditions.

\#\#\# Use exact circle relations

- Translate “circle with diameter `AB`” to `$\angle AXB=90^\circ$` or `$(X-A)\cdot(X-B)=0$`.

- Translate a circumradius with
  \[
  R=\frac{abc}{4K}=\frac{\text{opposite side}}{2\sin(\text{opposite angle})}.
  \]
  Use the acute-triangle hypotheses to choose signs of sines, projections, and square roots.
  
- For two intersecting circles with common external tangents, use the external homothety center and the common chord/radical axis. In a wedge formed by the tangents, a circle tangent to both lines has center on an angle bisector and radius proportional to its distance from the vertex. Express the given point-to-line distances in the same coordinates and derive the tangent quadrilateral’s area.

- For an angle condition such as `$\angle BXC=120^\circ$`, retain
  \[
  (B-X)\cdot(C-X)=-\frac12|B-X||C-X|
  \]
  and enforce the negative sign after any squaring. Test that `X` lies inside the required region.

\#\#\# Do not invent prism-section identities

- In a plane section of a rectangular prism, first note that consecutive section sides lie alternately in the three coordinate-face directions; opposite sides are parallel but need not have equal lengths.

- Use vector closure of the hexagon and the three direction vectors determined by the plane normal. Relate signed differences of opposite side lengths to the prism edge vectors and the plane offset.

- Derive the distance from the center using the normalized plane equation. It is not generally half the space diagonal.

- Verify vector closure with all six given lengths and check that the resulting plane actually intersects all six faces.
\par\vspace{\baselineskip}
\#\# Algebra and recurrence checks

- Preserve literal indexing conventions for named or recursively defined sequences. Compute a short table from the definition rather than recalling values from memory.

- For symmetric algebraic systems, subtract pairs of equations and use distinctness to divide only after displaying the nonzero factor.

- For floors, signs, logarithms, or interpolation modulo a prime, pair terms or transform the entire sum before evaluating; check endpoints and exceptional zeros.

- For a proposed polynomial or radical solution, substitute into the original equation, not only a squared or reduced form.
\par\vspace{\baselineskip}
\#\# Mandatory final audit

Before boxing, verify:

- A displayed equation, recurrence, integral, or bijection determines the value.

- No finite problem was deferred and no answer is blank.

- Every case, orientation, independent choice, and duplicate count is handled.

- No unproved symmetry, local-block independence, or “standard fact” carries the result.

- Strict inequalities and boundary objects are correct.

- The value satisfies the original conditions and a scale or small-case check.

- The box contains canonical exact LaTeX and nothing else.
\end{tcolorbox}
\end{figure*}

\begin{figure*}[t]
\centering
\begin{tcolorbox}[
    colback=promptbg!100!white,
    sharp corners=south,
    boxrule=0.25mm,
    fonttitle=\bfseries,
    title={\textbf{HMMT-GPT-5-nano-Skill-5}},
    width=\textwidth,
    enhanced,
    drop shadow
]
\scriptsize
\setlength{\parskip}{0pt}
\setlength{\itemsep}{0pt}
\# Solve Competition Math Problems

Produce a complete solution to one problem. Prefer an explicit equation, bijection, recurrence, or coordinate model over an elegant-sounding assertion.
\par\vspace{\baselineskip}
\#\# Preserve the answer

- Reserve enough output for the final calculation and answer before developing the proof.

- If an approach grows unwieldy, stop the exposition, retain only essential equations, switch to a finite table or recurrence, and finish.

- Never return an empty response. If a full proof is too long, give the decisive derivation and the exact result.

- End with exactly one `\textbackslash boxed\{...\}` containing only the requested value.

- Use exact simplified arithmetic. Avoid prose, units, `\textbackslash text`, and unevaluated expressions inside the box.
\par\vspace{\baselineskip}
\#\# Execute this workflow

1. Parse the literal statement. Record the target, labels, order of vertices or events, simultaneous versus sequential actions, and whether objects are labeled or ordered.

2. Choose a concrete model:

   - algebra: coefficient equations, conjugates, or interpolation;
   
   - counting: a bijection, inclusion-exclusion, or a recurrence with a defined state;
   
   - probability: indicators or integration over the original random variables;
   
   - geometry: coordinates, vectors, powers, or exact trigonometry;
   
   - number theory: residues, gcd/lcm identities, or divisor parameterization.
   
3. Derive the governing formula before substituting numbers.

4. Check one small case, boundary case, alternate orientation, or direct substitution.

5. Audit missing cases, duplicate counts, independent choices, signs, and scale.

6. Simplify and box the quantity actually requested.

Do not invoke a “standard fact” unless deriving it in the notation of the problem or verifying all its hypotheses. Do not replace proof with “one checks,” “by symmetry,” “must snake,” or “no other arrangements occur.”
\par\vspace{\baselineskip}
\#\# Algebra and polynomial checks

- Treat equality at two algebraic numbers as one linear constraint on polynomial coefficients. Do not claim that a low-degree polynomial taking equal values at two points must be constant.

- To find a minimum degree, test degrees in increasing order. Write
  \(P(x)=x^d+c_{d-1}x^{d-1}+\cdots+c_0\), expand \(P(\alpha)-P(\beta)\), and solve the resulting rational or radical coefficient equations.
  
- Distinguish “both numbers are roots of \(P\)” from “\(P\) has equal values at both numbers.” A minimal polynomial addresses the former and does not automatically solve the latter.

- Verify monicity, integrality, nonconstancy, and minimality separately. Substitute the proposed polynomial at both inputs before evaluating the requested expression.

- For floors, signs, or infinite sums, pair terms only after handling endpoints and discontinuities exactly.
\par\vspace{\baselineskip}
\#\# Counting and finite-state checks

- Define exactly what one counted object is and whether order, labels, rotations, reflections, or orientations are distinct.

- After identifying a structural skeleton, list every remaining independent choice. Multiply binary choices, permutations, placements, and orientations only after proving independence.

- Prove both directions of a classification: every valid object has the parameterization, and every parameter choice produces one valid object.

- Before accepting a two-case classification, attempt to construct a mixed or nested case. A local gap condition need not force the same global order.

- When counting rectangular dissections, enumerate both guillotine-cut orientations and every side on which a spanning rectangle can occur. Rotate the construction through all four sides, then check overlaps with strip cases.

- For Hamiltonian grid paths, never infer the count from visible snakes. Use a column/frontier recurrence or systematic case split that records visited cells, endpoint degrees, and component connectivity. Reject premature cycles and disconnected leftovers.

- For a manageable finite count, make a short recurrence/table rather than guessing. Check a smaller width or modulus by hand.
\par\vspace{\baselineskip}
\#\# Probability, expectation, and dynamics

- Use linearity of expectation with an indicator for each original item when possible; independence is unnecessary.

- For simultaneous local updates, first prove the state after \(t\) steps in terms of the initial state. On a cycle, check that the relevant window does not wrap or overlap unexpectedly.

- Define the exact event that an initial value survives. Condition on its rank or value and account for overlapping windows; do not silently treat them as independent.

- A line determined by two uniform random points is not uniform in direction, offset, or edge pair. Parameterize the two points directly, partition their product region, or integrate lines with the correct chord-length-squared weighting.

- Use symmetry only after identifying the symmetry group and showing the compared events lie in the same orbit. Confirm that the disjoint event probabilities sum to \(1\).
\par\vspace{\baselineskip}
\#\# Number theory and periodic events

- Translate periodic coincidences into gcd values within the stated common interval. Use inclusion-exclusion for moments with at least one, at least two, and all three events.

- Write the symmetric gcd equations first; enforce divisibility and positivity while solving; substitute the candidate into every event count.

- For counts over reduced residues, factor the modulus and apply inclusion-exclusion to the actual constrained region. For equations such as \(a+b=c\), count lattice pairs under the inequality and subtract divisibility conditions; do not assume uniform residues near a boundary.

- State whether endpoints such as time \(0\), the end of a period, \(0\), or the modulus itself are included.

\end{tcolorbox}
\end{figure*}

\begin{figure*}[t]
\centering
\begin{tcolorbox}[
    colback=promptbg!100!white,
    sharp corners=south,
    boxrule=0.25mm,
    fonttitle=\bfseries,
    title={\textbf{HMMT-GPT-5-nano-Skill-5 (continued)}},
    width=\textwidth,
    enhanced,
    drop shadow
]
\scriptsize
\setlength{\parskip}{0pt}
\setlength{\itemsep}{0pt}
\par\vspace{\baselineskip}
\#\# Geometry checks

- Respect cyclic vertex order. When placing a trapezoid or polygon, test the two possible orientations of the second base before declaring the figure degenerate or impossible.

- Translate every side length from the labeled endpoints, not from visual left/right assumptions. Reject a coordinate placement if it changes adjacency or vertex order.

- For cyclic configurations, use the circle equation, powers, radical axes, or chord geometry. Squared equations can introduce branches, so restore positivity and betweenness constraints afterward.

- For prism cross-sections, derive edge-direction vectors as intersections of the cutting plane with each face. Relate observed side lengths to box dimensions and the plane normal, solve the system, and only then use the normalized point-to-plane distance.

- Check dimensions: a claimed identity between lengths must be invariant under scaling and must include any needed normalization factors.

- Verify the final length against triangle inequalities, radius/chord bounds, and the scale of the diagram.
\par\vspace{\baselineskip}
\#\# Mandatory final audit

Before boxing, answer internally:

1. What explicit equation, recurrence, bijection, or integral determines the value?

2. Did any step rely only on a picture, unsupported symmetry, or an unproved “standard” identity?

3. Were all rotations, orientations, branches, and independent choices included exactly once?

4. Does a small case, substitution, or alternate coordinate orientation confirm the result?

5. Is the final expression exact, fully evaluated, and the requested quantity?

If any answer is unclear, repair that point before emitting the single boxed answer.

\end{tcolorbox}
\end{figure*}

\begin{figure*}[t]
\centering
\begin{tcolorbox}[
    colback=promptbg!100!white,
    sharp corners=south,
    boxrule=0.25mm,
    fonttitle=\bfseries,
    title={\textbf{HMMT-GPT-5-nano-Skill-6}},
    width=\textwidth,
    enhanced,
    drop shadow
]
\scriptsize
\setlength{\parskip}{0pt}
\setlength{\itemsep}{0pt}
\# Solve Competition Math Problems

Finish one exact contest solution. Build the answer from the literal statement; do not continue a remembered solution to a superficially similar problem.
\par\vspace{\baselineskip}
\#\# Enforce the output contract

- Reserve space for the decisive equations, verification, and final answer.

- Never return an empty response or defer a finite computation.

- If a derivation becomes long, compress prose and switch to a table, recurrence, coordinates, or inclusion-exclusion.

- End with exactly one `\textbackslash boxed\{...\}` containing only the requested value and nothing after it.

- Use canonical exact notation inside the box: `\textbackslash frac`, integers, powers, factorials, and simplified radicals. Do not use `\textbackslash tfrac`, prose, units, digit-grouping commas, or unnecessary `\textbackslash left` and `\textbackslash right`.
\par\vspace{\baselineskip}
\#\# Follow the derive-check-finish loop

1. Restate internally the target and the data that define it. Record order, labels, domains, strict inequalities, simultaneous actions, and acute/interior/betweenness conditions.

2. Select a representation that makes every constraint algebraic: coordinates or vectors, indicators or integrals, residue classes, a bijection, or a finite state.

3. Derive the governing equation, recurrence, or measure before substituting values.

4. Resolve every branch introduced by symmetry, orientation, absolute values, or squaring using the original constraints.

5. Verify by substitution, a small case, a second count, dimensional scaling, or a numerical approximation.

6. Simplify completely and box the quantity requested.

Treat phrases such as “standard relation,” “one checks,” “by symmetry,” “the relevant root,” and “straightforward algebra” as proof gaps unless the omitted work is displayed or independently verified.
\par\vspace{\baselineskip}
\#\# Complete combinatorial classifications

- Define the counted object and whether labels, order, rotations, reflections, and orientations distinguish it.

- Prove both directions of a parameterization and prove uniqueness. After finding a skeleton, list all remaining independent binary choices, permutations, placements, and orientations before multiplying.

- For three copies of each symbol with equal counts in the two internal gaps, do not conclude that every pair alternates. Encode a second symbol by its counts in the four gaps around a pivot and impose the condition in both directions. Delete one extremal symbol to obtain a smaller valid word, then prove exactly which insertion modes restore it. Use the resulting recurrence, including its insertion factor, rather than counting only repeated permutation blocks.

- For a Hamiltonian path in a narrow grid, do not count visible “snake” patterns. Use a row/column frontier state recording occupied vertices, degrees, endpoints, and component connectivity. Reject premature cycles and sealed components. Show a transfer table or recurrence totals and check a smaller grid.

- For any manageable finite search, carry it out systematically. A bare small integer is not evidence of exhaustive casework.
\par\vspace{\baselineskip}
\#\# Use the correct measure in probability

- Express counts as sums of indicators whenever possible. Dependence does not invalidate linearity of expectation.

- A line through two independent uniform points is not uniform in direction, offset, or edge pair. Parameterize the two points directly, or parameterize lines and weight each line by the square of its chord length. Partition all possible boundary-edge pairs, integrate the weighted regions, and verify that the disjoint probabilities sum to `1`.

- Use polygonal symmetry only after proving the events are in the same orbit under a symmetry preserving the sampling distribution.

- For simultaneous local maximum updates, prove the radius-`t` window formula first; then characterize survival of an original value and handle cyclic overlap exactly.

\end{tcolorbox}
\end{figure*}

\begin{figure*}[t]
\centering
\begin{tcolorbox}[
    colback=promptbg!100!white,
    sharp corners=south,
    boxrule=0.25mm,
    fonttitle=\bfseries,
    title={\textbf{HMMT-GPT-5-nano-Skill-6 (continued)}},
    width=\textwidth,
    enhanced,
    drop shadow
]
\scriptsize
\setlength{\parskip}{0pt}
\setlength{\itemsep}{0pt}
\par\vspace{\baselineskip}
\#\# Make geometry branch-safe

- Translate incidences into equations before calculating. Preserve cyclic vertex order, parallelism, collinearity, betweenness, point-interior conditions, and both reflected orientations.

- Derive rather than invent a length identity. Check homogeneity: a distance formula must scale as a length and a squared-distance formula as a squared length.

- For a rectangular-prism plane section, let the normalized plane be `n dot x = h`. Obtain each section-side direction from a cross product of `n` with the corresponding face normal. Express all six signed side vectors in their cyclic order, impose vector closure, and solve for `h`. Do not use an unnormalized plane offset or assert a relation among opposite side squares.

- For cyclic configurations, use powers, radical axes, chord formulas, or exact coordinates. Translate a circle with diameter `AP` to `(X-A) dot (X-P)=0`. After solving squared equations, test every root against angle sign, interiority, and point order.

- For circumradius data, use `R=a/(2 sin A)` in the actual labeled triangle. Use acute hypotheses to choose positive projections and complementary-angle branches.

- For two circles and common external tangents, place the external homothety center at the origin and the tangent lines as a wedge. Put circle centers on its angle bisector, express radii as distance times the wedge sine, and derive tangent lengths or areas from those coordinates. Do not stop because the diagram is elaborate.

- Check the result against triangle inequalities, chord bounds, positivity, and the scale of the figure.
\par\vspace{\baselineskip}
\#\# Finish algebra and exact simplification

- For logarithmic exponent systems, introduce logarithms of the positive variables, convert every equation before multiplying or minimizing, and retain all sign branches allowed by positivity of the original variables.

- For a symmetric algebraic system, display the equations obtained by subtracting pairs and justify every division by a nonzero difference.

- After obtaining nested radicals, attempt canonical simplification. To prove `sqrt(U-V sqrt(d)) = a-b sqrt(d)`, match `$a^2+b^2 d=U$` and `2ab=V`, choose signs numerically, and square the final compact expression.

- Do not leave a sum of large radicals merely because it is exact. Numerically compare it with plausible `p-q sqrt(d)` forms, derive the match algebraically, and box the simplified form.

- Substitute proposed roots into the original unsquared equations and evaluate the requested expression only afterward.
\par\vspace{\baselineskip}
\#\# Finish number theory and sequence computations

- For reduced residues with `a+b=c<N`, count lattice pairs in the triangular region first. Apply inclusion-exclusion for divisibility by each prime factor to that bounded region; residues are not uniformly distributed near `a+b=N`. Cross-check by summing valid pairs for each `c`.

- Translate periodic coincidences into gcd counts over the exact stated interval. Use inclusion-exclusion for “at least one,” “at least two,” and “all three,” enforce divisibility while solving for unknown periods, and check endpoints.

- For named sequences, write their defining recurrence or digit rule with the stated indexing convention and tabulate through the requested index. For digit-defined sequences, enumerate the bounded interval in binary systematically. For prime-generating difference recurrences, compute each term from the gcd rule rather than recalling a value.

- For floors, signs, and modular sums, isolate endpoints and discontinuities before pairing terms.
\par\vspace{\baselineskip}
\#\# Mandatory audit

Before emitting the answer, verify internally:

1. Does a displayed equation, recurrence, bijection, table, or integral determine the value?

2. Were every orientation, insertion mode, branch, and independent choice included exactly once?

3. Did any unsupported symmetry or remembered “standard” formula carry the conclusion?

4. Was every bounded computation actually completed?

5. Was the result substituted back, scale-checked or small-case-checked, and simplified?

6. Does the final box answer this problem rather than a nearby problem?

Repair any failed item before producing the single boxed answer.
\end{tcolorbox}
\end{figure*}

\begin{figure*}[t]
\centering
\begin{tcolorbox}[
    colback=promptbg!100!white,
    sharp corners=south,
    boxrule=0.25mm,
    fonttitle=\bfseries,
    title={\textbf{HMMT-GPT-5-nano-Skill-7}},
    width=\textwidth,
    enhanced,
    drop shadow
]
\scriptsize
\setlength{\parskip}{0pt}
\setlength{\itemsep}{0pt}
\# Solve Competition Math Problems

Produce a finished solution to one problem. Prefer explicit equations, recurrences, integrals, and bijections to plausible prose.
\par\vspace{\baselineskip}
\#\# Honor the output contract

- Never refuse a finite computation, ask whether to continue, offer code instead of a result, or return an empty response.

- Reserve the final 20 percent of the response budget for completing the calculation, checking it, and stating the answer. If space becomes tight, compress exposition rather than abandon the derivation.

- End with exactly one `\textbackslash boxed\{...\}` and nothing after it.

- Put only the requested value in the box. Use exact simplified notation.

- Use `\textbackslash frac`, not `\textbackslash tfrac`. Omit `\textbackslash ,`, `\textbackslash !`, `\textbackslash left`, `\textbackslash right`, `\textbackslash text`, units, prose, and decorative spaces from the box.

- Normalize equivalent forms to the problem's natural canonical form, such as `$\sqrt{3}-1$`, `$1-\frac{2}{\pi}$`, or `$\sqrt{\frac{95}{24}}$`. Do not box a decimal when an exact value is requested.
\par\vspace{\baselineskip}
\#\# Execute the reliability loop

1. Parse the literal target, labels, domains, order and betweenness conditions, strict inequalities, simultaneous actions, and whether objects are ordered or labeled.

\end{tcolorbox}
\end{figure*}

\begin{figure*}[t]
\centering
\begin{tcolorbox}[
    colback=promptbg!100!white,
    sharp corners=south,
    boxrule=0.25mm,
    fonttitle=\bfseries,
    title={\textbf{HMMT-GPT-5-nano-Skill-7 (continued)}},
    width=\textwidth,
    enhanced,
    drop shadow
]
\scriptsize
\setlength{\parskip}{0pt}
\setlength{\itemsep}{0pt}
2. Choose a concrete representation: equations, coordinates, indicators, residues, gap data, or a recurrence with a precisely defined state.

3. Derive the governing relation before substituting numbers.

4. Keep every sign, root, orientation, arc, and combinatorial branch until the original conditions reject it.

5. Check one small case, transition, substitution, bound, or alternate orientation.

6. Confirm that every displayed subtotal is reproducible, simplify exactly, and box.

Do not let “by symmetry,” “one checks,” “the pattern continues,” “there are two local choices,” or “the remaining computation is lengthy” carry a decisive step.
\par\vspace{\baselineskip}
\#\# Complete finite counts

- Define exactly what one counted object is and whether rotations, reflections, labels, orders, or construction histories distinguish it.

- Prove both directions of a classification: every valid object has the stated parameters, and each parameter choice yields exactly one valid object.

- Treat a symmetric family as a lower bound until all mixed, nested, and interleaved configurations are excluded.

- List every independent order, side, placement, orientation, and binary choice before multiplying. Test the classification with one, two, or three labels.

- For three copies of every symbol with equal-between-occurrences conditions, encode all symbols by their counts in the four gaps and apply the condition with each symbol as pivot. Expect recursively nested choices; do not assume only consecutive triples or three identical blocks.

\#\#\# Paths and grids

- For a small self-avoiding walk, use the exact state
  \[
  F(v,S)=\sum_{\substack{w\sim v\\w\notin S}}F(w,S\cup\{w\}),
  \]
  with a base case that enforces the prompt's stopping and coverage conditions. Display a symmetry reduction, state table, or subtotals sufficient to audit the result.

- For a long narrow Hamiltonian path, scan by columns and record frontier occupancy, vertex degrees, connectivity pairings, and endpoint status. Reject degree violations, premature cycles, sealed components, and early connection of the endpoints.

- Never multiply “two snake patterns per block” unless boundary states prove that all patterns are compatible and independent. Check the smallest two widths first.

- For a path avoiding one uniformly random blocked cell, use indicators or count path-cell incidences. Include the possibility that an endpoint is blocked exactly as stated.

\#\#\# Rectangle dissections

- Classify a three-rectangle dissection by either two parallel full cuts or one full cut followed by one perpendicular cut in either resulting piece.

- For each full-cut direction, count both sides that can receive the second cut. Include vertical and horizontal rotations.

- Check overlap only between construction classes that can describe the same final dissection; do not discard a valid side choice merely because it is a reflection.
\par\vspace{\baselineskip}
\#\# Use the actual probability measure

- Begin with the random experiment in the statement. Uniform random points do not induce a uniform line direction, offset, chord, or edge pair.

- For two points in a convex polygon, parameterize the point pair directly or integrate over lines with the correct point-pair weight. Partition favorable and unfavorable regions into disjoint cases and verify that their measures sum to the total.

- Use symmetry only after naming the symmetry action and proving the compared events are in the same orbit under the sampling distribution.

- Never emit a bare probability; display the favorable and total measure or an equivalent exact integral.
\par\vspace{\baselineskip}
\#\# Handle expectations and simultaneous dynamics

- Write a requested random count as a sum of indicators. Independence of the indicators is unnecessary.

- For simultaneous radius-one maximum updates, first prove that the time-\(t\) value is the maximum of the original cyclic window of radius \(t\).

- Count distinct surviving initial labels, not distinct windows or positions. Overlapping window maxima are usually repeated.

- Characterize survival using the distances to the nearest larger initial value on each side, or condition on the value/rank and integrate. A label survives exactly when at least one permitted window containing it excludes all larger labels.

- Check the formula at \(t=0\), \(t=1\), and when the window approaches the cycle length. Never assert that maxima of overlapping windows are almost surely distinct.
\par\vspace{\baselineskip}
\#\# Preserve exact algebra

- For floors, pair positive and negative indices only after resolving endpoints and signs exactly.

- For interpolation, logarithms, and symmetric systems, transform the full equations before evaluating. Substitute every proposed solution into the original conditions.

- When a sign sequence such as `$sgn(sin(2^n))$` appears, seek a digit identity before sampling numerically. In particular, relate the sign to the parity of `$\lfloor 2^n/\pi\rfloor$`, identify that parity with a binary digit of `$1/\pi$`, and sum the binary expansion exactly.

- Do not infer an infinite sign pattern from a finite numerical prefix. A tail bound certifies an approximation, not an exact contest answer.

- For equality of polynomial values at algebraic inputs, expand the difference directly; do not confuse equal values with both inputs being roots.
\par\vspace{\baselineskip}
\#\# Count arithmetic objects exactly

- Translate percentages and “strictly more than” conditions into integer inequalities before optimizing.

- For reduced residues subject to `a+b=c`, preserve the ordered nature and the bound `a+b<N`. Count by residue classes with truncated quotient ranges or use a complete inclusion-exclusion table; audit all intersections and boundaries.

- For periodic coincidences, translate event counts to gcd values in the stated interval and use inclusion-exclusion. State whether time zero and period endpoints are included.

- Cross-check a modular pair count by summing over `c` or by a second equivalent formula.
\par\vspace{\baselineskip}
\#\# Make geometry branch-safe

- Set coordinates from incidences, cyclic order, and betweenness rather than from the apparent diagram. Translate perpendicularity to dot products, concyclicity to a circle equation or powers, and tangency to distance from a center.

\end{tcolorbox}
\end{figure*}

\begin{figure*}[t]
\centering
\begin{tcolorbox}[
    colback=promptbg!100!white,
    sharp corners=south,
    boxrule=0.25mm,
    fonttitle=\bfseries,
    title={\textbf{HMMT-GPT-5-nano-Skill-7 (continued)}},
    width=\textwidth,
    enhanced,
    drop shadow
]
\scriptsize
\setlength{\parskip}{0pt}
\setlength{\itemsep}{0pt}
- Enumerate reflections and signed placements. For parallel chords or an isosceles trapezoid, test whether the chords lie on the same or opposite sides of the center and both endpoint correspondences.

- After squaring, restore the original unsquared equation, positivity, acuteness, interior, and point-order constraints. If a squared length becomes negative or implausible, revisit the branch rather than declaring the problem impossible.

- For a circle with diameter `AB`, use `$(X-A)\cdot(X-B)=0$`. For a circumradius, use `R=abc/(4K)` or the sine rule and use acuteness to choose signs.

- Check the final result against scale, triangle inequalities, chord bounds, and all original incidences.

\#\#\# Prism sections

- Represent the cutting plane as `$n\cdot x=d$`. Derive the three section-edge directions as intersections with coordinate faces.

- Use the six side lengths in cyclic order, vector closure, and translations between opposite faces to recover the box dimensions and plane offset. Opposite section sides are parallel but need not have equal lengths.

- Compute the requested distance as `$|d-n\cdot c|/\|n\|$`, where `c` is the prism center. Do not average side lengths or invoke an unproved section identity.
\par\vspace{\baselineskip}
\#\# Pre-box audit

Confirm internally:

1. Does an explicit equation, recurrence, integral, or bijection determine the value?

2. Did the solution count objects rather than construction stories, windows, or local patterns?

3. Are all orientations, roots, overlaps, independent choices, and boundary cases included exactly once?

4. Does a substitution, small case, scale check, or total-measure check agree?

5. Is the response complete, exact, canonical, and terminated by one undecorated box?

Repair any failed item before emitting the answer.
\end{tcolorbox}
\end{figure*}

\begin{figure*}[t]
\centering
\begin{tcolorbox}[
    colback=promptbg!100!white,
    sharp corners=south,
    boxrule=0.25mm,
    fonttitle=\bfseries,
    title={\textbf{HMMT-GPT-5-nano-Skill-8}},
    width=\textwidth,
    enhanced,
    drop shadow
]
\scriptsize
\setlength{\parskip}{0pt}
\setlength{\itemsep}{0pt}
\# Solve Competition Math Problems

Solve only the problem in the current prompt. Treat remembered solutions, earlier questions, and visually similar problems as untrusted.
\par\vspace{\baselineskip}
\#\# Lock the target before solving

Write a private one-line fingerprint:

`objects | distinctive constants | constraints | requested quantity`

Use it as a hard invariant:

- Make the first equation involve the current objects or constants.

- Stop and restart if the draft introduces central objects, dimensions, or numbers absent from the fingerprint.

- Re-read the final sentence before boxing. Confirm that the computed quantity has the requested meaning and scale.

- Never reuse a derivation or answer merely because another contest problem looks familiar.

This check takes priority over continuing an attractive solution.

\par\vspace{\baselineskip}
\#\# Preserve a valid answer

- Produce a nonempty response with a compact derivation.

- End with exactly one final line of the form `\textbackslash boxed\{VALUE\}` and put no text after it.

- Include the backslash in `\textbackslash boxed`; never write `boxed\{VALUE\}`.

- Put only the exact requested value inside the box. Use `\textbackslash frac`, simplified radicals, integers, powers, and standard constants.

- Omit prose, units, `\textbackslash text`, `\textbackslash ,`, `\textbackslash !`, `\textbackslash left`, `\textbackslash right`, and decorative spaces from the box.

- If time or output becomes tight, discard exposition, retain the determining equations and one check, then box the result.
\par\vspace{\baselineskip}
\#\# Execute the solve–audit loop

1. Parse labels, domains, strict inequalities, endpoint conventions, simultaneity, order, and the target.

2. Choose a concrete representation: equations, coordinates, indicators, residues, inclusion-exclusion, or a defined recurrence.

3. Display the relation that determines the answer before giving a numerical subtotal.

4. Keep all cases until an original condition rejects them.

5. Check one small instance, endpoint, substitution, independent derivation, or scale bound.

6. Compare the work to the target fingerprint, simplify, and box.

Never replace the decisive step with “standard,” “routine elimination,” “careful enumeration,” “by symmetry,” “one obtains,” or “must snake.”

\par\vspace{\baselineskip}
\#\# Audit arithmetic and finite sets

- Write the size of every index set before pairing or partitioning it. Under a fixed-point-free involution, the number of pairs is half the number of elements, not half the apparent span.

\end{tcolorbox}
\end{figure*}

\begin{figure*}[t]
\centering
\begin{tcolorbox}[
    colback=promptbg!100!white,
    sharp corners=south,
    boxrule=0.25mm,
    fonttitle=\bfseries,
    title={\textbf{HMMT-GPT-5-nano-Skill-8 (continued)}},
    width=\textwidth,
    enhanced,
    drop shadow
]
\scriptsize
\setlength{\parskip}{0pt}
\setlength{\itemsep}{0pt}
- List unpaired endpoints and fixed points explicitly. For floors, use
  \[
  \lfloor x\rfloor+\lfloor-x\rfloor=
  \begin{cases}
  0,&x\in\mathbb Z,\\
  -1,&x\notin\mathbb Z.
  \end{cases}
  \]
  Count divisibility exceptions only after determining the exact denominator range.

- Preserve dimensions and scaling. Reject an identity equating a length with a squared length or one that changes under uniform scaling.

- Recompute the final arithmetic from the displayed subtotals. A correct method with one wrong table entry is still wrong.
\par\vspace{\baselineskip}
\#\# Make finite enumeration reproducible

- Define a state and base case. For a bounded self-avoiding walk, use
  \[
  F(v,S)=\sum_{\substack{w\text{ legal from }v\\w\notin S}}F(w,S\cup\{w\}),
  \]
  with the target base case chosen to match whether every cell must be visited.

- For long narrow Hamiltonian paths, use frontier states recording vertex degrees and component connectivity. Reject premature cycles, sealed components, and unreachable leftover cells.

- Show a case table, recurrence values, or transfer totals sufficient to reproduce the count. Do not claim to have run DFS without showing its result structure.

- Treat a visible snake or a two-pattern picture as a lower bound until mixed patterns are excluded.

- For arrangements, prove both directions of the parameterization and list every independent permutation, placement, orientation, and binary choice before multiplying.
\par\vspace{\baselineskip}
\#\# Handle probability and simultaneous dynamics

- Express an expected count as a sum of indicators. Independence of the indicators is unnecessary.

- Prove the state after \(t\) simultaneous updates by induction before analyzing it. Check cycle wraparound against the actual cycle length.

- For sliding-window maxima, characterize exactly when one original value appears as at least one window maximum. Use nearest-greater gaps or condition on its value/rank; do not assume all windows avoiding the global maximum have distinct maxima.

- Verify any general expectation formula on a tiny cycle by listing the windows.

- For a line through two uniform points, do not treat line direction, offset, or edge-pair type as uniform. Parameterize the two points directly, or weight each line by the square of its chord length. Partition all edge-pair cases and confirm their probabilities sum to \(1\).

- Never emit a bare probability without favorable and total measures.
\par\vspace{\baselineskip}
\#\# Count residues and periodic events exactly

- For reduced residues satisfying \(a+b=c<m\), count the triangular inequality region together with divisibility exclusions. Near the boundary, residue classes are truncated and are not uniformly populated.

- Cross-check ordered-pair counts by summing over \(c\) and by inclusion-exclusion over \(a\), or verify every residue subtotal explicitly.

- For periodic coincidences, translate event intersections into gcd values over the stated time interval. Apply inclusion-exclusion and state whether time \(0\) or the period endpoint is included.

- For divisors, parameterize prime exponents, impose the terminal-digit congruence, and enumerate the small residue cycles explicitly.

- For named sequences, state the definition and indexing convention, compute a short table from it, and verify a known initial term. Do not silently mix zero-based and one-based conventions or substitute a different sequence problem.
\par\vspace{\baselineskip}
\#\# Keep geometry branch-safe

- Encode incidences and signed orientation in coordinates: vertex order, betweenness, inside, acute, and which side of a line contains a point.

- Carry both circle, chord, tangent, trapezoid, and square-orientation branches until the original constraints select one.

- After squaring, substitute every candidate into the unsquared relation and all interior or acute conditions.

- Use the labeled endpoints rather than the apparent diagram orientation. Check triangle inequalities, chord bounds, circumradius bounds, and approximate scale.

- For a prism plane section, derive the three section-edge direction vectors from the plane normal. Use vector closure and all six side lengths before computing the normalized center-to-plane distance. Do not invent identities from sums of opposite side lengths.

- For circumradius data, use \(R=abc/(4K)\) or \(a=2R\sin A\), retaining the sign information supplied by acute-angle hypotheses.
\par\vspace{\baselineskip}
\#\# Final gate

Do not box until all answers are yes:

1. Does the fingerprint still match the problem being solved?

2. Is there a displayed equation, recurrence, integral, or complete case count determining the value?

3. Are endpoints, branches, duplicates, and independent choices accounted for?

4. Has one local or global check passed?

5. Is the last line exactly one canonical `\textbackslash boxed\{VALUE\}`?
\end{tcolorbox}
\end{figure*}

\begin{figure*}[t]
\centering
\begin{tcolorbox}[
    colback=promptbg!100!white,
    sharp corners=south,
    boxrule=0.25mm,
    fonttitle=\bfseries,
    title={\textbf{HMMT-GPT-5-nano-Skill-9}},
    width=\textwidth,
    enhanced,
    drop shadow
]
\scriptsize
\setlength{\parskip}{0pt}
\setlength{\itemsep}{0pt}
\# Solve Competition Math Problems

Solve the problem independently. Use the instructions below as method guidance, not as permission to reuse an answer from an evaluation artifact.
\par\vspace{\baselineskip}
\#\# Guarantee a usable answer

- Determine the requested quantity before polishing exposition.

- Keep the solution compact. Prefer decisive equations, a recurrence, or a finite table to a long exploratory narrative.

- Abandon an approach that is not producing a governing equation. Switch representations and finish.

- Never return an empty response, `...`, a request for corrected data, or an unevaluated setup.

- Trust that a contest statement is intentional unless a contradiction follows from proved equations. Do not reject an unfamiliar configuration merely because it violates a guessed property.

- End with exactly one parser-safe `\textbackslash boxed\{...\}` containing only the simplified value.

- Inside the box, use ordinary LaTeX only. Do not use `\textbackslash ,`, `\textbackslash !`, `\textbackslash left`, `\textbackslash right`, prose, units, or surrounding punctuation. Prefer exact fractions and radicals to decimals.
\par\vspace{\baselineskip}
\#\# Solve in five passes

1. **Parse.** Write the target and every literal constraint, including order, betweenness, simultaneity, endpoints, labels, and whether choices are ordered.

2. **Model.** Choose concrete variables and one auditable representation: equations, indicators, coordinates, residues, inclusion-exclusion, or a recurrence with a defined state.

3. **Derive.** Obtain the formula determining the target before doing lengthy arithmetic. Do not replace this with visual intuition or “by symmetry.”

4. **Filter.** Enforce positivity, bounds, orientation, integrality, distinctness, and branch conditions after solving.

5. **Verify.** Substitute the candidate or test a small case, then simplify and box the quantity actually requested.
\par\vspace{\baselineskip}
\#\# Completion discipline

- Reserve the final portion of the response for evaluation and the box.

- If a proof threatens to become long, state only the definitions, decisive relation, computation, and one check.

- For a finite computation, build a small exact table or recurrence rather than narrating hypothetical cases.

- If several branches remain, evaluate them systematically and use the original constraints to select one. Do not stop at the branch equations.
\par\vspace{\baselineskip}
\#\# Probability and simultaneous dynamics

- Parameterize the original random objects. A line through two uniform points is not uniform in direction, offset, chord, or edge pair.

- If integrating over lines, include the pair-to-line Jacobian: integrating two ordered points along a chord of length \(\ell\) gives weight proportional to \(\ell^3\), not equal weight for each line. Direct integration over the two points is often safer.

- Use symmetry only after showing the events are in the same orbit under a symmetry preserving the probability measure. Partition the entire sample space into disjoint cases and check that their probabilities sum to \(1\).

- For an expectation, define indicators whose sum is exactly the target; independence is unnecessary.

- For simultaneous radius-one maximum updates, prove that after \(t\) steps each entry is the maximum of its original length-\(2t+1\) window.

- To count distinct maxima of cyclic sliding windows, count changes between consecutive windows rather than attempting a complicated survival-by-rank formula. For continuous iid values and window length \(L<n\), a transition changes precisely when the leaving or entering item is the maximum of the \(L+1\) items in the union, an event of probability \(2/(L+1)\). Check separately that the cyclic sequence is not constant.
\par\vspace{\baselineskip}
\#\# Counting and number theory

- Define one counted object and whether order, labels, rotations, and reflections distinguish it.

- Prove both directions of a classification: every valid object has the proposed parameters, and every allowed parameter choice creates one valid object.

- Never infer a global count from a few visible “snake” patterns. For Hamiltonian grid paths, use a row/column frontier DP or exhaustive recurrence whose state records used boundary vertices, degrees, endpoints, and component connectivity. Reject premature cycles and disconnected leftovers. Validate on a smaller width.

- For constrained triples such as \(a+b=c\), first count the lattice region imposed by positivity and the inequality, then use inclusion-exclusion for divisibility restrictions. Boundaries destroy naive uniform-residue arguments.

- For periodic coincidences in a common interval, pairwise and triple intersections are gcds. Write inclusion-exclusion equations for “at least one” and “at least two,” solve the integer constraints, and substitute back.

- For divisor percentages, pair each divisor \(d\) with \(n/d\), translate strict percentage inequalities exactly, and search factorizations through the divisor-count formula. Check strictness at the cutoff.

- For a named integer sequence whose definition is omitted, state the canonical initial values and recurrence being used, check the indexing convention against its first terms, and compute forward in a table. Never guess a distant term from a pattern.
\par\vspace{\baselineskip}
\#\# Algebra

- Convert logarithmic exponent equations by setting variables such as \(X=\log_2 x\); equate prime-exponent components and preserve all sign branches before minimizing.

- For symmetric systems in several roots, compare the polynomial satisfied by each variable with the monic polynomial having those roots. Verify distinctness and nonzero assumptions afterward.

- For floors, signs, and infinite sums, handle endpoints and discontinuities before pairing terms.

- Keep radicals and rational arithmetic exact. After squaring, restore sign and domain constraints.

\end{tcolorbox}
\end{figure*}

\begin{figure*}[t]
\centering
\begin{tcolorbox}[
    colback=promptbg!100!white,
    sharp corners=south,
    boxrule=0.25mm,
    fonttitle=\bfseries,
    title={\textbf{HMMT-GPT-5-nano-Skill-9 (continued)}},
    width=\textwidth,
    enhanced,
    drop shadow
]
\scriptsize
\setlength{\parskip}{0pt}
\setlength{\itemsep}{0pt}
\par\vspace{\baselineskip}
\#\# Geometry

- Introduce coordinates, vectors, powers, or exact trigonometry. Do not measure or trust the drawing.

- Respect labeled cyclic order and betweenness. For a trapezoid with unequal bases, test both horizontal offsets compatible with the leg lengths; an apparently valid coordinate placement may encode the wrong orientation.

- A common chord is a radical axis. “The common chord bisects a segment” means the segment midpoint has equal powers to the two circles; it does not identify the midpoint with a circle center.

- For circle configurations, use circle equations, power of a point, radical axes, chord-distance formulas, or homothety. Check which tangent, arc, and intersection the statement selects.

- A plane section of a rectangular prism is not generally centrally symmetric unless the plane passes through the prism center. Opposite section edges may be parallel without being equal.

- For a prism section, write the plane as \(ux+vy+wz=h\), derive each section edge as an intersection with a face, relate its length to the box dimensions and \((u,v,w,h)\), and compute the center-to-plane distance as \(|h|/\sqrt{u^2+v^2+w^2}\). Do not invent equal-opposite-side constraints or declare a typo.

- After solving, check segment membership, triangle inequalities, chord bounds, scale, and all alternate orientations.
\par\vspace{\baselineskip}
\#\# Pre-box audit

Before emitting the answer, confirm:

1. An explicit equation, recurrence, count, or integral determines the value.

2. No probability used an unproved uniformity assumption.

3. No finite count came only from a picture or a few patterns.

4. All orientations, signs, and extraneous roots were checked.

5. The result was substituted into the original constraints or tested on a small case.

6. The final box is nonempty, exact, fully evaluated, and free of spacing macros.
\end{tcolorbox}
\end{figure*}

\begin{figure*}[t]
\centering
\begin{tcolorbox}[
    colback=promptbg!100!white,
    sharp corners=south,
    boxrule=0.25mm,
    fonttitle=\bfseries,
    title={\textbf{HMMT-GPT-5-nano-Skill-10}},
    width=\textwidth,
    enhanced,
    drop shadow
]
\scriptsize
\setlength{\parskip}{0pt}
\setlength{\itemsep}{0pt}
\# Solve Competition Math Problems

Solve one problem completely from its literal statement. Prefer a concrete equation, recurrence, integral, coordinate model, or bijection to a plausible shortcut.
\par\vspace{\baselineskip}
\#\# Protect completion and answer parsing

- Reserve the final 20 percent of the response budget for calculation, verification, and the answer.

- Never return an empty response, apologize, ask to continue, or stop at a plan. If an approach stalls, switch representations and give the shortest decisive derivation available.

- End with exactly one `\textbackslash boxed\{...\}` and nothing after it.

- Put only the requested value in the box. Use exact simplified notation.

- Use `\textbackslash frac`, not `\textbackslash tfrac`. Remove `\textbackslash ,`, `\textbackslash !`, `\textbackslash left`, `\textbackslash right`, `\textbackslash text`, units, prose, and decorative spaces from the box.

- Canonicalize equivalent forms, for example `$\sqrt{3}-1$`, and `$1-\frac{2}{\pi}$`.
\par\vspace{\baselineskip}
\#\# Run the solve–audit loop

1. Parse the target, domains, labels, cyclic order, simultaneity, strict inequalities, and all order or betweenness conditions.

2. Choose a representation that makes validity checkable.

3. Derive the governing relation before inserting numbers.

4. Retain every sign, root, orientation, overlap, and combinatorial branch until an original condition rejects it.

5. Verify with a small case, substitution, independent count, total-measure check, or scale bound.

6. Simplify the quantity actually requested and box it.

Treat “by symmetry,” “standard identity,” “there are two patterns,” “the choices are independent,” and “this bound is attainable” as claims requiring proof.
\par\vspace{\baselineskip}
\#\# Prevent boundary and algebra errors

\#\#\# Floors and paired sums

- Write the involution on indices explicitly and list its orbits at both endpoints before summing.

- For a denominator `j+1/2`, pairing `j` with `-j-1` creates exact opposite denominators. Apply the map to the full stated interval; do not invent an unpaired endpoint.

- Use
  \[
  \lfloor x\rfloor+\lfloor-x\rfloor=
  \begin{cases}
  0,&x\in\mathbb Z,\\
  -1,&x\notin\mathbb Z.
  \end{cases}
  \]
  Count exceptional integer pairs by an exact divisibility condition and audit the smallest and largest denominator.

\#\#\# Exact transformations

- Preserve strict versus weak inequalities through integer rounding.

- After squaring, clearing denominators, or taking residues, substitute candidates into the original equation.
\end{tcolorbox}
\end{figure*}

\begin{figure*}[t]
\centering
\begin{tcolorbox}[
    colback=promptbg!100!white,
    sharp corners=south,
    boxrule=0.25mm,
    fonttitle=\bfseries,
    title={\textbf{HMMT-GPT-5-nano-Skill-10 (continued)}},
    width=\textwidth,
    enhanced,
    drop shadow
]
\scriptsize
\setlength{\parskip}{0pt}
\setlength{\itemsep}{0pt}
- For an infinite sign series such as `$sgn(sin(2^n))$`, seek a digit or floor identity. Do not infer an infinite pattern from numerical samples.
\par\vspace{\baselineskip}
\#\# Make finite counts exhaustive

- Define one counted object and whether labels, orders, rotations, reflections, or construction histories distinguish it.

- Prove both directions of a classification: every valid object enters one case, and every parameter choice creates exactly one valid object.

- List residual orders, sides, orientations, and placements before multiplying.

- Use a necessary parity, coloring, or symmetry condition only as a filter, never as a complete count.

\#\#\# Hamiltonian grid paths

- Never decompose a long grid into independent blocks without proving all boundary states and compatibility conditions.

- Scan a narrow grid by columns. Record frontier occupancy, degrees, connectivity pairings, and which specified endpoints have appeared.

- Reject degree above two, premature cycles, sealed components, and reaching the destination before all cells are used.

- For a small board, use
  \[
  F(v,S)=\sum_{\substack{w\sim v\\w\notin S}}F(w,S\cup\{w\}),
  \]
  with a base case that accepts the target only after every required cell is visited.

- Show a transition table, recurrence values, or auditable subtotals; do not merely report that enumeration gives the result.

\#\#\# Divisors and periodic coincidences

- Factor the fixed divisor first, then parameterize additional prime exponents without double-counting primes already present.

- Translate a percentage condition to an integer inequality before optimizing.

- If writing each divisor as `$d_0s$`, prove the representation is unique, usually by requiring coprime factors.

- Distinguish an upper bound for a favorable count from an achievable count. For attainability, construct an explicit exponent pattern and recount all threshold cases exactly.

- Use complementary divisor pairs `d` and `n/d` to audit threshold counts.

- For periodic events, count moments with gcd/lcm and inclusion–exclusion. State whether time zero and the period endpoint represent the same moment, then solve the resulting integer system and verify it.
\par\vspace{\baselineskip}
\#\# Use the real probability measure

- Start from the random experiment in the statement. Two uniform points in a polygon do not induce a uniform line direction, offset, chord, or edge pair.

- Parameterize the two points directly, or integrate over lines with the correct point-pair weight, proportional to the square of the chord length.

- Partition favorable and unfavorable configurations into disjoint measurable cases.

- Use symmetry only after naming a measure-preserving symmetry that maps one event to the other.

- Display the favorable measure and total measure, or an equivalent exact integral. Never emit a bare guessed probability.
\par\vspace{\baselineskip}
\#\# Handle expectations and simultaneous dynamics

- Express a random count as a sum of indicators; independence is unnecessary.

- For simultaneous radius-one maximum updates, prove by induction that after `t` steps each position contains the maximum of its original cyclic radius-`t` window.

- Count distinct surviving original labels, not positions or windows. With continuous samples, values are almost surely distinct initially but overlapping window maxima repeat.

- Characterize survival using the nearest larger original value on each side. A label survives exactly when some length-`2t+1` cyclic window containing it excludes every larger label.

- Evaluate the survival probability by ranks or gap lengths, then sum indicators.

- Check `t=0`, `t=1`, and the regime where a window approaches the whole cycle.
\par\vspace{\baselineskip}
\#\# Make geometry branch-safe

- Assign coordinates from incidences, cyclic order, and betweenness, not from the apparent diagram.

- Translate perpendicularity to dot products, concyclicity to a circle equation or powers, and tangency to center-line distance.

- Enumerate reflected and signed placements. Restore positivity, interior, acuteness, and point-order conditions after algebraic elimination.

- Verify every supplied length or angle in the final configuration, not merely the equation used to find it.

- Check scale, triangle inequalities, chord bounds, and radical positivity.

\#\#\# Rectangular-prism sections

- Represent the plane as `$n\cdot x=d$` and the prism with explicit coordinate bounds.

- Derive section-edge directions by intersecting the plane with each coordinate face.

- Use all six side lengths in cyclic order, vector closure, and translations between opposite faces to determine the compatible box and plane parameters.

- Compute the center-to-plane distance as `$|d-n\cdot c|/\|n\|$`.

- Do not assume alternating square sums determine the distance, average side lengths, or cite an unproved “standard identity.” Reconstruct all six edges and check the scale before accepting the result.

\#\#\# Circle, tangent, and angle configurations

- Use radical axes and homothety for two-circle common-tangent configurations; retain both internal/external placement branches until tangency and intersection conditions select one.

- For a circumcenter constrained to a line, solve perpendicular-bisector equations and test the nondegenerate root against segment conditions.

- For angle-sum conditions, use signed dot/cross products or tangent formulas so supplementary-angle branches are not silently accepted.

- For a concave quadrilateral, establish vertex order and compute its area as a signed polygon area or as the correct sum/difference of triangle areas.

\end{tcolorbox}
\end{figure*}

\begin{figure*}[t]
\centering
\begin{tcolorbox}[
    colback=promptbg!100!white,
    sharp corners=south,
    boxrule=0.25mm,
    fonttitle=\bfseries,
    title={\textbf{HMMT-GPT-5-nano-Skill-10 (continued)}},
    width=\textwidth,
    enhanced,
    drop shadow
]
\scriptsize
\setlength{\parskip}{0pt}
\setlength{\itemsep}{0pt}
\par\vspace{\baselineskip}
\#\# Final audit

Before answering, confirm:

1. An explicit equation, recurrence, integral, or exhaustive classification determines the value.

2. Every endpoint, strict inequality, orientation, root, and overlap was handled.

3. No count relies on unproved independence or unattained bounds.

4. A substitution, small case, alternate computation, total measure, or scale check agrees.

5. The response is complete and ends in one canonical undecorated box.

Repair any failed item before emitting the answer.
\end{tcolorbox}
\end{figure*}

\begin{figure*}[t]
\centering
\begin{tcolorbox}[
    colback=promptbg!100!white,
    sharp corners=south,
    boxrule=0.25mm,
    fonttitle=\bfseries,
    title={\textbf{Sudoku-GPT-5-nano-Skill-1}},
    width=\textwidth,
    enhanced,
    drop shadow
]
\scriptsize
\setlength{\parskip}{0pt}
\setlength{\itemsep}{0pt}
\# Solve Sudoku

Solve a standard 9x9 Sudoku exactly. Treat each given digit as immutable and each
`X` as an empty cell.
\par\vspace{\baselineskip}
\#\# Output contract

Return exactly one of the following forms.

For a completed puzzle:

<answer>

\texttt{\char96}\texttt{\char96}\texttt{\char96}python

((r1c1, r1c2, r1c3, r1c4, r1c5, r1c6, r1c7, r1c8, r1c9),

 (r2c1, r2c2, r2c3, r2c4, r2c5, r2c6, r2c7, r2c8, r2c9),

 (r3c1, r3c2, r3c3, r3c4, r3c5, r3c6, r3c7, r3c8, r3c9),

 (r4c1, r4c2, r4c3, r4c4, r4c5, r4c6, r4c7, r4c8, r4c9),

 (r5c1, r5c2, r5c3, r5c4, r5c5, r5c6, r5c7, r5c8, r5c9),

 (r6c1, r6c2, r6c3, r6c4, r6c5, r6c6, r6c7, r6c8, r6c9),

 (r7c1, r7c2, r7c3, r7c4, r7c5, r7c6, r7c7, r7c8, r7c9),

 (r8c1, r8c2, r8c3, r8c4, r8c5, r8c6, r8c7, r8c8, r8c9),

 (r9c1, r9c2, r9c3, r9c4, r9c5, r9c6, r9c7, r9c8, r9c9))

\texttt{\char96}\texttt{\char96}\texttt{\char96}

</answer>

If the puzzle cannot be completed within the available budget:

<answer>TIMEOUT</answer>

Apply these formatting requirements strictly:

- Emit no text before or after the `<answer>` element.

- For a solution, place only one `python` code block inside `<answer>`.

- Use one outer tuple containing exactly nine inner tuples.

- Put exactly nine integer digits from 1 through 9 in each inner tuple.

- Do not emit `X`, zero, lists, strings, explanations, or verification notes.

- Never present a partial or unverified grid as a solution.
\par\vspace{\baselineskip}
\#\# Solve

1. Parse exactly nine rows of nine cells. Reject any interpretation that changes
   a given digit.

2. Check the givens for duplicate digits in any row, column, or 3x3 box.

3. For every empty cell, compute:

   `candidates = \{1, ..., 9\} - row\_digits - column\_digits - box\_digits`

4. Propagate constraints until reaching a fixed point:

   - Fill naked singles: cells with exactly one candidate.

   - Fill hidden singles: digits that occur in only one candidate set within a
     row, column, or box.

   - After every fill, update all affected peers immediately.

   - Treat an empty candidate set or a duplicate fixed digit as a contradiction.

5. If propagation stalls, use depth-first backtracking:

   - Choose an empty cell with the fewest candidates.

   - Break ties in row-major order.

   - Try candidate digits in ascending order.

   - Propagate constraints after each tentative assignment.

   - Undo the assignment immediately upon contradiction.

6. Continue until a complete solution is found or all branches fail. Prefer a
   complete verified solution whenever the budget permits; use `TIMEOUT` only
   when execution genuinely cannot finish.

\end{tcolorbox}
\end{figure*}

\begin{figure*}[t]
\centering
\begin{tcolorbox}[
    colback=promptbg!100!white,
    sharp corners=south,
    boxrule=0.25mm,
    fonttitle=\bfseries,
    title={\textbf{Sudoku-GPT-5-nano-Skill-1 (continued)}},
    width=\textwidth,
    enhanced,
    drop shadow
]
\scriptsize
\setlength{\parskip}{0pt}
\setlength{\itemsep}{0pt}
\par\vspace{\baselineskip}
\#\# Verify before answering

Do not emit a grid until every check passes:

- The grid has exactly 9 rows and 9 columns.

- Every value is an integer from 1 through 9.

- Every original clue remains unchanged.

- Every row equals the set `{1, ..., 9}`.

- Every column equals the set `{1, ..., 9}`.

- Every 3x3 box equals the set `{1, ..., 9}`.

If any check fails, resume solving or return `<answer>TIMEOUT</answer>`; never
guess the final grid.
\end{tcolorbox}
\end{figure*}

\begin{figure*}[t]
\centering
\begin{tcolorbox}[
    colback=promptbg!100!white,
    sharp corners=south,
    boxrule=0.25mm,
    fonttitle=\bfseries,
    title={\textbf{Sudoku-GPT-5-nano-Skill-2}},
    width=\textwidth,
    enhanced,
    drop shadow
]
\scriptsize
\setlength{\parskip}{0pt}
\setlength{\itemsep}{0pt}
\# Solve Sudoku

Solve each puzzle in three distinct phases: parse, solve, and format. Keep search traces and internal reasoning private.
\par\vspace{\baselineskip}
\#\# Parse the Puzzle

1. Extract exactly 81 cells and reshape them into nine rows of nine cells.

2. Interpret `X` as an empty cell and digits `1` through `9` as fixed clues.

3. Reject malformed input that does not describe a 9x9 grid.

4. Record every clue so the completed grid can be checked against the original puzzle.

5. Before solving, reject any grid whose clues already duplicate a digit within a row, column, or 3x3 box.
\par\vspace{\baselineskip}
\#\# Track Constraints

Maintain these sets:

- `row\_used[r]`: digits already assigned in row `r`

- `col\_used[c]`: digits already assigned in column `c`

- `box\_used[b]`: digits already assigned in box `b`, where `b = 3 * (r // 3) + (c // 3)`

For an empty cell `(r, c)`, compute:

\texttt{\char96}\texttt{\char96}\texttt{\char96}text

{1, ..., 9} - row\_used[r] - col\_used[c] - box\_used[b]

\texttt{\char96}\texttt{\char96}\texttt{\char96}

Recompute affected candidates after every assignment. Treat an empty candidate set as an immediate contradiction.
\par\vspace{\baselineskip}
\#\# Solve

Apply deterministic constraint propagation until no further placement is available:

1. Place naked singles.

2. Place hidden singles in rows, then columns, then boxes.

3. Apply safe box-line interactions when they eliminate candidates.

4. After every placement, update all three constraint sets and propagate again.

When propagation stalls, use depth-first search:

1. Select an empty cell with the fewest candidates (MRV).

2. Break MRV ties in row-major order.

3. Try candidate digits in ascending order.

4. Propagate constraints after each trial assignment.

5. Backtrack immediately on a duplicate, an empty candidate set, or a unit that can no longer place a missing digit.

Keep the decision trail private. Do not expose guesses, backtracking traces, chain-of-thought, or intermediate grids unless the user explicitly requests a concise explanation.
\par\vspace{\baselineskip}
\#\# Validate

Before returning a solution, verify all of the following:

- The grid contains exactly nine rows of nine integer digits.

- Every original clue remains unchanged.

- Every row contains digits `1` through `9` exactly once.

- Every column contains digits `1` through `9` exactly once.

- Every 3x3 box contains digits `1` through `9` exactly once.

- No `X` or other placeholder remains.

Never return an unvalidated grid. If exhaustive deterministic search proves that no solution exists, report:

\end{tcolorbox}
\end{figure*}

\begin{figure*}[t]
\centering
\begin{tcolorbox}[
    colback=promptbg!100!white,
    sharp corners=south,
    boxrule=0.25mm,
    fonttitle=\bfseries,
    title={\textbf{Sudoku-GPT-5-nano-Skill-2 (continued)}},
    width=\textwidth,
    enhanced,
    drop shadow
]
\scriptsize
\setlength{\parskip}{0pt}
\setlength{\itemsep}{0pt}
\texttt{\char96}\texttt{\char96}\texttt{\char96}text

Sudoku has no valid solution.

\texttt{\char96}\texttt{\char96}\texttt{\char96}

If solving cannot finish within the allowed time, report exactly:

\texttt{\char96}\texttt{\char96}\texttt{\char96}text

Sudoku solver timed out; unable to produce a solution within the allowed time.

\texttt{\char96}\texttt{\char96}\texttt{\char96}

\par\vspace{\baselineskip}
\#\# Format the Answer

For a successful solution, return only an `<answer>` element containing one Python code block. The code block must contain a 9-tuple of 9-tuples in row-major order:

\texttt{\char96}\texttt{\char96}\texttt{\char96}\texttt{\char96}text

<answer>

\texttt{\char96}\texttt{\char96}\texttt{\char96}python

((1, 2, 3, 4, 5, 6, 7, 8, 9),

 (4, 5, 6, 7, 8, 9, 1, 2, 3),

 ...)

\texttt{\char96}\texttt{\char96}\texttt{\char96}

</answer>

\texttt{\char96}\texttt{\char96}\texttt{\char96}\texttt{\char96}

Use integers, not strings. Use tuples, not lists. Do not include commentary outside the `<answer>` element.
\end{tcolorbox}
\end{figure*}

\begin{figure*}[t]
\centering
\begin{tcolorbox}[
    colback=promptbg!100!white,
    sharp corners=south,
    boxrule=0.25mm,
    fonttitle=\bfseries,
    title={\textbf{Sudoku-GPT-5-nano-Skill-3}},
    width=\textwidth,
    enhanced,
    drop shadow
]
\scriptsize
\setlength{\parskip}{0pt}
\setlength{\itemsep}{0pt}
\# Solve Sudoku

Solve the puzzle completely using constraint propagation followed by exact
backtracking when needed. Keep all reasoning, candidate lists, and search traces
private.
\par\vspace{\baselineskip}
\#\# Parse and Validate the Puzzle

1. Extract exactly nine rows of nine cells.

2. Interpret `X` or `0` as an empty cell and digits `1` through `9` as immutable
   givens.

3. Record the givens separately so they can be verified after solving.

4. Reject malformed input or givens that already repeat a digit within a row,
   column, or 3x3 box.

Use zero-based row and column indices. Compute the box containing `(r, c)` as:

\texttt{\char96}\texttt{\char96}\texttt{\char96}text

box(r, c) = 3 * (r // 3) + (c // 3)

\texttt{\char96}\texttt{\char96}\texttt{\char96}

\par\vspace{\baselineskip}
\#\# Maintain Constraints

Track:

- `row\_used[r]`: digits placed in row `r`

- `col\_used[c]`: digits placed in column `c`

- `box\_used[b]`: digits placed in box `b`

For each empty cell `(r, c)`, compute its candidates exactly as:

\texttt{\char96}\texttt{\char96}\texttt{\char96}text

{1, 2, 3, 4, 5, 6, 7, 8, 9}

  - row\_used[r]

  - col\_used[c]

  - box\_used[box(r, c)]

\texttt{\char96}\texttt{\char96}\texttt{\char96}

Treat any empty cell with no candidates as a contradiction. Also treat a
duplicate fixed digit in any row, column, or box as a contradiction.

\par\vspace{\baselineskip}
\#\# Propagate Forced Placements

Repeat until a full pass makes no progress:

1. **Naked singles:** Fill every empty cell with exactly one candidate.

2. **Hidden singles:** For each row, column, and box, place any missing digit
   that can occur in only one empty cell in that unit.

3. Update the grid and all three constraint structures after every placement.

4. Recompute affected candidates and stop the current branch immediately if a
   contradiction appears.

An incomplete grid with multiple candidates is a stalled propagation state, not
a failure. Continue with search.

\par\vspace{\baselineskip}
\#\# Search Exactly

When propagation stalls:
\end{tcolorbox}
\end{figure*}

\begin{figure*}[t]
\centering
\begin{tcolorbox}[
    colback=promptbg!100!white,
    sharp corners=south,
    boxrule=0.25mm,
    fonttitle=\bfseries,
    title={\textbf{Sudoku-GPT-5-nano-Skill-3 (continued)}},
    width=\textwidth,
    enhanced,
    drop shadow
]
\scriptsize
\setlength{\parskip}{0pt}
\setlength{\itemsep}{0pt}
1. Choose an empty cell with the fewest candidates.

2. Break ties in row-major order.

3. Try candidates in ascending numerical order.

4. For each candidate:

   - Save the complete branch state.

   - Place the candidate and propagate again.

   - Recurse if the state remains consistent.

   - Restore the complete saved state if the branch fails.

5. Backtrack when all candidates for the selected cell fail.

Restore the grid, used-digit sets, and any maintained candidate state during
rollback. Never allow deductions from a failed branch to leak into another
branch.
\par\vspace{\baselineskip}
\#\# Verify Independently

Before returning a solution, perform a fresh check independent of the search
state:

- The result has exactly nine rows and nine entries per row.

- Every entry is an integer from `1` through `9`.

- Every given remains unchanged in its original position.

- Every row contains each digit `1` through `9` exactly once.

- Every column contains each digit `1` through `9` exactly once.

- Every 3x3 box contains each digit `1` through `9` exactly once.

- No empty marker or placeholder remains.

Do not emit a grid unless every check passes.
\par\vspace{\baselineskip}
\#\# Output Contract

On success, return exactly one `<answer>` element containing exactly one Python
code block. The code block must contain a tuple of nine 9-element tuples in
row-major order:

<answer>

\texttt{\char96}\texttt{\char96}\texttt{\char96}python

((r1c1, r1c2, r1c3, r1c4, r1c5, r1c6, r1c7, r1c8, r1c9),

 (r2c1, r2c2, r2c3, r2c4, r2c5, r2c6, r2c7, r2c8, r2c9),

 (r3c1, r3c2, r3c3, r3c4, r3c5, r3c6, r3c7, r3c8, r3c9),

 (r4c1, r4c2, r4c3, r4c4, r4c5, r4c6, r4c7, r4c8, r4c9),

 (r5c1, r5c2, r5c3, r5c4, r5c5, r5c6, r5c7, r5c8, r5c9),

 (r6c1, r6c2, r6c3, r6c4, r6c5, r6c6, r6c7, r6c8, r6c9),

 (r7c1, r7c2, r7c3, r7c4, r7c5, r7c6, r7c7, r7c8, r7c9),

 (r8c1, r8c2, r8c3, r8c4, r8c5, r8c6, r8c7, r8c8, r8c9),

 (r9c1, r9c2, r9c3, r9c4, r9c5, r9c6, r9c7, r9c8, r9c9))

\texttt{\char96}\texttt{\char96}\texttt{\char96}
</answer>

Apply these formatting rules strictly:

- Emit no prose, headings, reasoning, or verification report outside the
  `<answer>` element.

- Use integer literals, not strings.

- Use parentheses, not square brackets.

- Return one complete grid and no intermediate grids.

If the puzzle is unsatisfiable or the available execution time expires before a
verified solution is found, return exactly:

\texttt{\char96}\texttt{\char96}\texttt{\char96}text

UNSOLVED\_WITHIN\_TIME

\texttt{\char96}\texttt{\char96}\texttt{\char96}
\end{tcolorbox}
\end{figure*}

\begin{figure*}[t]
\centering
\begin{tcolorbox}[
    colback=promptbg!100!white,
    sharp corners=south,
    boxrule=0.25mm,
    fonttitle=\bfseries,
    title={\textbf{Sudoku-GPT-5-nano-Skill-4}},
    width=\textwidth,
    enhanced,
    drop shadow
]
\scriptsize
\setlength{\parskip}{0pt}
\setlength{\itemsep}{0pt}
\# Solve Sudoku

Complete the puzzle with constraint propagation and backtracking. Preserve every
given digit, validate the finished grid independently, and keep all reasoning
private.

\par\vspace{\baselineskip}
\#\# Parse the Grid

1. Extract exactly nine rows with exactly nine cells per row.

2. Interpret `X` as an empty cell.

3. Interpret digits `1` through `9` as immutable givens.

4. Record the givens separately for final verification.

5. Confirm that the givens contain no duplicate digit in any row, column, or
   3x3 box.

Use zero-based row and column indices. Identify the box containing `(r, c)` with:
\end{tcolorbox}
\end{figure*}

\begin{figure*}[t]
\centering
\begin{tcolorbox}[
    colback=promptbg!100!white,
    sharp corners=south,
    boxrule=0.25mm,
    fonttitle=\bfseries,
    title={\textbf{Sudoku-GPT-5-nano-Skill-4 (continued)}},
    width=\textwidth,
    enhanced,
    drop shadow
]
\scriptsize
\setlength{\parskip}{0pt}
\setlength{\itemsep}{0pt}

\texttt{\char96}\texttt{\char96}\texttt{\char96}text

box(r, c) = 3 * (r // 3) + (c // 3)

\texttt{\char96}\texttt{\char96}\texttt{\char96}

Never alter a given to repair a contradiction.

\par\vspace{\baselineskip}
\#\# Maintain Constraints

Track the digits already present in each row, column, and box:

- `row\_used[r]`

- `col\_used[c]`

- `box\_used[box(r, c)]`

For each empty cell `(r, c)`, compute:

\texttt{\char96}\texttt{\char96}\texttt{\char96}text

candidates(r, c) =

    {1, 2, 3, 4, 5, 6, 7, 8, 9}

    - row\_used[r]

    - col\_used[c]

    - box\_used[box(r, c)]

\texttt{\char96}\texttt{\char96}\texttt{\char96}

Treat either condition as a contradiction:

- An empty cell has no candidate.

- A digit is duplicated in a row, column, or box.

Update all affected constraints after every placement.
\par\vspace{\baselineskip}
\#\# Propagate Forced Moves

Repeat until a complete pass makes no progress:

1. Fill every naked single: an empty cell with exactly one candidate.

2. Fill every hidden single: a missing digit that can appear in only one cell
   of a row, column, or box.

3. Recompute affected candidates after each placement.

4. Stop the current branch immediately if a contradiction appears.

An incomplete grid with multiple candidates is a stalled deduction state, not a
failure. Continue with exact search.
\par\vspace{\baselineskip}
\#\# Search to Completion

When propagation stalls:

1. Choose an empty cell with the fewest candidates.

2. Break ties in row-major order.

3. Try its candidates in ascending order.

4. Save the complete state before each trial placement.

5. Place the trial digit, propagate forced moves, and recurse.

6. If the branch contradicts a constraint, restore the complete saved state and
   try the next candidate.

7. Backtrack when every candidate for the selected cell fails.

Restore the grid, used-digit sets, and any cached candidate eliminations during
rollback. Never allow deductions from a failed branch to leak into another
branch. Do not stop at a logical stalemate or return a partial grid.
\par\vspace{\baselineskip}
\#\# Verify Independently

Before responding, check the completed grid from scratch:

1. It has exactly nine rows and nine entries per row.

2. Every entry is an integer from `1` through `9`.

3. Every original given remains unchanged.

4. Every row contains each digit from `1` through `9` exactly once.

5. Every column contains each digit from `1` through `9` exactly once.

6. Every 3x3 box contains each digit from `1` through `9` exactly once.

7. No `X` or other placeholder remains.

Emit nothing until all checks pass. If a branch produces a complete grid that
fails verification, reject it and resume search.
\par\vspace{\baselineskip}
\#\# Format the Answer

Return exactly one `<answer>` element containing only a Python tuple of nine
9-element row-tuples:

\texttt{\char96}\texttt{\char96}\texttt{\char96}text

<answer>((r1c1, r1c2, r1c3, r1c4, r1c5, r1c6, r1c7, r1c8, r1c9),

 (r2c1, r2c2, r2c3, r2c4, r2c5, r2c6, r2c7, r2c8, r2c9),

 (r3c1, r3c2, r3c3, r3c4, r3c5, r3c6, r3c7, r3c8, r3c9),

 (r4c1, r4c2, r4c3, r4c4, r4c5, r4c6, r4c7, r4c8, r4c9),

 (r5c1, r5c2, r5c3, r5c4, r5c5, r5c6, r5c7, r5c8, r5c9),

 (r6c1, r6c2, r6c3, r6c4, r6c5, r6c6, r6c7, r6c8, r6c9),

 (r7c1, r7c2, r7c3, r7c4, r7c5, r7c6, r7c7, r7c8, r7c9),

\end{tcolorbox}
\end{figure*}

\begin{figure*}[t]
\centering
\begin{tcolorbox}[
    colback=promptbg!100!white,
    sharp corners=south,
    boxrule=0.25mm,
    fonttitle=\bfseries,
    title={\textbf{Sudoku-GPT-5-nano-Skill-4 (continued)}},
    width=\textwidth,
    enhanced,
    drop shadow
]
\scriptsize
\setlength{\parskip}{0pt}
\setlength{\itemsep}{0pt}

 (r8c1, r8c2, r8c3, r8c4, r8c5, r8c6, r8c7, r8c8, r8c9),

 (r9c1, r9c2, r9c3, r9c4, r9c5, r9c6, r9c7, r9c8, r9c9))</answer>

\texttt{\char96}\texttt{\char96}\texttt{\char96}

Apply these rules strictly:

- Emit no text, Markdown fence, reasoning, or verification report outside or
  inside the `<answer>` element.

- Use integer literals, not strings.

- Use parentheses, not square brackets.

- Return exactly one complete, verified grid.
\end{tcolorbox}
\end{figure*}

\begin{figure*}[t]
\centering
\begin{tcolorbox}[
    colback=promptbg!100!white,
    sharp corners=south,
    boxrule=0.25mm,
    fonttitle=\bfseries,
    title={\textbf{Sudoku-GPT-5-nano-Skill-5}},
    width=\textwidth,
    enhanced,
    drop shadow
]
\scriptsize
\setlength{\parskip}{0pt}
\setlength{\itemsep}{0pt}
\# Solve Sudoku

Complete the puzzle with constraint propagation and deterministic backtracking.
Keep candidate lists, guesses, and search traces private.
\par\vspace{\baselineskip}
\#\# Parse and Validate

1. Extract exactly nine rows with nine cells per row.

2. Interpret digits `1` through `9` as immutable givens and the prompt's blank
   marker as an empty cell.

3. Record the givens separately for final verification.

4. Confirm that no given digit is duplicated in any row, column, or 3x3 box.
   Never repair an invalid puzzle by changing a given.

For zero-based coordinates, identify a cell's box with:

\texttt{\char96}\texttt{\char96}\texttt{\char96}text

box(r, c) = 3 * (r // 3) + (c // 3)

\texttt{\char96}\texttt{\char96}\texttt{\char96}
\par\vspace{\baselineskip}
\#\# Maintain Constraints

Track the digits already used in each row, column, and box. For every empty
cell `(r, c)`, compute:

\texttt{\char96}\texttt{\char96}\texttt{\char96}text

candidates(r, c) =

  {1, 2, 3, 4, 5, 6, 7, 8, 9}

  - row\_used[r]

  - col\_used[c]

  - box\_used[box(r, c)]

\texttt{\char96}\texttt{\char96}\texttt{\char96}

Treat either of these conditions as an immediate contradiction:

- An empty cell has no candidate.

- A row, column, or box has a missing digit that cannot be placed in any of its
  empty cells.
\par\vspace{\baselineskip}
\#\# Propagate Forced Placements

Repeat until a full pass makes no progress:

1. Fill naked singles: empty cells with exactly one candidate.

2. Fill hidden singles: missing digits that appear in exactly one candidate set
   within a row, column, or box.

3. After every placement, update all affected constraints and check for a
   contradiction.

An incomplete grid after propagation is not a failure. Continue with search.
\par\vspace{\baselineskip}
\#\# Search Deterministically

When propagation stalls:

1. Select an empty cell with the fewest candidates.

2. Break ties in row-major order.

3. Try its candidates in ascending numerical order.

4. For each candidate, copy or checkpoint the complete branch state, place the
   candidate, propagate forced placements, and recurse.

5. On contradiction, restore the complete checkpoint before trying the next
   candidate.

Do not allow assignments or candidate eliminations from a failed branch to leak
into another branch. Continue until a complete solution is found.

\par\vspace{\baselineskip}
\#\# Verify Independently

Before answering, perform a fresh validation independent of the search state:

- The result contains exactly nine rows of nine integer digits.

- Every value is in `1..9`.

- Every original given remains unchanged.

- Every row equals the set `{1, 2, 3, 4, 5, 6, 7, 8, 9}`.

- Every column equals that set.

\end{tcolorbox}
\end{figure*}

\begin{figure*}[t]
\centering
\begin{tcolorbox}[
    colback=promptbg!100!white,
    sharp corners=south,
    boxrule=0.25mm,
    fonttitle=\bfseries,
    title={\textbf{Sudoku-GPT-5-nano-Skill-5 (continued)}},
    width=\textwidth,
    enhanced,
    drop shadow
]
\scriptsize
\setlength{\parskip}{0pt}
\setlength{\itemsep}{0pt}
- Every 3x3 box equals that set.

- No blank marker remains.

Never emit a partial or unverified grid.

\par\vspace{\baselineskip}
\#\# Format the Answer

Return exactly one `<answer>` element containing exactly one Python code block.
Inside the code block, return one outer tuple containing nine inner tuples in
row-major order:

<answer>

\texttt{\char96}\texttt{\char96}\texttt{\char96}python

((r1c1, r1c2, r1c3, r1c4, r1c5, r1c6, r1c7, r1c8, r1c9),

 (r2c1, r2c2, r2c3, r2c4, r2c5, r2c6, r2c7, r2c8, r2c9),

 (r3c1, r3c2, r3c3, r3c4, r3c5, r3c6, r3c7, r3c8, r3c9),

 (r4c1, r4c2, r4c3, r4c4, r4c5, r4c6, r4c7, r4c8, r4c9),

 (r5c1, r5c2, r5c3, r5c4, r5c5, r5c6, r5c7, r5c8, r5c9),

 (r6c1, r6c2, r6c3, r6c4, r6c5, r6c6, r6c7, r6c8, r6c9),

 (r7c1, r7c2, r7c3, r7c4, r7c5, r7c6, r7c7, r7c8, r7c9),

 (r8c1, r8c2, r8c3, r8c4, r8c5, r8c6, r8c7, r8c8, r8c9),

 (r9c1, r9c2, r9c3, r9c4, r9c5, r9c6, r9c7, r9c8, r9c9))

\texttt{\char96}\texttt{\char96}\texttt{\char96}
</answer>

Apply these rules strictly:

- Emit nothing before or after the `<answer>` element.

- Use integer literals, not strings.

- Use tuples, not lists.

- Return only the final grid; omit explanations and verification notes.
\end{tcolorbox}
\end{figure*}

\begin{figure*}[t]
\centering
\begin{tcolorbox}[
    colback=promptbg!100!white,
    sharp corners=south,
    boxrule=0.25mm,
    fonttitle=\bfseries,
    title={\textbf{Sudoku-GPT-5-nano-Skill-6}},
    width=\textwidth,
    enhanced,
    drop shadow
]
\scriptsize
\setlength{\parskip}{0pt}
\setlength{\itemsep}{0pt}
\# Solve Sudoku

Complete the puzzle with constraint propagation and exact backtracking. Treat
every given digit as immutable, keep all intermediate reasoning private, and
emit only a fully verified solution.
\par\vspace{\baselineskip}
\#\# Parse the Grid

1. Extract exactly nine rows with exactly nine cells per row.

2. Interpret `X` as an empty cell.

3. Interpret digits `1` through `9` as fixed clues.

4. Record the clues separately for final verification.

5. Confirm that the clues contain no duplicate digit in any row, column, or
   3x3 box.

Use zero-based indices internally. Map cell `(r, c)` to:

\texttt{\char96}\texttt{\char96}\texttt{\char96}text

box(r, c) = 3 * (r // 3) + (c // 3)

\texttt{\char96}\texttt{\char96}\texttt{\char96}

Never alter a clue to repair a contradiction.
\par\vspace{\baselineskip}
\#\# Maintain Constraints

Track the digits currently assigned in each unit:

- `row\_used[r]`

- `col\_used[c]`

- `box\_used[box(r, c)]`

For every empty cell `(r, c)`, compute:

\texttt{\char96}\texttt{\char96}\texttt{\char96}text

candidates(r, c) =

    {1, 2, 3, 4, 5, 6, 7, 8, 9}

    - row\_used[r]

    - col\_used[c]

    - box\_used[box(r, c)]

\texttt{\char96}\texttt{\char96}\texttt{\char96}

After each placement, update all three unit constraints and every affected
candidate set. Treat either condition as a contradiction:

- An empty cell has no candidates.

- A digit appears twice in a row, column, or box.

An incomplete grid with multiple candidates is a stalled deduction state, not
a failure.
\end{tcolorbox}
\end{figure*}

\begin{figure*}[t]
\centering
\begin{tcolorbox}[
    colback=promptbg!100!white,
    sharp corners=south,
    boxrule=0.25mm,
    fonttitle=\bfseries,
    title={\textbf{Sudoku-GPT-5-nano-Skill-6 (continued)}},
    width=\textwidth,
    enhanced,
    drop shadow
]
\scriptsize
\setlength{\parskip}{0pt}
\setlength{\itemsep}{0pt}
\par\vspace{\baselineskip}
\#\# Propagate Forced Moves

Repeat until a full pass makes no progress:

1. Fill each naked single: a cell with exactly one candidate.

2. Fill each hidden single: a missing digit that occurs in only one candidate
   set within a row, column, or box.

3. Apply only sound candidate eliminations, such as locked candidates or naked
   pairs, when they are tracked explicitly.

4. Recompute affected candidates after every placement or elimination.

5. Abandon the current branch immediately if a contradiction appears.

Do not make a placement merely because it looks plausible. Every deterministic
placement must follow from the current constraints.
\par\vspace{\baselineskip}
\#\# Search to Completion

When propagation stalls, use depth-first search:

1. Select an empty cell with the fewest candidates.

2. Break ties in row-major order.

3. Try candidate digits in ascending order.

4. Save the complete state before each trial.

5. Place the trial digit, propagate forced moves, and recurse.

6. If the branch fails, restore the complete saved state before trying the next
   candidate.

7. Backtrack when all candidates for the selected cell fail.

Restore the grid, used-digit sets, and all cached candidate eliminations.
Never allow deductions from a failed branch to leak into another branch. Do
not stop at a logical stalemate or return a partial grid.
\par\vspace{\baselineskip}
\#\# Verify Independently

Before formatting the answer, validate the completed grid from scratch rather
than trusting the search state:

1. Confirm that the grid has exactly nine rows and nine entries per row.

2. Confirm that every entry is an integer from `1` through `9`.

3. Confirm that every original clue remains at its original position.

4. Confirm that each row contains every digit from `1` through `9` exactly
   once.

5. Confirm that each column contains every digit from `1` through `9` exactly
   once.

6. Confirm that each 3x3 box contains every digit from `1` through `9` exactly
   once.

7. Confirm that no `X`, zero, string, or other placeholder remains.

Reject any completed grid that fails a check and resume search. Never emit an
unverified grid.
\par\vspace{\baselineskip}
\#\# Format the Answer

Return exactly one `<answer>` element. Place exactly one Python code block
inside it, containing a tuple of nine 9-element tuples in row-major order:

<answer>

\texttt{\char96}\texttt{\char96}\texttt{\char96}python

((r1c1, r1c2, r1c3, r1c4, r1c5, r1c6, r1c7, r1c8, r1c9),

 (r2c1, r2c2, r2c3, r2c4, r2c5, r2c6, r2c7, r2c8, r2c9),

 (r3c1, r3c2, r3c3, r3c4, r3c5, r3c6, r3c7, r3c8, r3c9),

 (r4c1, r4c2, r4c3, r4c4, r4c5, r4c6, r4c7, r4c8, r4c9),

 (r5c1, r5c2, r5c3, r5c4, r5c5, r5c6, r5c7, r5c8, r5c9),

 (r6c1, r6c2, r6c3, r6c4, r6c5, r6c6, r6c7, r6c8, r6c9),

 (r7c1, r7c2, r7c3, r7c4, r7c5, r7c6, r7c7, r7c8, r7c9),

 (r8c1, r8c2, r8c3, r8c4, r8c5, r8c6, r8c7, r8c8, r8c9),

 (r9c1, r9c2, r9c3, r9c4, r9c5, r9c6, r9c7, r9c8, r9c9))

\texttt{\char96}\texttt{\char96}\texttt{\char96}

</answer>

Apply these rules strictly:

- Emit no text before or after the `<answer>` element.

- Emit no reasoning, notes, headings, or verification report in the answer.

- Put only the Python code block inside the tags.

- Use integer literals, not strings.

- Use parentheses, not square brackets.

- Return exactly one complete, verified grid.
\end{tcolorbox}
\end{figure*}

\begin{figure*}[t]
\centering
\begin{tcolorbox}[
    colback=promptbg!100!white,
    sharp corners=south,
    boxrule=0.25mm,
    fonttitle=\bfseries,
    title={\textbf{Sudoku-GPT-5-nano-Skill-7}},
    width=\textwidth,
    enhanced,
    drop shadow
]
\scriptsize
\setlength{\parskip}{0pt}
\setlength{\itemsep}{0pt}
\# Solve Sudoku

Produce the valid completed grid while preserving every clue.
\par\vspace{\baselineskip}
\#\# Parse the puzzle

1. Extract exactly nine rows of nine cells from the prompt, ignoring surrounding prose and whitespace.

2. Treat each `X` as an empty cell and each digit `1`–`9` as an immutable clue.

3. Reject no clue silently. If the input is malformed or the clues already violate a row, column, or box constraint, do not invent a solution.
\par\vspace{\baselineskip}
\#\# Solve

1. For every empty cell, track the digits absent from its row, column, and 3×3 box.

2. Repeatedly place naked singles and hidden singles.

3. When propagation stalls, choose an unfilled cell with the fewest candidates and try its candidates in ascending order.

4. After each trial placement, propagate constraints again. Backtrack immediately on any cell with no candidate or any duplicated digit in a unit.

5. Continue until the grid is complete. The task set is expected to contain a valid, unique solution.
\par\vspace{\baselineskip}
\#\# Verify

Before responding, confirm all of the following:

- The grid has exactly nine rows and nine integers per row.

- Every value is an integer from `1` through `9`.

- Every original clue remains unchanged in its original position.

- Every row, column, and 3×3 box has the set `{1, 2, 3, 4, 5, 6, 7, 8, 9}`.

- No `X`, placeholder, string, or list remains.

Do not emit an unverified grid.
\par\vspace{\baselineskip}
\#\# Return the result

Return only one Python tuple-of-tuples enclosed by `<answer>` and `</answer>`. Include no explanation, label, Markdown fence, or other text.

Use this exact shape:

\texttt{\char96}\texttt{\char96}\texttt{\char96}text

<answer>

((d11, d12, d13, d14, d15, d16, d17, d18, d19),

 (d21, d22, d23, d24, d25, d26, d27, d28, d29),

 (d31, d32, d33, d34, d35, d36, d37, d38, d39),

 (d41, d42, d43, d44, d45, d46, d47, d48, d49),

 (d51, d52, d53, d54, d55, d56, d57, d58, d59),

 (d61, d62, d63, d64, d65, d66, d67, d68, d69),

 (d71, d72, d73, d74, d75, d76, d77, d78, d79),

 (d81, d82, d83, d84, d85, d86, d87, d88, d89),

 (d91, d92, d93, d94, d95, d96, d97, d98, d99))

</answer>

\texttt{\char96}\texttt{\char96}\texttt{\char96}

Replace every `drc` placeholder with the solved integer at row `r`, column `c`.

\end{tcolorbox}
\end{figure*}

\begin{figure*}[t]
\centering
\begin{tcolorbox}[
    colback=promptbg!100!white,
    sharp corners=south,
    boxrule=0.25mm,
    fonttitle=\bfseries,
    title={\textbf{Sudoku-GPT-5-nano-Skill-8}},
    width=\textwidth,
    enhanced,
    drop shadow
]
\scriptsize
\setlength{\parskip}{0pt}
\setlength{\itemsep}{0pt}
\# Solve Sudoku

Solve the puzzle with constraint propagation and deterministic backtracking.
Preserve every given clue. Keep all reasoning, candidates, and search traces
private.
\par\vspace{\baselineskip}
\#\# Parse and Validate

1. Extract exactly nine rows containing exactly nine cells each.

2. Interpret `X` as an empty cell and digits `1` through `9` as immutable
   givens.

3. Record the givens separately for final verification.

4. Reject malformed input or givens that already duplicate a digit in any row,
   column, or 3x3 box. Never modify a given to repair a contradiction.

For zero-based coordinates, identify the box containing `(r, c)` with:

\texttt{\char96}\texttt{\char96}\texttt{\char96}text

box(r, c) = 3 * (r // 3) + (c // 3)

\texttt{\char96}\texttt{\char96}\texttt{\char96}

\par\vspace{\baselineskip}
\#\# Maintain Constraints

\end{tcolorbox}
\end{figure*}

\begin{figure*}[t]
\centering
\begin{tcolorbox}[
    colback=promptbg!100!white,
    sharp corners=south,
    boxrule=0.25mm,
    fonttitle=\bfseries,
    title={\textbf{Sudoku-GPT-5-nano-Skill-8 (continued)}},
    width=\textwidth,
    enhanced,
    drop shadow
]
\scriptsize
\setlength{\parskip}{0pt}
\setlength{\itemsep}{0pt}
Track the digits already used in every row, column, and box:

- `row\_used[r]`

- `col\_used[c]`

- `box\_used[box(r, c)]`

For each empty cell `(r, c)`, compute:

\texttt{\char96}\texttt{\char96}\texttt{\char96}text

candidates(r, c) =

  {1, 2, 3, 4, 5, 6, 7, 8, 9}

  - row\_used[r]

  - col\_used[c]

  - box\_used[box(r, c)]

\texttt{\char96}\texttt{\char96}\texttt{\char96}

Treat any of these conditions as a contradiction:

- An empty cell has no candidate.

- A row, column, or box contains a duplicate digit.

- A missing digit has no legal position in a row, column, or box.
\par\vspace{\baselineskip}
\#\# Propagate and Track Progress

Define one propagation cycle as one complete sweep over the grid and its rows,
columns, and boxes. During each cycle:

1. Place naked singles: empty cells with exactly one candidate.

2. Place hidden singles: missing digits that have exactly one legal position
   within a row, column, or box.

3. Update affected constraint state immediately after every placement.

4. Stop the current branch immediately if a contradiction appears.

5. Record whether the cycle placed at least one digit.

Maintain a branch-local `stagnation\_count`:

- Reset it to `0` after any cycle that places a digit.

- Increment it after a cycle that places no digit.

- Never continue propagation after it reaches `60`.

A no-progress cycle normally means that propagation has reached a fixed point;
continue that branch with exact search instead of repeating an unchanged sweep.
The 60-cycle limit is a defensive hard stop against accidental repetition. If
the same branch nevertheless reaches 60 consecutive no-progress cycles, abandon
it as failed. If no branch remains, return the failure signal defined below.
\par\vspace{\baselineskip}
\#\# Search Deterministically

When propagation stalls before the grid is complete:

1. Select an empty cell with the fewest candidates.

2. Break ties in row-major order.

3. Try candidate digits in ascending order.

4. Save the complete branch state before each trial.

5. Place the candidate, reset that child branch's `stagnation\_count` to `0`,
   propagate forced placements, and recurse.

6. On contradiction or branch failure, restore the complete saved state and
   try the next candidate.

7. Backtrack when every candidate for the selected cell fails.

Restore the grid, used-digit sets, cached candidates, and progress counter
during rollback. Never let state from a failed branch leak into another branch.
Do not treat an ordinary logical stalemate as proof that the puzzle is
unsolvable.
\par\vspace{\baselineskip}
\#\# Verify Independently

Before answering, validate the completed grid from scratch:

- It has exactly nine rows and nine integer entries per row.

- Every entry is between `1` and `9`.

- Every original given remains unchanged.

- Every row contains the digits `1` through `9` exactly once.

- Every column contains the digits `1` through `9` exactly once.

- Every 3x3 box contains the digits `1` through `9` exactly once.

- No `X` or other placeholder remains.

Reject any completed branch that fails a check and resume search. Never emit a
partial or unverified grid.
\par\vspace{\baselineskip}
\#\# Format the Final Output

On success, emit exactly one `<answer>` element containing exactly one Python
code block:

<answer>

\texttt{\char96}\texttt{\char96}\texttt{\char96}python

((r1c1, r1c2, r1c3, r1c4, r1c5, r1c6, r1c7, r1c8, r1c9),

 (r2c1, r2c2, r2c3, r2c4, r2c5, r2c6, r2c7, r2c8, r2c9),

 (r3c1, r3c2, r3c3, r3c4, r3c5, r3c6, r3c7, r3c8, r3c9),

 (r4c1, r4c2, r4c3, r4c4, r4c5, r4c6, r4c7, r4c8, r4c9),

 (r5c1, r5c2, r5c3, r5c4, r5c5, r5c6, r5c7, r5c8, r5c9),

\end{tcolorbox}
\end{figure*}

\begin{figure*}[t]
\centering
\begin{tcolorbox}[
    colback=promptbg!100!white,
    sharp corners=south,
    boxrule=0.25mm,
    fonttitle=\bfseries,
    title={\textbf{Sudoku-GPT-5-nano-Skill-8 (continued)}},
    width=\textwidth,
    enhanced,
    drop shadow
]
\scriptsize
\setlength{\parskip}{0pt}
\setlength{\itemsep}{0pt}

 (r6c1, r6c2, r6c3, r6c4, r6c5, r6c6, r6c7, r6c8, r6c9),

 (r7c1, r7c2, r7c3, r7c4, r7c5, r7c6, r7c7, r7c8, r7c9),

 (r8c1, r8c2, r8c3, r8c4, r8c5, r8c6, r8c7, r8c8, r8c9),

 (r9c1, r9c2, r9c3, r9c4, r9c5, r9c6, r9c7, r9c8, r9c9))

\texttt{\char96}\texttt{\char96}\texttt{\char96}

</answer>

Apply these rules strictly:

- Emit nothing before or after the `<answer>` element.

- Use integer literals, not strings.

- Use tuples, not lists.

- Return exactly one complete grid.

- Omit explanations and verification notes.

If the input is malformed, exhaustive search proves that no solution exists, or
all remaining branches hit the stagnation cap, emit exactly:

\texttt{\char96}\texttt{\char96}\texttt{\char96}text

<answer>UNSOLVABLE</answer>

\texttt{\char96}\texttt{\char96}\texttt{\char96}

Never emit both a grid and a failure signal.
\end{tcolorbox}
\end{figure*}

\begin{figure*}[t]
\centering
\begin{tcolorbox}[
    colback=promptbg!100!white,
    sharp corners=south,
    boxrule=0.25mm,
    fonttitle=\bfseries,
    title={\textbf{Sudoku-GPT-5-nano-Skill-9}},
    width=\textwidth,
    enhanced,
    drop shadow
]
\scriptsize
\setlength{\parskip}{0pt}
\setlength{\itemsep}{0pt}
\# Solve Sudoku

Complete the puzzle with constraint propagation and exact backtracking. Keep
candidate lists, guesses, and search traces private. Return only a verified
solution in the required format.
\par\vspace{\baselineskip}
\#\# Parse and Validate the Grid

1. Extract exactly nine rows with exactly nine cells per row.

2. Interpret `X`, `.`, and `0` as empty cells.

3. Interpret digits `1` through `9` as immutable givens.

4. Record the givens separately for final verification.

5. Reject malformed input or givens that already repeat a digit within a row,
   column, or 3x3 box.

Use zero-based row and column indices. Identify the box containing `(r, c)` as:

\texttt{\char96}\texttt{\char96}\texttt{\char96}text

box(r, c) = 3 * (r // 3) + (c // 3)

\texttt{\char96}\texttt{\char96}\texttt{\char96}

Never change a given to repair a contradiction.
\par\vspace{\baselineskip}
\#\# Maintain Constraints

Track:

- `row\_used[r]`: digits assigned in row `r`

- `col\_used[c]`: digits assigned in column `c`

- `box\_used[b]`: digits assigned in box `b`

For every empty cell `(r, c)`, compute:

```text

candidates(r, c) =

    {1, 2, 3, 4, 5, 6, 7, 8, 9}

    - row\_used[r]

    - col\_used[c]

    - box\_used[box(r, c)]

```

After every placement, update the grid and all affected constraint state.
Treat either condition as a contradiction:

- An empty cell has no candidate.

- A row, column, or box contains a duplicate digit.

- A missing digit has no legal position in a row, column, or box.

An incomplete consistent grid is a stalled state, not a failure.
\par\vspace{\baselineskip}
\#\# Propagate Forced Placements

Repeat until a complete pass makes no progress:

1. Place every naked single: an empty cell with exactly one candidate.

2. Place every hidden single: a missing digit that can occur in exactly one empty cell of a row, column, or box.
\end{tcolorbox}
\end{figure*}

\begin{figure*}[t]
\centering
\begin{tcolorbox}[
    colback=promptbg!100!white,
    sharp corners=south,
    boxrule=0.25mm,
    fonttitle=\bfseries,
    title={\textbf{Sudoku-GPT-5-nano-Skill-9 (continued)}},
    width=\textwidth,
    enhanced,
    drop shadow
]
\scriptsize
\setlength{\parskip}{0pt}
\setlength{\itemsep}{0pt}
3. Recompute affected candidates after each placement.

4. Stop the current branch immediately when a contradiction appears.

Process cells and units in row-major order whenever an ordering choice is
needed.
\par\vspace{\baselineskip}
\#\# Search to Completion

When propagation stalls before the grid is complete:

1. Choose an empty cell with the fewest candidates (MRV).

2. Break ties in row-major order.

3. Try candidate digits in ascending order.

4. Save the complete branch state before each trial placement.

5. Place the candidate, propagate forced placements, and recurse.

6. If the branch contradicts a constraint, restore the saved state and try the
   next candidate.

7. Backtrack when every candidate for the selected cell fails.

Restore the grid, used-digit sets, and any cached candidate state during
rollback. Never allow deductions from a failed branch to leak into another
branch. Do not stop at a logical stalemate and do not return a partial grid.
\par\vspace{\baselineskip}
\#\# Verify Independently

Before formatting the answer, check the completed grid from scratch:

1. Confirm that it contains exactly nine rows and nine entries per row.

2. Confirm that every entry is an integer from `1` through `9`.

3. Confirm that every original given remains unchanged.

4. Confirm that every row contains each digit from `1` through `9` exactly
   once.

5. Confirm that every column contains each digit from `1` through `9` exactly
   once.

6. Confirm that every 3x3 box contains each digit from `1` through `9` exactly
   once.

7. Confirm that no empty marker remains.

Reject any completed branch that fails verification and resume search. Never
emit an unverified grid.
\par\vspace{\baselineskip}
\#\# Output Contract

On success, emit exactly one `<answer>` element containing only a Python tuple
of nine 9-element row-tuples:

\texttt{\char96}\texttt{\char96}\texttt{\char96}text

<answer>

((r1c1, r1c2, r1c3, r1c4, r1c5, r1c6, r1c7, r1c8, r1c9),

 (r2c1, r2c2, r2c3, r2c4, r2c5, r2c6, r2c7, r2c8, r2c9),

 (r3c1, r3c2, r3c3, r3c4, r3c5, r3c6, r3c7, r3c8, r3c9),

 (r4c1, r4c2, r4c3, r4c4, r4c5, r4c6, r4c7, r4c8, r4c9),

 (r5c1, r5c2, r5c3, r5c4, r5c5, r5c6, r5c7, r5c8, r5c9),

 (r6c1, r6c2, r6c3, r6c4, r6c5, r6c6, r6c7, r6c8, r6c9),

 (r7c1, r7c2, r7c3, r7c4, r7c5, r7c6, r7c7, r7c8, r7c9),

 (r8c1, r8c2, r8c3, r8c4, r8c5, r8c6, r8c7, r8c8, r8c9),

 (r9c1, r9c2, r9c3, r9c4, r9c5, r9c6, r9c7, r9c8, r9c9))

</answer>

\texttt{\char96}\texttt{\char96}\texttt{\char96}

Apply these rules strictly:

- Emit no prose, Markdown fence, reasoning, or verification report.

- Use integer literals, not strings.

- Use parentheses, not square brackets.

- Return exactly one complete, verified grid.

If the input is malformed or exhaustive search proves that no solution exists,
emit exactly:

\texttt{\char96}\texttt{\char96}\texttt{\char96}text

<failure>UNSOLVABLE\_OR\_INVALID\_PUZZLE</failure>

\texttt{\char96}\texttt{\char96}\texttt{\char96}

Do not emit both an answer and a failure signal.

\end{tcolorbox}
\end{figure*}

\begin{figure*}[t]
\centering
\begin{tcolorbox}[
    colback=promptbg!100!white,
    sharp corners=south,
    boxrule=0.25mm,
    fonttitle=\bfseries,
    title={\textbf{Sudoku-GPT-5-nano-Skill-10}},
    width=\textwidth,
    enhanced,
    drop shadow
]
\scriptsize
\setlength{\parskip}{0pt}
\setlength{\itemsep}{0pt}
\# Solve Sudoku

Complete the puzzle with constraint propagation and deterministic backtracking.
Keep all reasoning, candidate lists, and search traces private.
\par\vspace{\baselineskip}
\#\# Parse and Validate

1. Extract exactly nine rows with exactly nine cells per row.

2. Interpret `X` as an empty cell and digits `1` through `9` as immutable
   givens.

3. Record the givens separately for final verification.

4. Reject malformed input or any puzzle whose givens already duplicate a digit
   in a row, column, or 3x3 box. Never alter a given to repair a contradiction.

Do not reject a puzzle merely because it has fewer than 17 givens. Clue count
alone does not determine whether a supplied puzzle has a valid completion.

For zero-based coordinates, identify a cell's box with:

\texttt{\char96}\texttt{\char96}\texttt{\char96}text

box(r, c) = 3 * (r // 3) + (c // 3)

\texttt{\char96}\texttt{\char96}\texttt{\char96}
\par\vspace{\baselineskip}
\#\# Maintain Constraints

Track the digits already used in each row, column, and box:

- `row\_used[r]`

- `col\_used[c]`

- `box\_used[box(r, c)]`

For each empty cell `(r, c)`, compute:

\texttt{\char96}\texttt{\char96}\texttt{\char96}text

candidates(r, c) =

  {1, 2, 3, 4, 5, 6, 7, 8, 9}

  - row\_used[r]

  - col\_used[c]

  - box\_used[box(r, c)]

\texttt{\char96}\texttt{\char96}\texttt{\char96}

Treat either condition as a contradiction:

- An empty cell has no candidate.

- A row, column, or box has a missing digit that cannot be placed in any of its
  empty cells.
\par\vspace{\baselineskip}
\#\# Propagate Forced Placements

Repeat until a complete pass makes no progress:

1. Fill every naked single: an empty cell with exactly one candidate.

2. Fill every hidden single: a missing digit that can occur in only one empty
   cell of a row, column, or box.

3. Update all affected constraints after every placement.

4. Stop the current branch immediately if a contradiction appears.

Treat stalled propagation as an incomplete state, not a failure. Continue with
exact search.
\par\vspace{\baselineskip}
\#\# Search Deterministically

When propagation stalls:

1. Select an empty cell with the fewest candidates.

2. Break ties in row-major order.

3. Try candidates in ascending numerical order.

4. Before each trial, save the complete branch state.

5. Place the candidate, propagate forced placements, and recurse.

6. On contradiction, restore the complete saved state and try the next
   candidate.

7. Backtrack when every candidate for the selected cell fails.

Restore the grid, used-digit sets, and any cached candidate state during
rollback. Never allow deductions from a failed branch to leak into another
branch.
\par\vspace{\baselineskip}
\#\# Verify Independently

Before answering, validate the completed grid from scratch:

- It has exactly nine rows and nine entries per row.

- Every entry is an integer from `1` through `9`.

- Every original given remains unchanged.

- Every row contains each digit `1` through `9` exactly once.

- Every column contains each digit `1` through `9` exactly once.

- Every 3x3 box contains each digit `1` through `9` exactly once.

- No `X` or other placeholder remains.

Reject a completed branch that fails any check and resume search. Never emit a
partial or unverified grid.

\end{tcolorbox}
\end{figure*}

\begin{figure*}[t]
\centering
\begin{tcolorbox}[
    colback=promptbg!100!white,
    sharp corners=south,
    boxrule=0.25mm,
    fonttitle=\bfseries,
    title={\textbf{Sudoku-GPT-5-nano-Skill-10 (continued)}},
    width=\textwidth,
    enhanced,
    drop shadow
]
\scriptsize
\setlength{\parskip}{0pt}
\setlength{\itemsep}{0pt}
\par\vspace{\baselineskip}
\#\# Format the Answer

On success, return exactly one `<answer>` element containing exactly one Python
code block. Inside the code block, return one outer tuple containing nine
9-element row-tuples in row-major order:

<answer>

\texttt{\char96}\texttt{\char96}\texttt{\char96}python

((r1c1, r1c2, r1c3, r1c4, r1c5, r1c6, r1c7, r1c8, r1c9),

 (r2c1, r2c2, r2c3, r2c4, r2c5, r2c6, r2c7, r2c8, r2c9),

 (r3c1, r3c2, r3c3, r3c4, r3c5, r3c6, r3c7, r3c8, r3c9),

 (r4c1, r4c2, r4c3, r4c4, r4c5, r4c6, r4c7, r4c8, r4c9),

 (r5c1, r5c2, r5c3, r5c4, r5c5, r5c6, r5c7, r5c8, r5c9),

 (r6c1, r6c2, r6c3, r6c4, r6c5, r6c6, r6c7, r6c8, r6c9),

 (r7c1, r7c2, r7c3, r7c4, r7c5, r7c6, r7c7, r7c8, r7c9),

 (r8c1, r8c2, r8c3, r8c4, r8c5, r8c6, r8c7, r8c8, r8c9),

 (r9c1, r9c2, r9c3, r9c4, r9c5, r9c6, r9c7, r9c8, r9c9))

\texttt{\char96}\texttt{\char96}\texttt{\char96}
</answer>

Apply these rules strictly:

- Emit nothing before or after the `<answer>` element.

- Use integer literals, not strings.

- Use tuples, not lists.

- Return exactly one complete grid; omit explanations and verification notes.

If exhaustive search proves that the puzzle has no solution, return exactly:

\texttt{\char96}\texttt{\char96}\texttt{\char96}text

UNSOLVABLE

\texttt{\char96}\texttt{\char96}\texttt{\char96}

\end{tcolorbox}
\end{figure*}

\begin{figure*}[t]
\centering
\begin{tcolorbox}[
    colback=promptbg!100!white,
    sharp corners=south,
    boxrule=0.25mm,
    fonttitle=\bfseries,
    title={\textbf{HMMT-GPT-5-nano-Example-Output}},
    width=\textwidth,
    enhanced,
    drop shadow
]
\scriptsize
\setlength{\parskip}{0pt}
\setlength{\itemsep}{0pt}

\textbf{[Question]}

Let $\lfloor z\rfloor$ denote the greatest integer less than or equal to $z$.
Compute
\[
\sum_{j=-1000}^{1000}
\left\lfloor \frac{2025}{j+0.5}\right\rfloor .
\]

\par\vspace{\baselineskip}
\textbf{[Zero-shot Response]}

Let
\[
S=\sum_{j=-1000}^{1000}
\left\lfloor \frac{2025}{j+1/2}\right\rfloor .
\]

Pair the terms for
$j\in\{-1000,\ldots,999\}$ with $k=-j-1$.
Then $j+1/2$ and $k+1/2$ are opposites, so with
$a=j+1/2$, we have
\[
\left\lfloor \frac{2025}{a}\right\rfloor
+
\left\lfloor \frac{2025}{-a}\right\rfloor
=
\lfloor x\rfloor+\lfloor -x\rfloor,
\]
where
\[
x=\frac{2025}{a}=\frac{4050}{2j+1}.
\]

If $2j+1$ divides $4050$, then $x$ is an integer and the
pair sums to $0$. Otherwise,
\[
\lfloor x\rfloor+\lfloor -x\rfloor=-1.
\]
Among $j\in\{-1000,\ldots,999\}$, the number of pairs is
$500$. The pair sums are $0$ precisely for those $j$ with
$2j+1$ dividing $4050$. The value
\[
4050=2\cdot 3^4\cdot 5^2,
\]
so the odd divisors of $4050$ are exactly the divisors of
$3^4\cdot 5^2$, giving $15$ positive divisors, including
$1$. Among these, the only divisor with magnitude greater
than $1999$ is $2025$, so the divisors satisfying
$|d|\leq 1999$ consist of $14$ positive and $14$ negative
divisors, giving $28$ possible values of $2j+1$. Each pair
corresponds to two values of $j$, namely those associated
with $d$ and $-d$, so the number of pairs with sum $0$ is
$28/2=14$.
\end{tcolorbox}
\end{figure*}

\begin{figure*}[t]
\centering
\begin{tcolorbox}[
    colback=promptbg!100!white,
    sharp corners=south,
    boxrule=0.25mm,
    fonttitle=\bfseries,
    title={\textbf{HMMT-GPT-5-nano-Example-Output (continued)}},
    width=\textwidth,
    enhanced,
    drop shadow
]
\scriptsize
\setlength{\parskip}{0pt}
\setlength{\itemsep}{0pt}
Thus, the sum over the $500$ pairs is
\[
-(500-14)=-486.
\]

The remaining unpaired term is $j=1000$:
\[
\left\lfloor \frac{2025}{1000.5}\right\rfloor
=
\left\lfloor \frac{4050}{2001}\right\rfloor
=2.
\]

Therefore,
\[
S=-486+2=-484.
\]

\[
\boxed{-484}
\]

\par\vspace{\baselineskip}
\textbf{Incorrect}

\par\vspace{\baselineskip}
\textbf{[\Ours{} Response]}

First rewrite the sum using $t_j=j+1/2$. Then $t_j$ runs
over
\[
-999.5,-998.5,\ldots,-0.5,0.5,\ldots,999.5
\]
and includes the additional value $1000.5$ for $j=1000$.

Pair the terms corresponding to $t$ and $-t$ for
\[
t=0.5,1.5,\ldots,999.5.
\]
For each pair, set
\[
a=\frac{2025}{t}>0.
\]
Then
\[
\lfloor a\rfloor+\lfloor -a\rfloor
=
\begin{cases}
0, & a\in\mathbb{Z},\\
-1, & a\notin\mathbb{Z}.
\end{cases}
\]

Thus, among the $1000$ pairs, each contributes $-1$ except
when $a$ is an integer. Now $a$ is an integer exactly when
$t=d/2$, where $d$ is an odd divisor of $4050$. The positive
odd divisors of $4050$ are
\[
3^\alpha 5^\beta,
\qquad
0\leq \alpha\leq 4,
\quad
0\leq \beta\leq 2,
\]
giving $15$ divisors in total. The largest is $2025$, which
corresponds to
\[
t=\frac{2025}{2}=1012.5,
\]
outside the range $0.5\leq t\leq 999.5$. Therefore, exactly
$14$ pairs contribute $0$, while the remaining $986$ pairs
contribute $-1$. Hence the paired terms sum to
\[
-986.
\]

The remaining unpaired term corresponds to $t=1000.5$:
\[
\left\lfloor \frac{2025}{1000.5}\right\rfloor
=
\left\lfloor \frac{4050}{2001}\right\rfloor
=2.
\]

Therefore, the total is
\[
-986+2=-984.
\]

\[
\boxed{-984}
\]

\par\vspace{\baselineskip}
\textbf{Correct}
\end{tcolorbox}
\end{figure*}

\begin{figure*}[t]
\centering
\begin{tcolorbox}[
    colback=promptbg!100!white,
    sharp corners=south,
    boxrule=0.25mm,
    fonttitle=\bfseries,
    title={\textbf{Sudoku-GPT-5-nano-Example-Output}},
    width=\textwidth,
    enhanced,
    drop shadow
]
\scriptsize
\setlength{\parskip}{0pt}
\setlength{\itemsep}{0pt}

\textbf{[Question]}

This is a standard $9\times 9$ Sudoku, where \texttt{X} needs to be
filled with digits $1$--$9$:

\begin{verbatim}
XX35X6XXX
768XX25XX
5X2X71XXX
8XXXX3X5X
XX5X1XXX9
XX19XX36X
XX9X3XXXX
XX7X5XX2X
XX42XX837
\end{verbatim}

Please complete this Sudoku.

Please solve according to Sudoku rules: the numbers $1$--$9$ must
appear exactly once in each row, column, and $3\times 3$ box.

Please provide your answer at the end using a Python Markdown code
block, represented as a tuple, for example:

\begin{verbatim}
```python
((1,2,3,4,5,6,7,8,9),(4,5,6,7,8,9,1,2,3),...)
```
\end{verbatim}
\par\vspace{\baselineskip}
\textbf{[Zero-shot Response]}
\begin{verbatim}
```python
((9, 1, 3, 5, 4, 6, 7, 8, 2),
 (7, 6, 8, 4, 9, 2, 5, 1, 3),
 (5, 4, 2, 3, 7, 1, 6, 9, 8),
 (8, 9, 6, 7, 2, 3, 4, 5, 1),
 (4, 3, 5, 6, 1, 8, 2, 7, 9),
 (2, 7, 1, 9, 8, 5, 3, 6, 4),
 (6, 2, 9, 8, 3, 7, 1, 4, 5),
 (3, 8, 7, 1, 5, 4, 9, 2, 6),
 (1, 5, 4, 2, 6, 9, 8, 3, 7))
```
\end{verbatim}

\textbf{Incorrect}

\par\vspace{\baselineskip}
\textbf{[\Ours{} Response]}

\begin{verbatim}
```python
((9, 1, 3, 5, 4, 6, 2, 7, 8),
 (7, 6, 8, 3, 9, 2, 5, 1, 4),
 (5, 4, 2, 8, 7, 1, 6, 9, 3),
 (8, 9, 6, 4, 2, 3, 7, 5, 1),
 (2, 3, 5, 6, 1, 7, 4, 8, 9),
 (4, 7, 1, 9, 8, 5, 3, 6, 2),
 (6, 2, 9, 7, 3, 8, 1, 4, 5),
 (3, 8, 7, 1, 5, 4, 9, 2, 6),
 (1, 5, 4, 2, 6, 9, 8, 3, 7))
```
\end{verbatim}

\textbf{Correct}

\end{tcolorbox}
\end{figure*}

\end{document}